%% file: main.tex
\documentclass[12pt]{article}

\usepackage[utf8]{inputenc}
\usepackage[bookmarks=false]{hyperref}
\usepackage{amsmath}
\usepackage{amsthm}
\usepackage{amsfonts}
\usepackage{enumitem}
\usepackage[mathscr]{euscript}
 \let\mathscr\relax% just so we can load this and rsfs
\usepackage[scr]{rsfso}
\usepackage{bbm}
\usepackage{enumitem}
\usepackage{array}
\usepackage{booktabs}% http://ctan.org/pkg/booktabs
\usepackage{hhline}
\usepackage{makecell}
\usepackage{caption} 
\usepackage[dvipsnames]{xcolor}
\usepackage{graphicx}
\usepackage{mathtools}
\usepackage{subcaption} % Include this in your preamble
\usepackage{subfiles}
\usepackage{tikz}
\usetikzlibrary{arrows.meta, positioning}
\usepackage{authblk}
\usepackage{blindtext}
\usepackage{thmtools, thm-restate}
\usepackage{listings}
\usepackage{multirow}
\usepackage{enumitem}
\usepackage{tikz}
\usepackage{subcaption}
\usepackage{float} 
\usepackage[square]{natbib} % for bibliography

\usepackage[linesnumbered,ruled]{algorithm2e}
\usepackage{ulem}
\newcounter{myalgocounter}
\makeatletter
\let\c@algocf\c@myalgocounter
\makeatother

\newenvironment{myalgorithm}[1][htbp]{%
    \refstepcounter{myalgocounter}
    \noindent\textbf{Algorithm \thealgocf: #1}
    \ignorespaces
}{\par
}

\usepackage{tcolorbox}
\tcbuselibrary{breakable}
\tcbset{%any default parameters
  width=\textwidth,
  halign=justify,
  center,
  breakable,
  colback=white    
}

\usepackage[noabbrev]{cleveref}
\crefformat{equation}{#2#1#3}
\crefrangeformat{equation}{#3#1#4 to~#5#2#6}

\usepackage{geometry}
\graphicspath{{Figures/}}

\renewcommand{\cdot}[1][.5]{%
  \mathbin{\vcenter{\hbox{\scalebox{#1}{$\bullet$}}}}%
}

\newcommand{\bigcdot}[1][.5]{%
  \mathbin{\vcenter{\hbox{\scalebox{#1}{$\bullet$}}}}%
}

\DeclareMathOperator*{\argmax}{arg\,max}

\newcommand{\pabc}{\hat{p}_{\epsilon}}
\newcommand{\pabco}[1]{\hat{p}_{#1}}

\newtheorem{theorem}{Theorem}

\newtheorem{lemma}[theorem]{Lemma}
\newtheorem{proposition}[theorem]{Proposition}
\newtheorem{definition}{Definition}
\newtheorem{assumption}{Assumption}
\newtheorem{example}{Example}
\theoremstyle{remark}

\newenvironment{modenumerate}[1][]
  {\enumerate[#1]\setupmodenumerate}
  {\endenumerate}

\newif\ifmoditem
\newcommand{\setupmodenumerate}{%
  \global\moditemfalse
  \let\origmakelabel\makelabel
  \def\moditem##1{\global\moditemtrue\def\mesymbol{##1}\item}%
  \def\makelabel##1{%
    \origmakelabel{##1\ifmoditem\rlap{\mesymbol}\fi\enspace}%
    \global\moditemfalse}%
}

\title{Bayesian learning of the optimal action-value function in a Markov decision process}

\author[1]{Jiaqi Guo\textsuperscript{\dag}}
\author[1]{Chon Wai Ho\textsuperscript{\dag}}
\author[2]{Sumeetpal S. Singh}

\affil[1]{Signal Processing and Communications Laboratory,
    Department of Engineering, University of Cambridge, UK}
\affil[2]{NIASRA, School of Mathematics and Applied Statistics, University
 of Wollongong, Wollongong, Australia}

\date{}
\begin{document}

\maketitle

\begin{abstract}

\subfile{sections/abstract}

\end{abstract}

\renewcommand{\thefootnote}{\fnsymbol{footnote}} % Change footnotes to symbols temporarily
\footnotetext[2]{Equal contributions.\\
\hangindent=1.4em Contact: J. Guo \href{mailto:jg851@cam.ac.uk}{jg85138@cam.ac.uk}; C.W Ho \href{mailto:cwh38@cam.ac.uk}{cwh38@cam.ac.uk}; S.S Singh \href{mailto:sumeetpals@uow.edu.au}{sumeetpals@uow.edu.au}}
\renewcommand{\thefootnote}{\arabic{footnote}} % Revert back to numbering for the rest of the footnotes

\newpage
\tableofcontents

\subfile{sections/intro}

\subfile{sections/prelim}

\subfile{sections/relatedwork}

\subfile{sections/learning}

\subfile{sections/simpleexample}

\subfile{sections/sampling}

\clearpage

\subfile{sections/experiment}

\subfile{sections/conclude}

\section*{Acknowledgements}
S.S.~Singh holds the Tibra Foundation professorial chair and gratefully acknowledges research funding as follows: ``This material is based upon work supported by the Air Force Office of Scientific Research under award number FA2386-23-1-4100''. C.W. Ho was supported by the UK Engineering and Physical Sciences Research Council (EPSRC) grant EP/T517847/1 for the University of Cambridge Doctoral Training Programme. J. Guo was supported by the China Scholarship Council for the PhD programme.

\clearpage

	\bibliographystyle{apalike}
	\bibliography{bib}

\clearpage

 \appendix

\subfile{sections/appendix}

\end{document}

%% file: sections/abstract.tex
The Markov Decision Process (MDP) is a popular framework for sequential decision-making problems, and uncertainty quantification is an essential component of it to learn optimal decision-making strategies. In particular, a Bayesian framework is used to maintain beliefs about the optimal decisions and the unknown ingredients of the model, which are also to be learned from the data, such as the rewards and state dynamics. However, many existing Bayesian approaches for learning the optimal decision-making strategy are based on unrealistic modelling assumptions and utilise approximate inference techniques. This raises doubts whether the benefits of Bayesian uncertainty quantification are fully realised or can be relied upon.

We focus on infinite-horizon and undiscounted MDPs, with finite state and action spaces, and a terminal state. We provide a full Bayesian framework, from modelling to inference to decision-making. For modelling, we introduce a likelihood function with minimal assumptions for learning the optimal action-value function based on Bellman's optimality equations, analyse its properties, and clarify connections to existing works. For deterministic rewards, the likelihood is degenerate and we introduce artificial observation noise to relax it, in a controlled manner, to facilitate more efficient Monte Carlo-based inference. For inference, we propose an adaptive sequential Monte Carlo algorithm to both sample from and adjust the sequence of relaxed posterior distributions. For decision-making, we choose actions using samples from the posterior distribution over the optimal strategies. While commonly done, we provide new insight that clearly shows that it is a generalisation of Thompson sampling from multi-arm bandit problems. Finally, we evaluate our framework on the Deep Sea benchmark problem and demonstrate the exploration benefits of posterior sampling in MDPs.\newline

\noindent Keywords: Bayesian Reinforcement Learning, Uncertainty quantification, Posterior Annealing, Sequential Monte Carlo (SMC), Markov chain Monte Carlo (MCMC)

%% file: sections/intro.tex
\section{Introduction}
In many sequential decision-making problems, the task is to optimise some quantifiable objective,  such as the cumulative rewards accrued over a series of decisions, but without all the information necessary to make the optimal choices. Consequently, interaction with the problem's environment is needed to acquire more information. However, frequent interactions may be computationally expensive or otherwise infeasible.\newline

A suitable strategy for interacting with the environment for the purpose of searching for the sequence of decisions that optimise the objective is thus required. An {\it exploitive} strategy is one that executes the ``best guess" of the optimal decision based on the available information at the decision times.  Alternatively, not using the best-guessed optimal decision, or even a random decision, is another option. This is known as an \emph{explorative} strategy and it could reveal new information that eventually leads to better decisions. Ideally, one should use a data-efficient strategy which balances the trade-offs between exploration and exploitation. In particular, it should produce a sequence of decisions to discover highly rewarding regions while ensuring sustained high rewards over the long term.\newline

An example of a strategy that balances exploration and exploitation can be found in the context of the multi-arm bandit (MAB) problem \citep{suttonbartobook,tstutorial}, which is a class of well-studied decision-making problems. Thompson sampling (TS)---also known as probability matching or posterior sampling \citep{ts1933,tstutorial}---is a Bayesian strategy that chooses decisions according to the posterior probability that the chosen action is optimal. TS has been proven to be near-optimal compared to other exploration-exploitation strategies \citep{tsagrawal2013contextual,tsshidong2018contextualbayesregret}. \newline

This idea of using the Bayesian posterior distribution to select the best action can be generalised to a more general class of decision-making problems, namely a Markov decision process (MDP) \citep{tsdeardenbayesq,tsstren,randomisedvaluefunction}. Unlike the MAB problem, in a MDP, the available decisions and the rewards received are determined by a time-varying internal state process that is Markovian \citep{putermanrl,suttonbartobook}; see Section \ref{sec:mdp} for more details. In this paper, we adopt the TS methodology to learn the optimal decisions of a MDP. We refer to this as posterior sampling. This raises two important challenges to be addressed. Firstly, how do we construct a Bayesian framework that meaningfully quantifies the uncertainty of an action being optimal? Secondly, how do we access the resulting posterior distribution? Both of these challenges are discussed further below as the modelling and inference challenges.\newline

\textbf{Modelling.} Extending the Bayesian formulation of the MAB problem to a MDP is not entirely straightforward. Unlike a MAB problem, in a MDP, the reward is both state and action dependent. Actions affect the state transition, through its transition probability density, and thus also the future rewards to be received. Consequently, actions may be taken for higher long-term incentives even when short-term incentives are low. This makes the Bayesian posterior of the optimality of decisions more challenging to characterise as it is no longer as straightforward as modelling the probability distribution of the \emph{immediate rewards}, as in the MAB problem. \newline

In a MDP, the optimal \emph{action-value function}, denoted $Q^*$, is the expected cumulated rewards when following the state transition dynamics under the optimal policy. It characterises the optimality of actions and uniquely satisfies a set of simultaneous equations known as the Bellman optimality equations (BOEs). Many existing Bayesian formulations (to be discussed in Section \ref{sec:relatedwork}) that learn $Q^*$ stem from the $Q$-learning algorithm, which is a stochastic approximation algorithm \citep{bertsekas2019} that incrementally updates $Q^*$ using the BOEs \citep{qlearning1992}. Specifically, in these works, the chosen likelihood function is motivated by a stochastic approximation procedure. Additionally, some also rely on unjustified and/or implicit assumptions. Thus, the resulting Bayesian formulation is highly nuanced, lacks interpretability, and may not faithfully quantify the residual uncertainty after assimilating the data---through the likelihood---with the adopted prior distribution. This could diminish the effectiveness of TS too, as it relies on this posterior distribution to make the action choices. \newline

 In this paper, we propose a new parametric Bayesian formulation for learning $Q^*$ that avoids these mentioned shortcomings. Unlike previous works, we construct the likelihood function using the BOEs directly.\newline

\textbf{Inference.} To generate new policies using TS, we need to sample from an updated posterior distribution regularly. Existing works that adopt a Bayesian treatment for learning $Q^*$ have primarily used optimisation-based methods to produce samples from the posterior distribution \citep{dropout,randomisedvaluefunction}. Furthermore, as remarked previously, these works have posterior distributions defined differently from ours, meaning that their inference methods are not directly applicable to our formulation. In contrast, we aim to better understand and more faithfully leverage the posterior distribution for exploration, for which Bayesian sampling methods seem necessary. We employ sequential Monte Carlo (SMC) combined with Markov chain Monte Carlo (MCMC) mutation kernels to sample from the sequence of posterior distributions, which helps maintain particles scattered around regions of high probability density, thereby reducing the risk of samples getting stuck in around a single local mode--a common issue with MCMC alone \citep{smc2006}. See Section \ref{sec:relatedwork} for greater details.\newline

However, sampling from the resulting posterior is challenging for several reasons. Firstly, as we will discuss later, incomplete exploration of the MDP state space implies only a subset of BOEs are observed. In such scenarios, there are generally infinitely many solutions that satisfy this subset of BOEs within a parametric class which is at least as expressive as the tabular representation of $Q^*$. Due to this lack of identifiability, the resulting posterior will have mass spanning over a large volume in Euclidean space with non-convex contours if the prior is not sufficiently localised. Secondly, for deterministic rewards, the likelihood is degenerate and requires the introduction of observation noise for the samplers to function. But, as we demonstrate, it is crucial for the noise to be small to approximate the true posterior distribution well. This leads to a trade-off between sampling error and approximation error. Thirdly, SMC can perform poorly without intermediate tempering distributions to bridge between successive target distributions, and the problem of monitoring and ensuring the mutation kernel remains effective in an online exploration-exploitation Reinforcement Learning (RL) setting hasn't been addressed before. Finally, without further approximations to the sampler, the computational cost is quadratic in time as data arrives due to the evaluations of the likelihood by the MCMC kernels. Thus, there is a trade-off between the computational budget and the overall error of the posterior samples.\newline

For the moment, we remark that \citet{tdsmc} also uses SMC for learning $Q^*$. However, their posterior formulation differs from ours and their approach emphasises computational efficiency by using data sub-sampling to approximate the costly components of the algorithm without monitoring the MCMC effectiveness. A more detailed discussion of related work will be given in Section \ref{sec:relatedwork}.\newline

\textbf{Our approach and contributions.} We propose a Bayesian solution for learning the optimal policy for a MDP, with an emphasis on data efficiency. Our focus is on finite state-space MDPs; with rewards that are either deterministic or have to be learned from a parameterised family of distributions; and the transition probabilities of a state-action pair are known or revealed when explored--as discussed in the paper, these restrictions can be relaxed to cover more general MDPs. Furthermore, our exposition is for an infinite-horizon and undiscounted MDP, with an absorbing terminal state, which is a class of problems also known as \emph{stochastic shortest path} (SSP) \citep{putermanrl,bertseka1991}. The absence of a discount factor requires an absorbing terminal state for the objective to be well-defined. The main contributions of this work are as follows:
\begin{enumerate}
\item We formulate a likelihood function for learning $Q^*$ that directly enforces the subset of BOEs implied by the state-action pairs in the dataset. We characterise the properties of the likelihood when the adopted parametrisation of $Q^*$ creates recurrent non-goal states, which can be difficult to avoid in undiscounted infinite-horizon problems. For MDPs with deterministic rewards, additional observation noise is introduced to ensure effective Monte Carlo sampling. This artificial noise is treated in our methodology as a further layer of approximation that we control, in contrast to other Bayesian reinforcement learning approaches that arbitrarily set a fixed noise level.
\item With an appropriate Bayesian formulation in hand, we apply posterior sampling for exploration and show how it connects to, and generalises, Thompson sampling as used in MAB problems. We also derive the exact posterior probabilities (up to Gaussian integrals) for selecting optimal actions under the tabular representation of $Q^*$. However, since its computation does not scale well with the dimensions of the state and action spaces, we instead pursue Monte Carlo methods.
\item For inference, we use SMC to update the sequence of posterior distributions as data arrives sequentially. We propose an annealing scheme to bridge the target distributions by gradually decreasing the observation noise, while being guided by the effective sample size (ESS) \citep{abcsmc2011}. We use Hamiltonian Monte Carlo (HMC) as the MCMC mutation kernel, with hyperparameters adapted using the SMC particles following a modification of \cite{smchmctuning}. For MDPs with deterministic rewards, we adaptively adjust the artificial observation noises as the RL episodes progress. The adjustments are guided by monitoring the MCMC effectiveness and improvements in the empirical expected squared error of the BOEs under the SMC samples. Our methodology aims to ensure effective MCMC performance while maintaining low noise levels.
\item We present extensive numerical experiments on the Deep Sea benchmark problem \citep{randomisedvaluefunction} to demonstrate our framework's ability to quantify uncertainty and highlight the exploration benefits of posterior sampling, namely its data-efficient properties.
\item We discuss further challenges, and suggest potential solutions, for Bayesian learning of $Q^*$, such as: (i) Addressing the intractable likelihood for an infinite state-space or with stochastic state transitions. (ii) Selecting appropriate priors that incorporate additional information that the likelihood will fail to capture when exploration is incomplete. (iii) Convergence monitoring during sampling, managing the trade-off between approximation and sampling errors, and mitigating the non-linear computational complexity over time.
\end{enumerate}

The paper is structured as follows. Section \ref{sec:prelim} provides a brief introduction to MDPs and the conditions sufficient to guarantee the uniqueness of solutions to the BOEs. Section \ref{sec:relatedwork} discusses related works and other existing Bayesian frameworks. Our Bayesian framework and its application for exploration via posterior sampling are detailed in Section \ref{sec:bayesianlearning}. Section \ref{sec:example} presents illustrative examples that highlight some challenges arising from the framework and motivate the issues the sampler must address. Section \ref{sec:sampling} describes the sampling algorithm. Experiments are presented in Section \ref{sec:experiment}. Finally, limitations, unresolved challenges and future directions are discussed in Section \ref{sec:conclude}.

\subsection{Notations}

For any set $\mathcal{X}$, let $\mathscr{P}(\mathcal{X})$ denote the set of all probability distributions over $\mathcal{X}$. For any $x,x^\prime \in \mathcal{X}$, let $\delta_{x^\prime}(x)$ be the Dirac delta function at $x$ centred at $x^\prime$. For any distribution $p \in \mathscr{P}(\mathcal{X})$ associated with a random variable $X$ taking values in $\mathcal{X}$, we denote its probability density function (pdf), if continuous, or probability mass function (pmf), if discrete, evaluated at $x$ as $p(x)$, with $p(\bigcdot)$ serving as the pdf or pmf. For discrete spaces, we use $\int_{\mathcal{X}} \mathrm{d} x$ and $\sum_{x \in \mathcal{X}}$ interchangeably. The expectation of any function $f:\mathcal{X} \rightarrow \mathbb{R}$ with respect to distribution $p$ is denoted as $\mathbb{E}_{X \sim p(\bigcdot)}[f(X)] = \int f(x) p(x) \mathrm{d}x$, or more simply as $\mathbb{E}_p[f(X)]$ or $\mathbb{E}[f(X)]$ when unambiguous. Let $\mathcal{Y}$ be another set. For conditional distributions $p$ of the form $\mathcal{X} \rightarrow \mathscr{P}(\mathcal{Y})$, let $X$ be a random variable taking values in $\mathcal{X}$ and $Y$ be a random variable taking values in $\mathcal{Y}$. The conditional distribution, pdf or pmf, of $Y$ given $X=x$ is denoted by $p(\bigcdot|x) \in \mathscr{P}({\mathcal{Y}})$, and $p(y|x)$ is the conditional pdf (or pmf )evaluated at $y$. Let $\mathcal{N}(\cdot;\mu,\epsilon^2)$ denote the univariate normal density with mean $\mu \in \mathbb{R}$ and variance $\epsilon^2$.\newline

For a real-value function $x \mapsto f(x)$, its support is defined to be $\mathrm{supp}(f)=\{x|f(x)\neq 0\}$. If $\mathcal{X} \subseteq \mathbb{R}^n$ for some  $n>1$, the $i$-th component of $x \in \mathcal{X}$ is denoted by $x_i$. Similarly, for integers $i<j$, we denote the vector $(x_i,x_{i+1},\dots,x_j)^T$ by $x_{i:j}$.\newline

%% file: sections/prelim.tex
\section{Preliminaries}
\label{sec:prelim}
\subsection{Introduction to MDPs}
\label{sec:mdp}
A discrete-time time-homogeneous infinite-horizon MDP is denoted by the collection of objects $\mathcal{M}=\{\mathcal{S}, \mathcal{A}, p^S, p^R, \rho\}$ \citep{putermanrl}, where $\mathcal{S}$ is the state space, $\mathcal{A}$ is the action space of the form $\mathcal{A} = \cup_{s \in \mathcal{S}} \mathcal{A}_s$, and $\mathcal{A}_s$ is the set of admissible actions for state $s \in \mathcal{S}$. For any $s \in \mathcal{S}$, $a \in \mathcal{A}_s$, $p^S(\bigcdot|s,a) \in \mathscr{P}(\mathcal{S})$ is the transition kernel of state-action pair $(s,a)$, along with the reward distribution $p^R(\bigcdot|s,a) \in \mathscr{P}(\mathbb{R})$. $\rho \in \mathscr{P}(\mathcal{S})$ is the initial state distribution.\newline

Let $\Pi = \{\pi: \mathcal{S} \rightarrow \mathscr{P}(\mathcal{A})| \forall s \in \mathcal{S}, \text{supp}(\pi(\bigcdot|s)) = \mathcal{A}_s\}$ be the set of Markovian decision rules.\footnote{A Markovian decision rule is conditioned on the current state only.} Actions are to be chosen at times $t \in \mathbb{Z}_{\geq 0}$ using a predetermined collection of such decision rules $\{\pi_t\}_{t \in \mathbb{Z}_{\geq 0}}$, where $\pi_t \in \Pi$; any such collection is called a policy. A policy is said to be stationary if it deploys the same decision rule $\{\pi,\pi,\ldots\}$ at all times, where $\pi \in \Pi$. We use the notation  $\pi \in \Pi$  for this stationary policy.\newline

At time $t=0$, the initial state $s_0$ is drawn from $\rho$. Assume the agent is using the stationary policy $\pi \in \Pi$. At any time $t\geq0$, let the agent be in state $s_t$.  The agent samples from the stationary policy, $A_t \sim \pi(\bigcdot|s_t)$,  to get a specific action $A_t=a_t$ to be applied. The reward $R_t \sim p^R(\bigcdot|s_t,a_t)$ and the next state $S_{t+1} \sim p^S(\bigcdot|s_t,a_t)$ are subsequently drawn by the MDP, realising $R_t=r_t$ and $S_{t+1}=s_{t+1}$. This process continues indefinitely, and up to time $t=\tau$, it induces a sequence of random variables $(S_0,A_0,R_0,S_1,\dots,S_{\tau-1},A_{\tau-1},R_{\tau-1},S_\tau)$ with corresponding realisations $(s_0,a_0,r_0,s_1,\dots,s_{\tau-1},a_{\tau-1},r_{\tau-1},s_{\tau})$, following the distribution
\begin{equation*}
p^\pi_{S_{0:\tau},A_{0:\tau-1},R_{0:r-1}}(s_{0:\tau},a_{0:\tau-1},r_{0:\tau-1}) = \rho(s_0)\prod_{t=0}^{\tau-1} \Big[\pi(a_t|s_t)p^R(r_t|s_t,a_t)p^S(s_{t+1}|s_t,a_t)\Big],
\end{equation*}
where $\pi$ emphasises its dependence on a policy $\pi$.\footnote{A time-inhomogeneous finite-horizon MDP, where $p^S$ and $p^R$ are time-dependent and the process terminates at a fixed time, can be reformulated as a time-homogeneous infinite-horizon MDP \citep{putermanrl} by augmenting the state space to include time.} To emphasise the dependence of $R_t$ on $S_t$ and $A_t$, we write $R(S_t,A_t) := R_t$. From here onwards, to simplify notation and enhance readability, we drop the subscript when unambiguous and denote the density $p^\pi(s_{0:\tau},a_{0:\tau-1},r_{0:\tau-1}) := p^\pi_{S_{0:\tau},A_{0:\tau-1},R_{0:r-1}}(s_{0:\tau},a_{0:\tau-1},r_{0:\tau-1})$. The same holds for its marginal and conditional probabilities, such as $p^\pi(s_{0:\tau},a_{0:\tau-1}) := p^\pi_{S_{0:\tau},A_{0:\tau-1}}(s_{0:\tau},a_{0:\tau-1})$. $\mathbb{E}^\pi$ will be used to denote expectation over $p^\pi$.\newline

Denote $\mathcal{S} \otimes \mathcal{A} := \bigcup_{s \in \mathcal{S}} \{s\} \times \mathcal{A}_s$. Our goal is to search for a policy, the optimal policy, that maximises the expected (discounted) cumulative rewards function, also known as the action value function, $Q^\pi:\mathcal{S} \otimes \mathcal{A} \rightarrow \mathbb{R}$:
\begin{equation*}
    Q^\pi(s,a) := \mathbb{E}^\pi\Bigg[\sum_{t=0}^{\infty} \gamma^t R_t \Bigg| S_0=s, A_0=a\Bigg],
\end{equation*}
where $s \in \mathcal{S}$, $a \in \mathcal{A}_s$, and $0\leq \gamma \leq 1$ is the discount factor. Under mild conditions \citep{putermanrl}, the optimal policy is Markovian, stationary and deterministic\footnote{A policy is deterministic if its decision rules are deterministic, i.e. $\pi(\bigcdot|s)$ is supported on one action only for all $s \in \mathcal{S}$.} for infinite-horizon MDPs. Assume that $|\mathcal{A}|$ is finite. Let the optimal action value function $Q^*(s,a) := \sup_{\pi \in \Pi}Q^\pi(s,a)$, and define an operator $\mathcal{B}^*_q$ on $\mathcal{S} \otimes \mathcal{A} \rightarrow \mathbb{R}$ such that for any $Q \in \{\mathcal{S} \otimes \mathcal{A} \rightarrow \mathbb{R}\}$,
\begin{equation}
    \mathcal{B}_q^*(Q)(s,a) := \mathbb{E} \Big[R_0+\gamma \max_{a_1 \in \mathcal{A}_{S_1}}Q(S_1, a_1) \Big|S_0=s, A_0=a\Big]. \label{eqn:bellmanoptim}
\end{equation}

It is easy to show that $Q^*$ satisfies the Bellman optimality equations (BOEs) $\mathcal{B}_q^*(Q^*) = Q^*$. Furthermore, define the function $V^\pi$ and $V^*$ such that $V^\pi(s) := \mathbb{E}^{\pi}[Q^\pi(S_0,A_0)|S_0=s]$ and $V^*(s) = \max_{a \in \mathcal{A}_s}Q^*(s,a)$. A policy $\pi^* \in \Pi$ is optimal if $V^{\pi^*} \equiv V^*$. \newline

Finally, suppose there exists a non-empty subset $\mathcal{S}^g \subset \mathcal{S}$ of absorbing states, where $\mathcal{S}^g = \{s^g \in \mathcal{S}| A_{s^g}=\{a^g\},p^S(s|s^g,a^g)=\delta_{s^g}(s), p^R(r|s^g,a^g)=\delta_{0}(r)\}$. When $|\mathcal{S}^g|=1$, it is commonly known as a stochastic shortest path model (SSP), which is a special class of MDPs where $s^g \in \mathcal{S}^g$ acts as the goal state\footnote{Any MDPs with non-empty $\mathcal{S}^g$ can be reformulated as an SSP by adding a single ultimate absorbing state that all existing absorbing states transition to. To see this, for MDPs which $|\mathcal{S}^g| > 1$, we can augment and extend the MDP by introducing an overall absorbing state-action pair $s_{\bar{g}}, a_{\bar{g}}$ such that for any $s^g \in \mathcal{S}^g$ with $A_{s^g} = \{a^g\}$, we redefine $p^S(s|s^g,a^g) = \delta_{s_{\bar{g}}}(s)$ and $p^R(r|s^g,a^g) = \delta_0(r)$ and set $s_{\bar{g}}$,$a_{\bar{g}}$ as the unique goal state-action pair.\newline}. In this paper, we restrict our discussion to MDPs with non-empty $\mathcal{S}^g$, which we simply refer to as SSPs, and set $\gamma=1$ by assuming that $p^{\pi^*}(S_t \in \mathcal{S}^g \text{ for some } t \in \mathbb{Z}_{>0})=1$. Our formulation can be generalised to a wider class of MDPs by setting $0 \leq \gamma<1$. Furthermore, we assume that $|\mathcal{S}|$ and $|\mathcal{A}|$ are finite, unless otherwise specified.

\subsection{Uniqueness of solutions to Bellman optimality equations}

We now give a sufficient condition for the Bellman operator $\mathcal{B}^*_q$ to have $Q^*$ as its unique fixed point. Assume $\mathcal{S}^g=\{s^g\}$. A deterministic stationary policy $\pi \in \Pi$ is proper if $\lim\limits_{t \rightarrow \infty} p^\pi(S_t=s^g|S_0=s_0) = 1$ for all $s_0 \in \mathcal{S}$, otherwise it is improper.

\begin{assumption}
    Assume that $\gamma=1$ and there exists a unique absorbing state $s^g$ with $\mathcal{A}_{s^g}=\{a^g\}$. In addition, assume that a proper deterministic stationary policy exists, and for any improper deterministic stationary policy $\pi$, there exists an initial state $s \in \mathcal{S}$ such that $V^\pi(s) = -\infty$. \label{ass:boeunique}
\end{assumption}

When Assumption \ref{ass:boeunique} holds, we have the following result for $V^*$.

\begin{theorem}[\citep{bertseka1991}]
    Suppose Assumption \ref{ass:boeunique} holds. Furthermore, the rewards are deterministic, i.e. $p^R$ is a Dirac function. Denote the conditional random variable (which is deterministic) as $r(s,a):=R_t|S_t=s,A_t=a$. Let $\mathcal{B}_v^*$ be an operator on $\mathcal{S} \rightarrow \mathbb{R}$ such that for any $V \in \mathcal{S} \rightarrow \mathbb{R}$,
    \begin{equation*}
    \mathcal{B}_v^*(V)(s) := \max_{a \in \mathcal{A}_s} r(s,a)+ \sum_{s^\prime \in \mathcal{S}} V(s^\prime) p^S(s^\prime|s,a) 
    \end{equation*} for all $s \in \mathcal{S}$. Then, 
    \begin{enumerate}
        \item 
        $V^*$ is the unique fixed point of $\mathcal{B}_v^*$ under $\{V:\mathcal{S} \rightarrow \mathbb{R}|V(s^g) = 0\}$.

    \item There exists an optimal stationary policy $\pi^*$ which is deterministic and proper, and is of the form 
    \begin{equation*}
        \pi^*(a|s) = \mathbbm{1}(a \in \argmax\limits_{a \in \mathcal{A}_s} r(s,a) + \sum_{s^\prime \in \mathcal{S}} V^*(s^\prime)p^S(s^\prime|s,a))  = \mathbbm{1}(a \in \argmax\limits_{a \in \mathcal{A}_s}(Q^*(s,a))).
    \end{equation*}
    \end{enumerate}
\label{thmvunique}
\end{theorem}

Now, the uniqueness result of Theorem \ref{thmvunique} can be applied to $Q^*$.
\begin{lemma}
     Suppose the conditions of Theorem \ref{thmvunique} hold. $V^*$ is the unique fixed point of $\mathcal{B}_v^*$ under $\{V:\mathcal{S} \rightarrow \mathbb{R}|V(s^g)=0\}$ if and only if $Q^*$ is the unique fixed point of $\mathcal{B}^*_q$ under $\{Q:\bigcup_{s \in \mathcal{S}} \{s\} \times \mathcal{A}_s \rightarrow \mathbb{R}|Q(s^g,a^g)=0\}$. \label{lem:qvuniqueness}
\end{lemma}
\begin{proof}
See Appendix \ref{apd:proofbellmanuniqueness}.
\end{proof}
Other sufficient conditions can be found, e.g. in \citep{bertseka1991,putermanrl,bertsekas2019,ssppolyhedral}. These results motivate us to find $Q^*$ using the BOEs and derive a policy from $Q^*$.\newline

%% file: sections/relatedwork.tex
\section{Related work}
\label{sec:relatedwork}

 Many existing works that adopt a Bayesian perspective to model the action-value function $Q^\pi$ or $Q^*$ as random functions are built upon classical non-Bayesian algorithms like Q-learning \citep{qlearning1992} and Bellman residual minimisation \citep{residualgradient_baird1995}. As we discuss below, applying these optimisation algorithms directly to a Bayesian setting often poses challenges. Simplifying but inadequately justified assumptions are often made for modelling ease and computational feasibility, leading to issues such as questionable time-inconsistent posterior definitions and biased likelihoods. The advantages of the resulting Bayesian model compared to its optimisation counterparts are thereby diminished.\newline

 \textbf{Related works based on temporal difference methods.} Q-learning can be viewed as a stochastic approximation method that iteratively estimates $Q^*(s,a)$ as a lookup table for each state-action pair. When extended to other (almost-everywhere differentiable) parametric approximations of $Q^*$ \citep{residualgradient_baird1995,fittedqiter,dqn_humanlevel}, denoted as $Q_\theta$, each iteration can be generalised as a stochastic gradient descent step for the sequence of mean-squared temporal difference error (MSTDE) minimisation objectives
$$\text{MSTDE}(\theta;\theta^t) := \mathbb{E}_{\substack{S,A \sim d(\bigcdot) \\ S^\prime \sim p^S(\cdot|S,A)}} \Big[(R(S,A) + \max_{a^\prime \in \mathcal{A}_{S^\prime}} Q_{\theta^t} (S^\prime,a^\prime) - Q_\theta(S,A))^2\Big],$$ where $\theta^{t+1}$ approximately minimises $\text{MSTDE}(\theta;\theta^t)$ given $\theta^t$, and $d(\bigcdot)$ is a state-action distribution specific to the algorithm \citep{qlearningtheory, td_optimperspective}. However, the iterative algorithm does not have an overall explicit optimisation objective, and may not converge for general classes of $Q_\theta$ unless $Q_\theta$ has a tabular representation \citep{qlearning1992,residualgradient_baird1995}. In practice, the $Q$-learning update is performed with various additional tricks and safeguards such as data sub-sampling (i.e. experience replay) and delayed update of $\theta^t$ (i.e. target network) for better stability \citep{neuralfittedqiter, dqn_humanlevel, doubledqn}.
\newline

 To learn $Q^*$ in a Bayesian framework, a common approach is to reframe the MSTDE objectives as a Bayesian regression problem. In particular, a prior is chosen for $\theta$ and the other learnable parameters; at iteration $t$, the MSTDE criterion is used to define the iteration-dependent Gaussian likelihood $L(\theta;\theta^t) := \prod_{i=0}^t \mathcal{N}(r_i+\max_{a^\prime \in \mathcal{A}_{s_{i+1}}}Q_{\theta^t}(s_{i+1},a^\prime);Q_{\theta}(s_i,a_i),\epsilon^2)$ with variance $\epsilon^2$, where $\theta^t$ acts as a deterministic point estimate for the true $Q^*$ at iteration $t$ chosen from past iterations; and the resulting posterior distribution is further approximated along with additional stability tricks. For example, see \citet{bayesianqlearning,dropout,randomisedpriorwithensembleosband,bootstraposband,randomisedvaluefunction, tdsmc}. However, these approaches raise several concerns. Firstly, the evolving definition of the likelihood function, due to the dependence of $\theta^t$ and the application of stability techniques, results in a shifting and inconsistent interpretation of the posterior uncertainty. Secondly, the Gaussian likelihood assumption with a predetermined observation variance is unrealistic except for the simplest MDPs, because the target $r_i + \max_{a^\prime \in \mathcal{A}_{s_{i+1}}} Q_{\theta^i} (s_{i+1},a^\prime)$ conditional on $\theta$ and $s_i$, $a_i$, $\theta^i$ is a non-linear transformation of the reward noise from $p^R$ and the transition noise from $p^S$. Thirdly, the randomness of $\theta^t$ is not properly taken into account in the construction of posterior distributions. Finally, just as $Q_{\theta^t}$ may fail to converge to $Q^*$ in $Q$-learning (e.g. due to model misspecifications and implementation tricks), the posterior of the Bayesian reformulation may also fail to concentrate at $Q^\ast$.  In fact, it is not clear that the posterior will at all concentrate. \newline

For inference, tabular methods such as \citet{bayesianqlearning} use modelling approximations to maintain a closed-form posterior. \citet{dropout} parametrised $Q_\theta$ with a neural network and optimised the MSTDE objective with stochastic gradient descent and dropout, which can be interpreted as a variational approximation to the posterior. However, \citep{randomisedpriorwithensembleosband} noted that such approximate posteriors do not concentrate with more data and can result in sub-optimal policies even as more data are collected. To address this, \citet{randomisedpriorwithensembleosband,randomisedvaluefunction} suggested an alternative posterior approximation through optimisation,  which injects noise to the maximum-a-posteriori objectives for approximated moments matching (exact for linear Gaussian models) along with nonparametric bootstrap (data sub-sampling) \citep{bootstrap} to generate ensemble estimates. More recently, \citet{tdsmc} applied SMC using noisy weight updates with stochastic gradient MCMC \citep{howgoodbayespos} as the mutation kernel, both facilitated with data subsampling and leading to asymptotic bias estimations. In contrast, our method differs from these approaches in targeting different posterior distributions. Also, our SMC sampling framework prioritises accuracy by suggesting relevant techniques and addressing these more computationally efficient yet biased approximations primarily through informal discussions.\newline

\textbf{Related works based on Bellman residuals.} Another objective to minimise to learn $Q^*$ is the mean-squared Bellman error,
$$\text{MSBE}(\theta) := \mathbb{E}_{S,A \sim d(\bigcdot)}\Big[(R+ \mathbb{E}_{S^\prime \sim p^S(\cdot|S,A)}\big[\max_{a^\prime \in \mathcal{A}_{S^\prime}} Q_\theta(S^\prime,a^\prime)|S,A\big]-Q_\theta(S,A))^2\Big],$$ in which the global minima matches $Q^*$ for MDPs with a unique solution to the BOEs\footnote{Note that since a similar objective can be applied to learning $Q^\pi$ for fixed policy $\pi$ with the Bellman equations \citep{residualgradient_baird1995,policyevaluationsurvey,suttonbartobook}, which is arguably simpler as it avoids the $\max$ operator, we review works that either learn $Q^\pi$ or $Q^*$ for MSBE methods. The Bellman equation has the form $\mathbb{E}^\pi \Big[R_0+\gamma Q^\pi(S_1, A_1) \Big|S_0=s, A_0=a\Big] = Q^\pi(s,a)$ for any $s \in \mathcal{S}, a \in \mathcal{A}_s$}. For online learning, it is known as the residual gradient method and $\theta$ is updated via stochastic gradient descent. In contrast to MSTDE-based approaches, the MSBE objective is time-consistent, does not depend on point estimates, and can be optimised offline directly after dataset collection \citep{lstdlinear}.\newline

Likewise, Bayesian regression can be constructed from the MSBE objective by defining a likelihood function $L(\theta) := \prod_{i=0}^t \mathcal{N}(r_i; Q_\theta(s_i,a_i) - \mathbb{E}_{S_i^\prime \sim p^S(\cdot|s_i,a_i)}[\max_{a^\prime \in \mathcal{A}_{S^\prime_i}} Q_{\theta} (S^\prime_i,a^\prime)|s_i,a_i], \epsilon^2)$. However, a key challenge of MSBE methods is that the inner expectation of the objective may become intractable when $p^S$ is unknown analytically or $\mathcal{S}$ is large or continuous, an issue that any Bayesian methods derived from the MSBE objective also inherit. To address this issue, several approaches have been proposed to obtain an unbiased estimator of the gradient of the empirical MSBE, commonly referred to as the double-sampling problem\footnote{To see this, note that the gradient of the empirical MSBE is proportional to $(\sum_{i=0}^t r_i + \mathbb{E}[\max_{a^\prime \in \mathcal{A}_{S_{i+1}}} Q_{\theta} (S_{i+1},a^\prime)|S_i=s_i, A_i=a_i] - Q_\theta(s_i,a_i))(\nabla_\theta (\sum_{i=0}^t \mathbb{E}[\max_{a^\prime \in \mathcal{A}_{S_{i+1}}} Q_{\theta} (S_{i+1},a^\prime)|S_i=s_i, A_i=a_i] - Q_\theta(s_i,a_i)))$. Thus, independent samples of $S_{i+1}$ are required to unbiasedly estimate the product $(\sum_{i=0}^t \mathbb{E}[\max_{a^\prime \in \mathcal{A}_{S_{i+1}}} Q_{\theta} (S_{i+1},a^\prime)|S_i=s_i, A_i=a_i])(\nabla_\theta (\sum_{i=0}^t \mathbb{E}[\max_{a^\prime \in \mathcal{A}_{S_{i+1}}} Q_{\theta} (S_{i+1},a^\prime)|S_i=s_i, A_i=a_i]))$. Similar reasoning applies to obtaining a low MSE (but generally biased) estimator of the (log) likelihood.} \citep{residualgradient_baird1995,policyevaluationsurvey}, or a low MSE estimator of the likelihood. \newline

 A common approach is to use additional independent Monte Carlo samples of state transitions in place of the inner expectation in a selected inference algorithm\citep{residualgradient_baird1995, policyevaluationsurvey}. Another strategy involves modelling $p^S$. For example, \citet{gprl-rasmussen} utilised Gaussian processes (GPs) to separately model $p^\mathcal{S}$ and $V^\pi$ and aligned their means according to the Bellman Equations. Alternatively, some methods introduce additional but often invalid likelihood assumptions to compensate for the bias when using the empirical next states to approximate the inner expectation instead. For instance, \citet{gpsarsa_engel} assumed the cumulative (discounted) rewards initiated at $(s,a)$ to follow a Gaussian distribution centred at a GP or a linear parametric model of $Q^\pi$ independently for every state-action pair $(s,a)$ for a given policy $\pi$, with pre-defined variances and computed the resulting posterior analytically. \citet{kalmantd} extended this to non-linear models of $Q^\pi$ and employed a state-space approach to address non-stationarity arising from the invalid assumptions, with the posterior estimated via the Kalman filtering paradigm. Note that the latter method is not directly applicable to learning $Q^*$ due to some undesirable dependencies implied by the assumptions \citet{kalmantd}. Finally, \citet{bayesianbellmanoperator} proposed directly learning the distribution underlying the inner expectation and optimising the posterior predictive MSBE objective.\newline

\textbf{Other relevant works.} Other Bayesian methods include \citet{probplanningsmc}, which uses SMC to target the distribution of MDP trajectories under a framework known as control as inference\citep{controlasinference_levine}, where trajectories are conditioned on being optimal under a specific notion of optimality. In contrast, our algorithm aims to sample from the posterior distribution of optimal policy under the notion of maximising $Q_\theta(s,a)$ (See Section \ref{sec:possamplingexploration}).\newline

%% file: sections/learning.tex
\section{Bayesian learning}
\label{sec:bayesianlearning}
Let $\mathcal{M}$ be the MDP of interest. In this paper, we make the assumption for $\mathcal{M}$ that the reward $R_t$ at any given time $t$ can be decomposed into its mean and a zero-mean noise, and the zero-mean noise has a known distribution.

\subsection{Bayesian formulation of learning \texorpdfstring{$Q^*$}{Q*}}
\subsubsection{A definition via Bellman optimality equations}
To turn the problem of learning $Q^*$ into a Bayesian problem, we model $Q^*$ of a MDP as a random variable with a prior distribution, and we are interested in the posterior given the interactions observed in the environment. In particular, we consider a parametric approximation to $Q^*$ as $Q_\theta$, such that $\theta \in \Theta \subseteq \mathbb{R}^{d_\Theta}$, and for any $s \in \mathcal{S}$, $a \in \mathcal{A}_s$, $Q_\theta(s,a) \in \mathbb{R}$. \newline

Define the prior distribution on $\Theta$ as $p^\Theta$. Let $\bar{r}_{s,a} := \mathbb{E}[R_0|S_0=s,A_0=a] \equiv \mathbb{E}[R_t|S_t=s,A_t=a]$ by stationarity, and for $s,a \in \mathcal{S} \otimes \mathcal{A}$, let $g_{s,a}: \Theta \rightarrow \mathbb{R}$ such that
$$g_{s,a}(\theta) := Q_\theta(s,a) - \mathbb{E}\Big[\max_{a^\prime \in \mathcal{A}_{S_{1}}}Q_\theta(S_{1}, a^\prime)|S_0=s, A_0=a\Big].$$

For a single realisation of $\mathcal{M}$ up to time $t=\tau$, let $\mathcal{D}_\tau^{\mathcal{S},\mathcal{A}} := \{(s,a)|s=s_t, a=a_t \text{ for some } t \in \{0,\dots,\tau\}\}$ and $\mathcal{D}^{\bar{r}}_\tau := \{\bar{r}_{s,a}|s,a \in \mathcal{D}_\tau^{\mathcal{S},\mathcal{A}}\}$. Assume the MDP of interest has a unique solution to the BOEs, and that there exists a $\theta^* \in \Theta$ such that $Q_{\theta^*} \equiv Q^*$, we impose the equality constraints to the likelihood function $p^*$ defined on $\bar{r}_{s,a} \in \mathcal{D}^{\bar{r}}_\tau$, of the form

\begin{equation*}
    p^*(\bar{r}_{s,a}|\theta,s,a) := \delta_{g_{s,a}(\theta)}(\bar{r}_{s,a}).
\end{equation*}

Then, we can apply Bayes' rule to infer the posterior
\begin{equation}
p^*(\theta|\mathcal{D}^{\bar{r}}_\tau,\mathcal{D}_\tau^{\mathcal{S},\mathcal{A}}) \propto p^\Theta(\theta) \prod_{s,a \in \mathcal{D}_\tau^{\mathcal{S},\mathcal{A}}} p^*(\bar{r}_{s,a}|\theta,s,a).
\label{eqn:llhmodel}
\end{equation}

Hence, the uncertainty of $Q^*$ after observing the data originates from our prior belief constrained on the subset of $\Theta$ such that the corresponding subsets of BOEs are satisfied. In practice, modifications to Equation \ref{eqn:llhmodel} are needed depending on whether rewards are deterministic or stochastic for traceability, and to account for cases where the parametric class of $Q_\theta$ may not contain $Q^*$, i.e. the likelihood is misspecified.

\subsubsection{Deterministic rewards}

When $p^R$ is deterministic, the expected rewards $\bar{r}_{s,a}$ are observed directly. However, the degenerate likelihood in Equation \ref{eqn:llhmodel} needs to be relaxed to maintain tractability and allow for easier inference. For example, designing a proposal kernel for an MCMC algorithm becomes challenging without relaxation, as it must ensure that the proposed candidates remain within the support of the target posterior distribution. To address this, we propose using the idea of Approximate Bayesian Computation \citep{abcwikinson,abcmarin}, which approximates the posterior when the likelihood function cannot be evaluated but can be sampled from. Let $K_\epsilon: \mathbb{R} \times \mathbb{R} \rightarrow \mathbb{R}$ be a similarity kernel function such that $K_\epsilon(x,y) = K_\epsilon(y,x)$, with $K_\epsilon(x,y) = d_\epsilon(|x-y|)$ for some non-decreasing function $d_\epsilon:\mathbb{R}_{\geq 0} \rightarrow \mathbb{R}_{\geq 0}$ such that $d_\epsilon(z) \rightarrow \delta_0(z)$ as $\epsilon \rightarrow 0$. $\epsilon$ is referred to as the ABC tolerance or the bandwidth. The approximation is associated to the rejection sampling algorithm that samples $\theta \sim p^\Theta(\theta)$, computes $\hat{r}_{s,a}=g_{s,a}(\theta)$ and retains $\theta$ with probability $K_\epsilon(\hat{r}_{s,a},\bar{r}_{s,a})/d_\epsilon(0)$. The resulting approximated posterior $\hat{p}_\epsilon$ has the form: 
\small\begin{equation}
    \pabc(\theta|\mathcal{D}^{\bar{r}}_\tau,\mathcal{D}_\tau^{\mathcal{S},\mathcal{A}}) \propto p^\Theta(\theta) \prod_{s,a \in \mathcal{D}_\tau^{\mathcal{S},\mathcal{A}}} \int p^*(\hat{r}_{s,a}|\theta,s,a) K_\epsilon(\hat{r}_{s,a},\bar{r}_{s,a}) \mathrm{d} \hat{r}_{s,a}
    = p^\Theta(\theta) \prod_{s,a \in \mathcal{D}_\tau^{\mathcal{S},\mathcal{A}}} K_\epsilon(g_{s,a}(\theta),\bar{r}_{s,a}).
    \label{eqn:abcpos}
\end{equation}\normalsize

The tolerance can be interpreted as representing our belief regarding the discrepancy between the model's best estimate and the observed data at the time of decision-making, while also acknowledging the model $Q_\theta$ may not fully capture $Q^*$ \citep{errorinterpreationkennedy,abcwikinson}.  Common kernels for ABC include the uniform kernel $K_\epsilon(x,y) = \frac{1}{2\epsilon} \mathbbm{1}(|x-y|<\epsilon)$ and the Gaussian kernel $K_\epsilon(x,y) = \mathcal{N}(y;x,\epsilon^2)$. Notice that when the Gaussian kernel is used, the implied posterior in Equation \ref{eqn:abcpos} takes the same form as if the likelihood function of $\bar{r}_{s,a}$ were Gaussian with variance $\epsilon$ centred at $g_{s,a}(\theta)$. This provides an interpretation for the choice of Gaussian likelihood in previous works described in Section \ref{sec:relatedwork} from the perspective of ABC. In cases where $Q^*$ does not lie within the parametric class, it can be shown that the posterior collapses to the maximiser of $\prod_{s,a \in \mathcal{D}_\tau^{\mathcal{S},\mathcal{A}}} K_\epsilon(g_{s,a}(\theta),\bar{r}_{s,a})$ for appropriate kernels as $\epsilon$ tends to zero. A justification for the Gaussian kernel is provided in Appendix \ref{apd:abcemptyset}.

\subsubsection{Stochastic rewards}

When $p^R$ is not deterministic, we only observe samples of the conditional random variable $R(s,a) := R_t|S_t=a,A_t=a$ through interactions with the environment. Hence, $\bar{r}_{s,a}$ is intractable without knowing the analytical form of $p^R$. In this setting, for any $s \in \mathcal{S}$, $a \in \mathcal{A}_s$, we assume an addictive zero-mean noise model for $R(s,a)$,

\begin{equation}
    R(s,a) = \mathbb{E}[R(s,a)] + \sigma(\phi)\eta_{s,a},
\end{equation} where $\eta_{s,a} \sim p^H(\bigcdot|s,a)$ is a zero-mean noise with known distribution $p^H$, and $\sigma:\Phi \rightarrow \mathbb{R}$ is a known scaling function dependent on the unknown parameter $\phi \in \Phi \subseteq \mathbb{R}^{d_\Phi}$. In other words, the reward is corrupted by a zero-mean noise known up to $\phi$.\newline

For a realisation of $\mathcal{M}$, we define the likelihood function $p^*$ on $R_t=r_t$ given $S_t=s_t$, $A_t=a_t$ by setting $\mathbb{E}[R_t|S_t=s_t,A_t=a_t]$ as $g_{s_t,a_t}(\theta)$, which gives
\begin{equation}
p^*(r_t|\theta,\phi,s_t,a_t) := \sigma(\phi)^{-1}p^H(\sigma(\phi)^{-1}(r_t - g_{s_t,a_t}(\theta))|s_t,a_t). \label{eqn:llhpstarnoisy}
\end{equation}

Then, up to time $t=\tau$, define $\mathcal{R}^{s,a}_\tau:=\{r_t|s_t=s,a_t=a \text{ for some } t \in \{0,\dots,\tau\}\}$, and $\mathcal{D}^{\mathcal{R}}_\tau:= \{\mathcal{R}^{s,a}_\tau|s,a \in \mathcal{D}_\tau^{\mathcal{S},\mathcal{A}}\}$. The overall likelihood function $p^*$ on $R_{0:\tau}=r_{0:\tau}$ has the form:
\small
\begin{equation}
    p^*(r_{0:\tau}|\theta,\phi,s_{0:\tau},a_{0:\tau}) := \prod_{t=0}^\tau p^*(r_t|\theta,\phi,s_t,a_t) = \prod_{s,a \in \mathcal{D}_\tau^{\mathcal{S},\mathcal{A}}} \prod_{r_{s,a} \in \mathcal{R}^{s,a}_\tau} p^*(r_{s,a}|\theta,\phi,s,a) =: p^*(\mathcal{D}^{\mathcal{R}}_\tau|\theta,\phi,\mathcal{D}_\tau^{\mathcal{S},\mathcal{A}}),
    \label{eqn:llhpstarnoisyfull}
\end{equation}
\normalsize

where the first equality is justified by the fact that given $S_t$ and $A_t$, $R_t$ is conditionally independent of the remaining variables.\newline

The overall posterior, therefore, has the form
\begin{equation}
p^*(\theta,\phi|r_{0:\tau},s_{0:\tau},a_{0:\tau}) = p^*(\theta,\phi|\mathcal{D}^{\mathcal{R}}_\tau,\mathcal{D}_\tau^{\mathcal{S},\mathcal{A}}) \propto p^\Theta(\theta) p^\Phi(\phi) p^*(\mathcal{D}^{\mathcal{R}}_\tau|\theta,\phi,\mathcal{D}_\tau^{\mathcal{S},\mathcal{A}}),
\end{equation}
with $p^\Phi$ the prior distribution on $\Phi$.\newline

Hence, the deterministic $p^R$ is a special case of the more general framework of Equation \ref{eqn:llhpstarnoisy} when $\eta_{s,a} \equiv 0$, and $\prod_{r_{s,a} \in \mathcal{R}^{s,a}_\tau} p^*(r_{s,a}|\theta,\phi,s,a)$ collapses to $\delta_{g_{s,a}(\theta)}(\bar{r}_{s,a})$.\newline

Table \ref{tab:pos_summary} summarises the distributions. The likelihood functions in both cases factorise with respect to $\mathcal{D}_\tau^{\mathcal{S},\mathcal{A}}$ and $\mathcal{D}_\tau^{\mathcal{R}}$ (or $\mathcal{D}_\tau^{\bar{r}}$). Thus, when there are multiple MDP realisations, the overall likelihood also factorises. We refer to this as episodic learning. For notational simplicity, as one episode realisation ends and the next begins, the time subscript $t$ in the datasets is incremented. This should not be confused with the definition of $Q^*$ and other relevant functions, which is the expected sum of rewards for each individual MDP realisation.\newline

\begin{table}[h!]
\centering
\begin{tabular}{l||l||l}
\hline \hline
 & Target posterior & Likelihood of $\mathcal{D}^\mathcal{R}_\tau$ (or $\mathcal{D}^{\bar{r}}_\tau$)\\
\hline \hline
Tractable $p^*$          &    $p^*(\theta,\phi|\mathcal{D}^{\mathcal{R}}_\tau,\mathcal{D}_\tau^{\mathcal{S},\mathcal{A}})$ & $\prod\limits_{s,a \in \mathcal{D}_\tau^{\mathcal{S},\mathcal{A}}} \prod\limits_{r_{s,a} \in \mathcal{R}^{s,a}_\tau} p^*(r_{s,a}|\theta,\phi,s,a)$     \\ \hline
Degenerate $p^*$ & $\pabc(\theta|\mathcal{D}^{\bar{r}}_\tau,\mathcal{D}_\tau^{\mathcal{S},\mathcal{A}})$ & $\prod\limits_{s,a \in \mathcal{D}_\tau^{\mathcal{S},\mathcal{A}}} K_\epsilon(g_{s,a}(\theta),\bar{r}_{s,a})$\\ \hline \hline
\end{tabular}
\caption{Target Posterior Distributions}
\label{tab:pos_summary}
\end{table}

\subsection{Discussions on the intractable expectation within the likelihood}
\label{sec:intractllh}
Notice that in the definition of $p^*$, $g_{s,a}(\theta)$ is dependent on an expectation over $p^S$, which is often intractable when $|\mathcal{S}|$ is infinite or when $p^S$ is unknown analytically. In this paper, we limit our contributions to the discussion of a few potential solutions to this problem for the stochastic reward case. The deterministic case follows similarly as a special case.\newline

Several factors need to be considered, including whether a computationally cheap state transition simulator is accessible and the size of the state space. When the state space is discrete and small, intractability arises from the unknown transition probabilities. Therefore, a straightforward solution is to model the transition probabilities in addition to $Q^*$, with the possibility of including additional simulations to improve the model. The derivation is in Appendix \ref{apd:tabmodelbased}. When the state space is large or continuous, however, computing the expectation may become computationally expensive or involve an intractable integral. This stems from the same underlying cause as the double sampling problem faced by MSBE-based methods previously discussed. In such a scenario, biased solutions include:\newline

\emph{Without Simulator:} Replace the random variable $S^\prime$ with the empirical next state $s^\prime$ following $(s,a)$, i.e. $\mathbb{E}[\max\limits_{a^\prime \in \mathcal{A}_{S_1}} Q_\theta(S_1,a^\prime)|S_0=s,A_0=a] \approx \max\limits_{a^\prime \in \mathcal{A}_{s^\prime}} Q_\theta(s^\prime,a^\prime)$ in $g_{s,a}(\theta)$ within $p^*(r_{s,a}|\theta,\phi,s,a)$ for any $r_{s,a} \in \mathcal{R}_t^{s,a}$, which is the same idea as MSTDE-based methods discussed above. This approach does not require any additional simulations but may incur a greater bias.\newline
 
\emph{With Simulator:} Approximate the intractable expectation using new Monte Carlo samples each time a likelihood evaluation is performed, assuming a simulator for $p^S$ is available without revealing subsequent rewards. In other words, let $Z_{s,a}^1,\dots,Z_{s,a}^m \sim p^S(\bigcdot|s,a)$ independently for some $m \in \mathbb{Z}_{\geq 1}$ and denote $Z_{s,a} = \{Z_{s,a}^i\}_{i=1}^m$, mutually independent across all $(s,a)$. The reward likelihood for any $(s,a)$ in Equation \ref{eqn:llhpstarnoisyfull} can be approximated using an alternative likelihood
$$\prod_{r_{s,a} \in \mathcal{R}^{s,a}_t} p^*(r_{s,a}|\theta,\phi,s,a) \approx \mathbb{E}\Bigg[\prod_{r_{s,a} \in \mathcal{R}_t^{s,a}} \sigma(\phi)^{-1} p^H (\sigma(\phi)^{-1}(r_{s,a} - \hat{g}_{s,a}^m(\theta,Z_{s,a}))|s,a) \Bigg],$$
where $\hat{g}_{s,a}^m(\theta,Z_{s,a}) := Q_\theta(s,a) - \frac{1}{m} \sum_{i=1}^m \max_{a^\prime \in \mathcal{A}_{Z_{s,a}^i}} Q_\theta(Z_{s,a}^i,a^\prime)$. In practice, it is further approximated by a Monte Carlo sample of $Z_{s,a}$. This is therefore an asymptotically unbiased approximation. Alternatively, independent Monte Carlo samples can be generated dynamically as needed for each $(s,a)$ as new data arrives and incorporated into the average in $\hat{g}_{s,a}^m$. This approach may increase memory consumption and introduce more complex dependencies within the chosen inference algorithm, but may also reduce bias over time. \newline

From here onwards, we do not distinguish between $\theta$ and $\phi$ but denote all unknown parameters modelled by a Bayesian prior as $\theta$. And for ease of notation, denote $\mathcal{D}_\tau := \{(s,a,r)| (s,a) \in \mathcal{D}^{\mathcal{S},\mathcal{A}}_\tau, r \in \mathcal{R}^{s,a}_\tau \}$ for tractable $p^*$ and $\mathcal{D}_\tau := \{(s,a,r)|(s,a) \in \mathcal{D}^{\mathcal{S},\mathcal{A}}_\tau, r=\bar{r}_{s,a} \}$ for their degenerate counterparts. Hence, the target posterior is denoted as $p(\theta|\mathcal{D}_\tau)$ for both cases.

\subsection{Theoretical form of posterior under tabular \texorpdfstring{$Q_\theta$}{Qθ} and Gaussian likelihood}
\label{sec:postabtheorem}
When $\mathcal{S}$ and $\mathcal{A}$ are finite spaces, the number of deterministic admissible policies is finite. For the special cases when $p^*$ is Gaussian with a fixed variance or $K_\epsilon$ is a Gaussian kernel, and $Q_\theta$ is tabular with a Gaussian prior, the posterior is tractable up to multivariate Gaussian integrals. Here is a definition of a tabular $Q_\theta$.\\

\begin{definition}
   $Q_\theta(s,a):\mathcal{S} \otimes \mathcal{A} \rightarrow \mathbb{R}$ has a tabular form if $d_{\Theta}=|(\mathcal{S} \setminus \mathcal{S}^g) \otimes \mathcal{A}|$, and 
    there exists an index function $\nu:\mathcal{S} \otimes \mathcal{A} \rightarrow \{1,\dots,d_{\Theta}+|\mathcal{S}^g|\}$ such that it is a bijection and $\nu((s^g,a^g)) \in \{d_{\Theta}+1,\dots,d_{\Theta}+|\mathcal{S}^g|\}$ for $s^g \in \mathcal{S}^g$, $a^g \in A_{s^g}$. Then, for $\theta \in \Theta$, define $\theta_j:= Q_\theta(\nu^{-1}(j))$ for $j \in \{1,\dots,d_{\Theta}\}$. Furthermore, with abuse of notation, we denote $\theta_k:= Q_\theta(\nu^{-1}(k)) \equiv 0$ for $k \in \{d_{\Theta}+1,\dots,d_{\Theta}+|\mathcal{S}^g|\}$.
    \label{def:tabular}
\end{definition}

Then, given a dataset consisting of collections of state, action, and reward, we can partition $\Theta$ in a way specific to the dataset such that the likelihood function, which contains the $\max$ operator, can be written as a sum of linear Gaussian likelihood within each partition. This allows us to compute the form of the posterior density and its cumulative distribution provided that we can evaluate Gaussian integrals numerically.\\

\begin{theorem}
\label{thm:psrl}
Let $\mathcal{D}=\{(s_i,a_i,r_i)\}_{i=1}^n$ be the dataset storing the unique transitions of a MDP, where $s_i \in \mathcal{S} \setminus \mathcal{S}^g$, $a_i \in \mathcal{A}_{s_i}$. Let $Q_\theta(s,a):\mathcal{S} \otimes \mathcal{A} \rightarrow \mathbb{R}$ be a tabular form with index bijection $\nu$, and define $$p(\theta,r_{1:n}|s_{1:n},a_{1:n}) := \prod_{i=1}^n \mathcal{N}\Big(r_i;\theta_{\nu(s_i,a_i)} - \sum_{s_i^\prime \in \mathcal{S}} p^S(s_i^\prime|s_i,a_i) \max_{a_i^\prime \in \mathcal{A}_{s_i^\prime}} \theta_{\nu(s_i^\prime,a_i^\prime)},\epsilon^2\Big) \prod_{j=1}^{d_{\Theta}} \mathcal{N}(\theta_j; 0, \sigma^2),$$
i.e., the marginal posterior is of the form $p(\theta|\mathcal{D})=p(\theta,r_{1:n}|s_{1:n},a_{1:n})/p(r_{1:n}|s_{1:n},a_{1:n})$.
Denote $\mathcal{S}^{\prime\mathcal{D}} = \bigcup_{i=1}^n \text{supp}(p^S(\bigcdot|s_i,a_i))$ and $\mathcal{A}^{\prime\mathcal{D}}=\prod_{s^\prime \in \mathcal{S}^{\prime\mathcal{D}}} \mathcal{A}_{s^\prime}$. Let $\ell^{\prime\mathcal{D}}=\{\ell:\mathcal{S}^{\prime\mathcal{D}} \rightarrow \mathcal{A}^{\prime\mathcal{D}}|\ell(s^\prime) \in \mathcal{A}_{s^\prime} \,\, \forall s^\prime \in \mathcal{S}^{\prime\mathcal{D}}\}$. Define $$E^\ell := \Theta \cap \bigcap_{\substack{s^\prime \in \mathcal{S}^{\prime\mathcal{D}} \\ s^\prime \notin \mathcal{S}^g}} \bigcap_{\substack{a^\prime \in \mathcal{A}_{s^\prime} \\ a^\prime \neq \ell(s^\prime)}}
\{\theta \in \Theta| \theta_{\nu(s^\prime,a^\prime)} - \theta_{\nu(s^\prime,\ell(s^\prime))} \leq 0\}.$$
Then, the marginal likelihood $$p(r_{1:n}|s_{1:n},a_{1:n})=\sum\limits_{\ell \in \ell^{\prime\mathcal{D}}}\mathcal{N}(r_{1:n};0,(\Gamma^{\ell})^{-1})\int_{E^\ell} \mathcal{N}(\theta;\mu^\ell_{\theta|r},\Sigma^\ell_{\theta|r})\mathrm{d}\theta,$$
where $\Gamma^{\ell} := (\sigma^2 B^\ell {B^\ell}^T + \epsilon^2 I_n)^{-1}$, $\mu^\ell_{\theta|r} := \sigma^2 {B^\ell}^T\Gamma^{\ell} r_{1:n}$, $\Sigma^\ell_{\theta|r}:= \sigma^2 I_{d_{\Theta}} - \sigma^4 {B^\ell}^T \Gamma^{\ell} B^\ell$, and $B^\ell \in \mathcal{R}^{n\times d_{\Theta}}$, $B^\ell_{i,j} = \mathbbm{1}(j=\nu(s_i,a_i)) - \sum_{s^\prime \in \mathcal{S}^{\prime\mathcal{D}}} p(s^\prime|s_i,a_i) \mathbbm{1}(j=\nu(s^\prime,\ell(s^\prime)))$, for $i \in \{1,\dots,n\}$, $j \in \{1,\dots,d_{\Theta}\}$.\newline

In particular, for any $E^* \subseteq \Theta$, 
$$p\Big(\theta \in E^* \Big|\mathcal{D}\Big) = (p(r_{1:n}|s_{1:n},a_{1:n}))^{-1}\Bigg(\sum\limits_{\ell \in \ell^{\prime\mathcal{D}}} \mathcal{N}(r_{1:n};0,(\Gamma^{\ell})^{-1}) \int_{E^*\cap E^\ell} \mathcal{N}(\theta;\mu^\ell_{\theta|r},\Sigma^\ell_{\theta|r})\mathrm{d}\theta\Bigg).$$
\end{theorem}

\begin{proof}
    See Appendix \ref{apd:proofpsrl}.
\end{proof}

The form of $p(\theta \in E^*|\mathcal{D})$ can be used to compute the probability that an action $a^* \in \mathcal{A}_{s^*}$ is optimal for a given state $s^* \in \mathcal{S} \setminus \mathcal{S}^g$ by setting $E^*=\bigcap_{\substack{a \in \mathcal{A}_{s^*} \\ a \neq a^*}} \{\theta \in \Theta| \theta_{\nu(s^*,a)} - \theta_{\nu(s^*,a^*)} \leq 0 \} $ (See Section \ref{sec:possamplingexploration}. Similarly, the probability that a deterministic policy $\mu:\mathcal{S} \rightarrow \mathcal{A}$ such that $\mu(s) \in \mathcal{A}_s$ for all $s \in \mathcal{S}$ is optimal can be calculated by setting $E^*=\bigcap_{s \in \mathcal{S}}\bigcap_{\substack{a \in \mathcal{A}_{s} \\ a \neq \mu(s)}} \{\theta \in \Theta| \theta_{\nu(s,a)} - \theta_{\nu(s,\mu(s))} \leq 0 \}$. Note that while this probability can be used to construct the exploration policy, this approach does not scale well with $|\mathcal{S}|$ and $|\mathcal{A}|$ and $d_{\Theta}$ because the summation is over the set $\ell^{\prime\mathcal{D}}$ and the Gaussian integrals are on $\Theta$. As we will discuss in Section \ref{sec:tspsrl}, an equivalent sampling algorithm is used in practice. Nevertheless, these exact probabilities can offer insights for theoretical analysis. Analysing the theoretical behaviour of the policy under our Bayesian framework is beyond the scope of this paper and is left for future studies.

\subsection{From Thompson sampling to posterior sampling for MDPs}
We now discuss and provide additional insights on how the posterior we defined can be used to construct an exploration policy for MDPs. To build intuition, we first introduce the multi-armed bandit problem and draw connections to Thompson sampling, a widely used strategy. We then highlight its connection to the commonly used posterior sampling strategy for MDPs in the literature.
\label{sec:tspsrl}

\subsubsection{Thompson sampling for multi-armed bandits}
Assume there are $K$ arms, or slot machines, labelled  $1,\dots,K$. Suppose a sequence of arms $(A_1,\dots,A_\tau)$ for an multi-armed bandit (MAB), where $A_t \in \{1,\dots,K\}$, are pulled at integer time steps $t \in \{1$,\dots,$\tau\}$. Pulling arm $k$ at time $t$ yields a real-valued random reward, $R_t\sim p(\bigcdot | k)$, where $p(\bigcdot | k)$ is the time-independent conditional pdf or pmf of the reward for pulling arm $k$. The expected reward is $\bar{r}_k \equiv \mathbb{E}[R_t|A_t=k]$, which is unknown. The goal is to pull the arm with the highest expected reward as many times as possible. This is an MAB problem. Thompson sampling (TS) is a Bayesian strategy that selects an action according to the posterior probability that that action is optimal.\newline

Specifically, let $p(\bigcdot|k,\theta)$ be the assumed pdf or pmf of the rewards for choosing action $k$, which is parameterised by $\theta \in \Theta$. Let $p^\Theta(\theta)$  be the prior probability density on $\Theta$. The posterior density of $\theta$ given the dataset $\mathcal{D}_\tau=\{a_t,r_t\}_{t=1}^\tau$ for $A_t=a_t$, $R_t=r_t$ is 
$$p^\Theta(\theta|\mathcal{D}_{\tau}) \propto p^\Theta(\theta) \prod_{t=1}^\tau p(r_t|a_t,\theta).$$ 
 At time $\tau +1$, the TS strategy is to play arm $k^\star$ according to the \emph{posterior probability} that the arm has the largest expected reward. That is, assuming that $\argmax_k \bar{r}_k(\theta)$ is $p(\bigcdot|\mathcal{D}_\tau)$-almost-surely unique,
\begin{equation} \mathbb{P}(A_{\tau+1}=k^\star|\mathcal{D}_{\tau}) := \mathbb{P}(\text{arm }k^\star \text{ is optimal}|\mathcal{D}_{\tau}):= \int_{\Theta} \mathbbm{1}(k^\star \in \argmax_k \bar{r}_k(\theta)) \; p(\theta|\mathcal{D}_{\tau})\mathrm{d}\theta,
\label{eq:ts_mab}
\end{equation}
where $\bar{r}_k(\theta) := \int r p(r|k,\theta) \mathrm{d} r$. When the argmax uniqueness assumption does not hold, such as when $\bar{r}_k(\theta)$ takes discrete values, a tie-breaking rule can be introduced and an arm can be redefined as optimal if it both maximises the expected reward and is selected by the rule in case of a tie. See Appendix \ref{apd:psrl_mab} for more discussions.\newline

Implementing this strategy is straightforward: Sample $\theta \sim p(\theta|\mathcal{D}_{\tau})$, followed by selecting $a_{\tau +1} \in \argmax_{k} \bar{r}_k(\theta)$ under a defined tie-breaking rule. Note that this has the same probability as if we sample according to the computed pmf in Equation \ref{eq:ts_mab}.\newline

TS naturally incorporates both \emph{exploration and exploitation}. Sampling an arm to play according to Equation \ref{eq:ts_mab} will tend to return the arm that the data suggests is the most rewarding.  Playing this arm corresponds to the agent \emph{exploiting} the``best guess'' decision. However, arms that have a low posterior probability of being the most rewarding will also occasionally be sampled. Playing these arms is exploratory since it may reveal new information that could lead to better decisions eventually.

\subsubsection{Posterior sampling for exploration for MDPs}
\label{sec:possamplingexploration}

 With the parametrisation $Q_\theta$ of $Q^*$, which defines optimality in MDPs, we can extrapolate TS to the MDP settings by sampling from the posterior distribution over the set of admissible optimal deterministic policies\footnote{A deterministic admissible policy $\mu:\mathcal{S} \rightarrow \mathcal{A}$ is optimal if and only if $\mu(s) \in \argmax_{a^\prime \in \mathcal{A}_s}Q^*(s,a^\prime) \,\, \forall s \in \mathcal{S}$: $(\Rightarrow)$ Assume $V^*(s) = V^\mu(s)$ for any $s \in \mathcal{S}$. Suppose $\exists \bar{s} \in \mathcal{S}\text{ such that } \mu(\bar{s}) \notin \argmax_{a^\prime \in \mathcal{A}_{\bar{s}}}Q^*(\bar{s},a^\prime)$. Let $\bar{a} \in \argmax_{a^\prime \in \mathcal{A}_{\bar{s}}}Q^*(\bar{s},a^\prime)$. Then, $V^\mu(\bar{s}) = Q^\mu(\bar{s},\mu(\bar{s})) \leq Q^*(\bar{s},\mu(\bar{s})) < Q^*(\bar{s},\bar{a}) = V^*(\bar{s})$, contradiction. $(\Leftarrow)$ $\mu(s) \in \argmax_{a^\prime \in \mathcal{A}_s}Q^*(s,a^\prime) \,\, \forall s \in \mathcal{S}$ implies that $Q^\mu(s,a) = \mathbb{E}[R(s,a) + Q^\mu(S_1,\mu(A_1)|S_0=s,A_0=a] = \mathbb{E}[R(s,a) + \max_{a^\prime \in \mathcal{A}_{S_1}} Q^\mu(S_1,a^\prime)|S_0=s,A_0=a]$ implies that BOEs are satisfied, i.e. $V^\mu(s) \equiv V^*(s)$.}, namely
\begin{equation}\mathbb{P}(\mu \text{ is an optimal deterministic policy}|\mathcal{D}_\tau)
:=p^\Theta(\{\theta| \forall s \in \mathcal{S}, \mu(s) \in \argmax_{a \in \mathcal{A}_s} Q_\theta(s,a)\}|\mathcal{D}_\tau).
\label{eq:ts_mdp}
\end{equation} assuming that $\argmax_{a \in \mathcal{A}_s} Q_\theta(s,a)$ is $p^\Theta(\bigcdot|\mathcal{D}_\tau)$-almost-surely unique (See Appendix \ref{apd:psrl_mdp} when this does not hold). This is analogous to picking an optimal arm in TS for MABs using the posterior probability in the sense of optimality in a Bayesian setting.\newline

To fully realise the cumulative rewards following an optimal policy sample from Equation \ref{eq:ts_mdp}, it can be beneficial to retain the same policy from the commencement of the task to its completion time for episodic problems, like SSP-type problems with finite (possibly random) terminating times. This leads to the phenomenon called \emph{deep exploration} \citep{osband2013moreefficient}, which is attributed to the execution until task completion of policies sampled from the posterior distribution over policy optimality in Equation \ref{eq:ts_mdp}, as illustrated in the benchmark Deep Sea problem \citep{randomisedvaluefunction} (See Section \ref{sec:experiment}). This idea was originally explained \citep{tsstren,osband2013moreefficient,randomisedvaluefunction} as a strategy in MDPs that not only take actions for immediate rewards but also consider a consistent set of actions that lead the agent towards regions with potential information gain, even if these regions are many steps away and offer low intermediate incentives along the way. In practice, while deploying a single policy may work for short episodes, for other applications with longer episodes or without a goal state, the current policy cannot incorporate new information during execution. A more general approach is to resample policies at times $\mathcal{T}=\{t_0,t_1,t_2,\dots\}$, where $0=t_0 < t_1 < t_2 < \dots$ and act greedily\footnote{A deterministic policy $\mu:\mathcal{S} \rightarrow \mathcal{A}$ is greedy if $\forall s \in \mathcal{S}$, $\mu(s) \in \argmax_{a \in \mathcal{A}_{s}} Q^*(s,a)$} to the most recent policy. \newline

In fact, deploying a policy from Equation \ref{eq:ts_mdp} for several timesteps is equivalent to the practical TS generalisation suggested by \citet{tsstren}, which samples $\theta \sim p^\Theta(\theta|\mathcal{D}_{t_i-1})$ at time $t_{i}$, constructs the greedy policy $\mu(s) = \argmax_{a \in \mathcal{A}_s} Q_\theta(s,a)$, and deploys it until time $t_{i+1}-1$ before resampling a new policy from an updated posterior with the latest data. A formal justification is provided in Appendix \ref{apd:psrl_mdp}. This practical implementation avoids the need to compute the optimal policy pmf, which is typically computationally expensive even if tractable (see Section \ref{sec:postabtheorem}. With suitable time intervals, this approach has been shown to be more data-efficient than resampling at every timestep and is widely adopted in subsequent works, and has been demonstrated to be both theoretically and empirically competitive in data efficiency to other exploration algorithms \citep{osband2013moreefficient,randomisedvaluefunction,bootstraposband,randomisedpriorwithensembleosband,psrlinfhorizon}, such as optimism-based methods \citep{ucrl2regret}.\newline

Note that while this strategy's practical implementation has become popular, its explicit and formal connection to the posterior distribution over optimal deterministic policies as defined in Equation \ref{eq:ts_mdp} has received limited emphasis in the existing literature to our knowledge, with mostly informal mentions \citep{osband2013moreefficient}. In this paper, we adopt this practical implementation, with the discussion above serving as a clarification of its connection to the original definition of TS in MABs, and offering additional insights into the tradeoff between regenerating a policy with an up-to-date posterior and maintaining a consistent policy under a slightly outdated posterior. In our framework, we interpret the posterior sampling exploration strategy as deploying a policy according to the posterior probability of it being optimal, conditioned on the BOEs for the encountered state-action pairs being (almost) satisfied and our prior belief.\newline

Finally, another advantage of maintaining the same policy for multiple steps is that it provides learning stability. Since the reward likelihood is centred at $Q_\theta(s,a) - \mathbb{E} [\max_{a^\prime \in \mathcal{A}_{S_1}}Q_\theta(S_{1},a^\prime)|S_0=s,A_0=a]$ for non-goal state-action pairs $s,a$, it does not inform the mean of $Q_\theta(s,a)$ until a goal state $s \in \mathcal{S}^g$ is encountered. For further discussions, see Section \ref{sec:example}. Therefore, for episodic learning with short episode lengths, the policy update interval can be conveniently set to the episode length.\newline

%% file: sections/simpleexample.tex
\section{Illustrative examples}
\label{sec:example}
We now present some MDP examples to illustrate the challenging landscape of the resulting posterior that the sampling algorithm must navigate under our framework; the prior choices to alleviate this complexity; and the necessity of reducing the ABC tolerance for deterministic MDPs.\newline

In this section, for deterministic rewards, we apply the Gaussian similarity kernel with tolerance $\epsilon$. The discussions are focused on the tabular modelling of $Q^*$ unless otherwise specified.

\begin{figure}[ht!]
\captionsetup{justification=raggedright}  

\centering
\begin{minipage}{0.45\textwidth} % Left diagram
\centering
\begin{tikzpicture}[
    >={Stealth[round]}, % Arrow tip
    state/.style={draw, circle, minimum size=0.5cm,font=\small}, % Style for state nodes
    edge label/.style={midway, font=\small}, % Style for edge labels
]

% Define nodes
\node[state] (s1) at (0, 0) {$s^1$};
\node[state] (s2) at (2, 0) {$s^2$};

% Add edges with labels
\draw[->] (s1) to[bend left] node[edge label, above] {$a^2,r^2$} (s2);
\draw[->] (s1) .. controls (-1.5, 1) and (-1.5, -1) .. node[edge label, left] {$a^1,r^1$} (s1);
\draw[->] (s2) .. controls (3.5, 1) and (3.5, -1) .. node[edge label, right] {$a^g,0$} (s2);

\end{tikzpicture}
\vspace{57pt}
    \caption{A deterministic 2-state MDP with a non-goal recurrent state. Each edge is labelled as (action, reward).}
    \label{fig:2dmdp}
\end{minipage}%
\begin{minipage}{0.45\textwidth} % Right diagram
\centering
\begin{tikzpicture}[
    >={Stealth[round]}, % Arrow tip
    state/.style={draw, circle, minimum size=0.5cm,font=\small}, % Style for state nodes
    edge label/.style={midway, font=\small}, % Style for edge labels
]

% Define nodes
\node[state] (s1) at (0, 0) {$s^1$};
\node[state] (s2) at (-1, -1) {$s^2$};
\node[state] (s3) at (1, -1) {$s^3$};
\node[state] (s4) at (-2, -2) {$s^4$};
\node[state] (s5) at (2, -2) {$s^5$};

\draw[->] (s1) -- (s2) node[edge label, midway, left, yshift=7pt] {$a^1, r^1$};
\draw[->] (s1) -- (s3) node[edge label, midway, right, yshift=7pt] {$a^2, r^2$};
\draw[->] (s2) -- (s4) node[edge label, midway, left, yshift=7pt] {$a^1,r^3$};
\draw[->] (s3) -- (s5) node[edge label, midway, right, yshift=7pt] {$a^2,r^4$};

% Add self-loops
\draw[->] (s4) .. controls (-2.75, -3.25) and (-1.25, -3.25) .. node[edge label, below] {$a^g,0$} (s4);
\draw[->] (s5) .. controls (2.75, -3.25) and (1.25, -3.25) .. node[edge label, below] {$a^g,0$} (s5);

\end{tikzpicture}
\caption{A deterministic 5-state MDP with tractable posterior. Each edge is labelled as (action, reward).}
        \label{fig:5dmdp}
\end{minipage}

\end{figure}

\subsection{The challenging posterior density landscape}
\label{sec:example_landscape}
The following deterministic MDP example will help illustrate key points in our discussions to follow.
\begin{example}
    Consider the 2-state deterministic MDP shown in Figure \ref{fig:2dmdp}, where $\mathcal{S}=\{s^1,s^2\}$, $\mathcal{A}_{s^1}=\{a^1,a^2\}$, and $s^2$ is the goal state with absorbing action $a^g$. At $s^1$, taking action $a^1$ transitions to $s^1$ and receives reward $r^1$, while taking action $a^2$ transitions to $s^2$ and receives reward $r^2$. Let $r^1=r^2=-1$. Denote the partial dataset as $\mathcal{D}_1 = \{(s^1,a^1,-1)\}$ and the complete dataset as $\mathcal{D}_2 = \{(s^1,a^1,-1),(s^1,a^2,-1)\}$. This implies $Q^*(s^1,a^1)=-2$, $Q^*(s^1,a^2)=-1$ and $Q^*(s^2,a^g)=0$. A tabular model for $Q^*$ therefore requires two scalar parameters, $\theta=(\theta_1,\theta_2)^T$, where $\theta_1$ models $Q^*(s^1,a^1)$ and $\theta_2$ models $Q^*(s^1,a^2)$.
    \label{ex:2dmdp}
\end{example}

\begin{figure}[t!]
    \centering
    \includegraphics{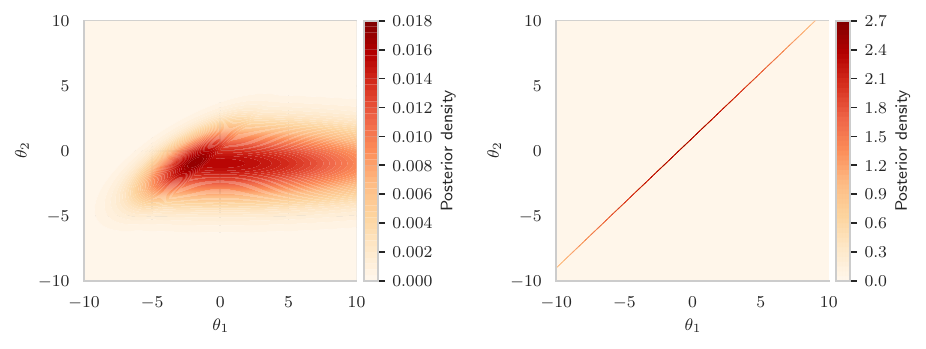}
    \caption{Left: Contour plot of the posterior of Example \ref{ex:2dmdp} with complete dataset $\mathcal{D}_2$, Gaussian prior with $\sigma=10$, tolerance $\epsilon=2$. Right: Contour plot of the posterior for Example \ref{ex:2dmdp} using the partial dataset $\mathcal{D}_1$, a zero-mean Gaussian prior with $\sigma = 10$, and tolerance $\epsilon = 0.01$.}
    \label{fig:2dcontour}
\end{figure}

In online learning, exploring only part of the MDP can result in an incomplete dataset and an overparameterised model, which in turn leads to an unidentifiable likelihood function. Consider the two-state deterministic MDP in Example \ref{ex:2dmdp}. The likelihood function is given by $L(\theta;\mathcal{D}_1)=\mathcal{N}(-1;\theta_1-\max(\theta_1,\theta_2),\epsilon^2)$, and the corresponding approximate posterior $\pabc(\theta|\mathcal{D}_1) \propto L(\theta;\mathcal{D}_1) p^\Theta(\theta)$ contracts towards the line $\theta_2=\theta_1 +1$ as $\epsilon \rightarrow 0$, as shown in Figure \ref{fig:2dcontour}. This is intuitive because $r^2$, the reward leading to the goal state, has not been revealed yet, leaving the magnitude of $Q^*$ at the preceding state-action pairs, $(s^1, a^1)$ and $(s^1,a^2)$ in this example, undetermined apart from being restricted by the prior distribution. This phenomenon extends similarly to MDPs with higher-dimensional state spaces: the posterior will contract towards the prior-constrained manifold that contains $Q^*$ and satisfies the subset of BOEs implied by a given incomplete dataset. Thus, when $\epsilon$ is sufficiently small, the posterior quantifies the uncertainty in estimating $Q^*$ by reflecting the prior belief confined to $Q_\theta$ that satisfy, or approximately satisfy under the $\epsilon$ tolerance, the subset of the BOEs evaluated at a specific time $t$.\newline

This contrasts with optimisation-based methods that aim to minimise MSBE, as discussed in Section \ref{sec:relatedwork}, where a local minimiser serves as a point estimate to represent the manifold. As demonstrated in the 2D example, any point estimate is likely a poor representation of the manifold, which may explain the limited success of MSBE-based optimisation methods. Furthermore, an estimator $Q_\theta$ with a small (or zero) empirical MSBE of a given dataset can still incur a large empirical mean-squared error to $Q^*$ and vice versa \citep{bellmanvsmse}. This motivates our use of Bayesian particle methods to represent the manifold, as detailed in Section \ref{sec:sampling} later.  When the dataset is incomplete, our goal is not to recover $Q^*$. Rather, we aim to identify probable optimal actions by examining the manifold's position, as described by the empirical distribution of the particles, within the parameter space derived from the BOEs.\newline

On the other hand, when the dataset is complete, i.e. all state-action pairs have been explored but the tolerance is not sufficiently small, the posterior contours of a tabular model are formed by merged hyper-ellipsoids connected by non-differentiable boundaries (hyperplanes) that partition the parameter space, as already illustrated in Theorem \ref{thm:psrl}. To see this, consider Example \ref{ex:2dmdp} again, but with the complete dataset $\mathcal{D}_2$. When the prior of $\theta$ is independent Gaussian and when $\epsilon$ is not sufficiently small, the resulting posterior contours, as illustrated in Figure \ref{fig:2dcontour}, are made of two ellipses fused along the line $\theta_1=\theta_2$ due to the max function in its likelihood. Each ellipse corresponds to the maximum being taken as either $\theta_1$ or $\theta_2$. This creates concave contour lines and non-differentiable boundaries and therefore poses challenges to sampling algorithms such as MCMC, as they may struggle to traverse the landscape efficiently. Nonetheless, when $\epsilon$ is small, the posterior contracts to $Q^*$ because all the state-action transition data have been collected.

\subsection{Non-goal recurrent states}
\label{sec:example_loop}
Sampling becomes particularly challenging for MDPs that include improper polices, even with a complete dataset. Specifically, we show below that, for any MDP with an improper policy, there exists a subset of $\Theta$, which may be unbounded if $\Theta$ is unbounded, such that the likelihood remains constant along a half line originating within the subset.\newline

A state $s^r \in \mathcal{S}$ is a non-goal recurrent state under a deterministic policy $\pi \in \Pi$ if $s^r \notin \mathcal{S}^g$ and $p^\pi(S_t=s^r \text{ for some } t \in \mathbb{Z}_{\geq 1}|S_0=s^r)=1$, and there exists an initial state $s_0^r \in \text{supp}(\rho)$ such that $p^\pi(S_t=s^r \text{\, for some \,} t \in \mathbb{Z}_{\geq 0}|S_0=s_0^r) > 0$. It is clear that if an improper policy exists and $\mathcal{S}$ is finite, an $s^r$ must exist, and the converse is also true. We now present the following result.\newline

\begin{theorem}
Assume the MDP has finite $\mathcal{S}$ and $\mathcal{A}$, satisfies Assumption \ref{ass:boeunique}, and has either deterministic rewards or independent Gaussian rewards. Consider a tabular model for $Q^*$ as in Definition \ref{def:tabular} with index function $\nu$ and $\Theta = \mathbb{R}^{d_\Theta}$, leading to the likelihood
$$ L(\theta|\mathcal{D}) = \prod_{(s,a,r) \in \mathcal{D}} \mathcal{N}(r;\theta_{\nu(s,a)} - \sum_{s^\prime \in \mathcal{S}} p^S(s^\prime|s,a) \max_{a^\prime \in \mathcal{A}_{s^\prime}} \theta_{\nu(s^\prime,a^\prime)}, \epsilon^2).$$
Furthermore, assume that the MDP contains a recurrent non-goal state $s^r$ corresponding to some improper deterministic policy, and let $u \in [0,1]^{d_\Theta}$ be such that $u_{\nu(s,a)}$ is the maximum probability leading to $s^r$ from a given state-action pair $(s,a)$. Then there exists a subset $\mathcal{O} \subseteq \mathbb{R}^{\mathrm{d}_\Theta}$, with non-zero Lebesgue measure, such that for any $\theta \in \mathcal{O}$, there exists a constant $c_\theta \leq 0$ satisfying $L(\theta|\mathcal{D}) = L(\theta+cu|\mathcal{D})$ for all $c > c_\theta$, $\epsilon>0$ and datasets $\mathcal{D}$.
\label{thm:loopthm}
\end{theorem}

\begin{proof}
    See Appendix \ref{apd:nondecaylkhproof}.
\end{proof}

To remark, this implies that for any $\theta$ in the subset $\mathcal{O}$ (as given formally in the proof), all points along the half line (or the full line if $c_\theta$ is unbounded) $\{\theta + cu|c> c_\theta\}$ yield the same error for the BOEs, and thus the likelihood is unchanged.\newline

Consequently, if $\mathcal{O}$ carries a considerable prior probability mass, such as under a Gaussian prior with large variance, the posterior density becomes elongated along certain directions of $\theta$, especially if $\epsilon$ is also large. To illustrate, Example \ref{ex:2dmdp} is a specific case of the MDPs involved in Theorem \ref{thm:loopthm}, with $s^0$ being the non-goal recurrent state under the deterministic transition dynamics. As shown in Figure \ref{fig:2dcontour}, the posterior density remains elongated towards large positive $\theta_1$, instead of contracting uniformly towards $Q^*$. The likelihood function satisfies
\begin{equation*}
    L(\theta;\mathcal{D}_2) = L((c+\theta_1,\theta_2)^T;\mathcal{D}_2) = \mathcal{N}(-1;0,\epsilon^2)\mathcal{N}(-1;\theta_2,\epsilon^2)\qquad \text{if } \min\{\theta_1,c+\theta_1\} \geq \theta_2.
\end{equation*}

To understand this in the context of the MDP, the approximated likelihood function can be interpreted as a Gaussian likelihood on noisy rewards, where $r^1\sim \mathcal{N}(\theta_1-\max(\theta_1,\theta_2), \epsilon^2)$. Consequently, when $\theta_1 \geq \theta_2$, which corresponds to a greedy policy that does not leave $s^1$ when started at $s^1$, the likelihood of observing $r^1$ becomes $\mathcal{N}(r^1;0,\epsilon^2)$. Thus, this region of the parameter space $\Theta$ violates Assumption \ref{ass:boeunique}, leading to the breakdown of the uniqueness assumption of the BOEs and the elongated contours. Note that this issue does not arise in MDPs without non-goal recurrent states (under any policies) when the dataset is complete and $Q_\theta$ is tabular. In such cases, the magnitude of $Q_\theta$ can be backpropagated from $Q_\theta(s^g,a^g)=0$ through the BOEs, all the way to that of the initial state-action pairs.\newline

For MDPs with non-goal recurrent states, and therefore improper policies, one may argue that the posterior distribution is not robust to the likelihood approximation via the Gaussian kernel. However, this stems from the fact that the support of the Gaussian prior contains $Q_\theta$ corresponding to improper policies that violate Assumption \ref{ass:boeunique}. A Gaussian prior can result in favourable theoretical properties, as demonstrated in Theorem \ref{thm:psrl}. A prior with a large variance is chosen for reasons such as lack of knowledge of the scale of $Q^*$, or to introduce optimism to facilitate exploration \citep{randomisedvaluefunction,provablyefficientpossampling}.  However, it fails to incorporate the prior knowledge that improper policies, if they exist, result in negative-infinite rewards into $Q_\theta$ -- the very motivation for using the BOEs to solve for $Q^*$. When $\epsilon$ is also too large to penalise such $\theta$, two issues arise: (1) the posterior mass shifts away from $Q^*$ towards regions associated with the recurrent states, introducing significant bias in the estimation of the optimal actions probabilities under the true model; (2) sampling becomes challenging due to the dispersed posterior. These issues apply similarly to stochastic rewards MDPs. However, as $\epsilon \rightarrow 0$ for deterministic rewards MDPs or as more rewards data are gathered for stochastic rewards MDPs, the likelihood of the $\theta$s within the region that violates the assumptions required for the BOEs uniqueness is small, and this pathological effect diminishes. \newline

One straightforward mitigating solution is by introducing the discount factor $\gamma$ as mentioned in Section \ref{sec:mdp}, which can serve as an approximation when Assumption \ref{ass:boeunique} holds. When $\gamma<1$, the uniqueness of the BOEs solutions is assured under milder conditions, avoiding the need of the improper policy Assumption \ref{ass:boeunique}, and the augmented $Q^*$ remains well-defined \citep{putermanrl}. This implicitly permits the existence of improper policies with non-negative-infinite rewards in the posterior modelling, while relying on the data to infer the augmented $Q^*$ through the BOEs. However, extreme values for components of $\theta$ may still arise with non-negligible density if $\gamma$ is close to $1$, as the implied incentive to take the improper policy is still high. Conversely, choosing a low $\gamma$ effectively alters the MDP formulation, resulting in a different optimal policy and deviating from the original $Q^*$.\newline

Alternatively, a Bayesian approach would be to elicit a prior to restricts the support of $Q_\theta$ so that the implied rewards of the MDP are consistent with Assumption \ref{ass:boeunique}. This can be done by leveraging the known structure of $p^S$ and $p^R$ to determine the range of values of $Q^*(s,a)$ consistent with the assumption, along with any additional knowledge not captured by the BOEs. The prior can then penalise or exclude $\theta$ incompatible with such preliminary information. This approach is problem-specific, and defining a unifying prior for all parametrisations of $Q^*$ and MDPs is a challenging problem and is not addressed here. We provide additional discussions for specific MDPs with tabular $Q_\theta$ in Appendix \ref{apd:prior}.

\subsection{The necessity of reducing \texorpdfstring{$\epsilon$}{ε}}
\label{sec:example_epsilon}

\begin{figure}[t!]
    \centering
    \includegraphics{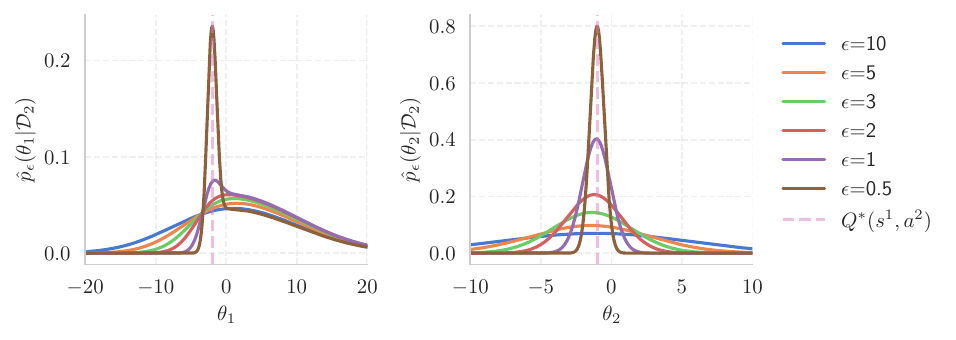}
    \caption{The marginal posterior of $\theta_1$ and $\theta_2$ respectively of Example \ref{ex:2dmdp} with complete dataset $\mathcal{D}_2$, Gaussian prior $\sigma=10$, and various tolerances.}
    \label{fig:illustrative_plot}
\end{figure}

 Although eliciting an appropriate prior is important, it may not always be feasible in practice. For deterministic transition MDPs, as shown in Figure \ref{fig:illustrative_plot}, the posterior mean in Example \ref{ex:2dmdp} deviates significantly from $Q^*$ when the tolerance is large. Thus, it becomes crucial to reduce the tolerance sufficiently so that the likelihood can dominate the uninformative or misspecified prior.\newline

In the MDP example below, in which the posterior of the optimal policies has an analytical form, we demonstrate that an insufficiently small tolerance can lead to incorrect decisions even when the entire MDP has been explored.

\begin{example}
    Consider the 5-state deterministic MDP shown in Figure \ref{fig:5dmdp}, where $\mathcal{S} = \{s^i\}_{i=1}^5$. At $s^1$, taking action $a^1$ leads to $s^2$ and receives reward $r^1$, whereas taking action $a^2$ leads to $s^3$ and receives reward $r^2$. At $s^2$, taking the only admissible action $a^1$ leads to $s^4$ and receives reward $r^3$. At $s^3$, taking the only admissible action $a^2$ leads to $s^5$ and receives reward $r^4$. Both $s^4$ and $s^5$ are absorbing states. A tabular model comprises 4 parameters $\theta = (\theta_1,\theta_2,\theta_3,\theta_4)^T$, which models $Q^*(s^1,a^1)$ as $\theta_1$, $Q^*(s^1,a^2)$ as $\theta_2$, $Q^*(s^2,a^1)$ as $\theta_3$, $Q^*(s^3,a^2)$ as $\theta_4$.
\end{example}

Suppose the MDP has been fully explored. As this problem does not involve the max function in the likelihood function, the posterior of $\theta$ is Gaussian with closed-form mean and variance when a Gaussian prior is used. This problem's cumulative reward depends solely on the decision made at the initial state $s^1$. The posterior probability of choosing $a^1$ over $a^2$ at $s^1$ is as follows:

\begin{lemma}
    Let $\mathcal{D}=\{(s^1,a^1,r^1),(s^1,a^2,r^2),(s^2,a^1,r^3),(s^3,a^2,r^4)\}$ and define the prior $p^\Theta(\theta)= \mathcal{N}(\theta;0,\sigma^2I)$. Then, the posterior probability of choosing the action $a^1$ over $a^2$ at $s^1$ is
    $$\pabc(\theta_1 > \theta_2 |\mathcal{D},\sigma) = \Phi\Bigg(\frac{kd-c}{\sigma\sqrt{2k(k+2)(k^2+3k+1)}}\Bigg),$$
    where $d = r^1-r^2$ and $c = r^2+r^4-r^1-r^3$, $k=\frac{\epsilon^2}{\sigma^2}$ and $\Phi:\mathbb{R} \rightarrow [0,1]$ is the $\mathcal{N}(0,1)$ cumulative distribution function.
    \label{lem:5dmdp}
\end{lemma}

\begin{proof}
    See Appendix \ref{apd:5dexampleproof}.
\end{proof}

Suppose $r^1 > r^2$ but $r^2+r^4> r^3 + r^1$, meaning that the action $a^1$ seems more favourable than $a^2$ initially, but $a^2$ leads to a higher cumulative reward. Due to the way the prior interacts with the approximated likelihood function, the result demonstrates that when the tolerance is not sufficiently small compared to the prior variance, i.e. $k = \frac{\epsilon^2}{\sigma^2} > \frac{r^2+r^4-r^1-r^3}{r^1-r^2} = \frac{c}{d}$, the probability of choosing the sub-optimal action $a^1$ exceeds $0.5$. Furthermore, for fixed $\sigma$, there exists rewards $r^1$,$r^2$,$r^3$,$r^4$ and $\epsilon$ such that this probability can be arbitrarily close to $1$. For any $\sigma > 0$ and rewards such that $c>0$, though, the probability converges to $0$ as $\epsilon$ approaches $0$, which corresponds to always selecting the optimal action $a^2$.\newline

%% file: sections/sampling.tex
\section{Sampling}
\label{sec:sampling}
To target the posterior of interest outlined in Table \ref{tab:pos_summary}, we consider two scenarios: (i) Offline learning - where we seek the posterior of $\theta$ after we have stopped collecting data at time $\tau$; (ii) Online learning - where we seek the intermediate posterior distributions of $\theta$ as data arrive sequentially. For the former, we use Hamiltonian Monte Carlo (HMC), an MCMC algorithm that targets the full posterior, and for the latter we use sequential Monte Carlo (SMC) to sample from a sequence of posterior distributions which can be used for constructing policies for exploration as described in Section \ref{sec:tspsrl}. In addition, this sequence of distributions also includes intermediate distributions that provide better algorithmic stability by appropriately interpolating between these target distributions.\newline

In the subsection below, we give a brief introduction to HMC and SMC. Readers who are familiar with these can jump to Section \ref{sec:offlinelearning} for offline learning or Section \ref{sec:onlinelearning} for online learning.

\subsection{Preliminaries for MCMC and SMC}

\subsubsection{MCMC}
We consider Hamiltonian Monte Carlo (HMC) as the Markov chain Monte Carlo (MCMC) method for problems that we are only interested in one posterior distribution or for mutations in sequential Monte Carlo (see Section \ref{sec:smc}). Let $p^\Theta$ be the target distribution supported on $\Theta$. MCMC samples from $p^\Theta$ by repeatedly applying a transition kernel to construct a Markov chain that is stationary and (under further conditions) ergodic for $p^\Theta$. An MCMC transition kernel acting on $\Theta$ is defined as $\kappa(\bigcdot,\bigcdot): \Theta \times \Theta \rightarrow \mathbb{R}_{\geq 0}$ such that $\kappa(\theta,\bigcdot) \in \mathscr{P}(\Theta)$ for any $\theta \in \Theta$. Hence, given $\theta \in \Theta$, $\theta$ is moved to a new position $\theta^\prime \in \Theta$ following the density $\kappa(\theta,\theta^\prime)$. Now, for any $\theta \in \Theta$, let $p \in \mathbb{R}^{d_\Theta}$ and $z=(\theta^T, p^T)^T$. Define the Hamiltonian function $H:\mathbb{R}^{2d_\Theta} \rightarrow \mathbb{R}$ such that $H(z) = H((\theta^T,p^T)^T) = -\log p^\Theta(\theta) + p^T C^{-1} p/2$, $C \in \mathbb{R}^{d_\Theta \times d_\Theta}$ a symmetric positive-definite matrix, known as the mass matrix. Let $\Psi_t^{C,p^\Theta}$ be the flow of the differential equation $\frac{\mathrm{d}\theta}{\mathrm{d}t} = C^{-1}p$, $\frac{\mathrm{d}p}{\mathrm{d}t} = \nabla_\theta \log p^\Theta(\theta)$, and define the distribution $p^Z$ with density $p^Z((\theta^T,p^T)^T) = k \exp(-H((\theta^T,p^T)^T))$ for normalising constant $k$, which has $p^\Theta$ as the marginal. If $(\theta^T,p^T)^T \sim p^Z$, $p^Z$ is stationary and time-reversible with respect to the Markov kernel $({\theta^\prime}^T, -{p^\prime}^T) = \Psi_t^{C,p^\Theta}((\theta^T,p^T)^T)$ for any $t$. In general, the flow does not have an analytical form and an approximated discretised solution is constructed using the leapfrog integrator, which has favourable properties such as volume preservation and time-reversibility. To ensure stationarity to $p^Z$, the Metropolis-Hastings correction is used. These properties of HMC then allow us to construct an ergodic MCMC transition kernel using $L$ leapfrog steps with step-size $\delta$ denoted as $\hat{\Psi}_{L,\delta}^{C,p^\Theta}$, targeting the marginal distribution $p^Z$\citep{hmcneal}. In other words, $({\theta^\prime}^T, {-p^\prime}^T)^T = \hat{\Psi}_{L,\delta}^{C,p^\Theta}((\theta^T,p^T)^T)$ is proposed and is accepted with probability $\min(1,\exp(H((\theta^T,p^T)^T)-H(({\theta^\prime}^T,{p^\prime}^T)^T)))$. See Appendix \ref{apd:massmatrix} for more discussions on the choice of $C$, and the algorithm is illustrated in Algorithm \ref{alg:hmc} in Appendix \ref{apd:algo}.

\subsubsection{SMC}
\label{sec:smc}
Define a sequence of distributions $p^\Theta_0,\dots,p^\Theta_J$ supported on $\Theta$. Sequential Monte Carlo (SMC) is a sampling algorithm that generates weighted particles to approximate $p^\Theta_j(\cdot)$ sequentially from $j=0$ to $j=J$ \citep{smc2006}. We say that weight-particle pairs $\{\omega^{j,(n)},\theta^{j,(n)}\}_{n=1}^N$ approximate $p_j^\Theta$ if its approximation has the form $\hat{p}_j^\Theta(\theta) = \sum_{n=1}^N \omega^{j,(n)} \delta_{\theta^{j,(n)}}(\theta)$, where $0 \leq \omega^{j,(n)} \leq 1$ such that $\sum_{n=1}^N \omega^{j,(n)} = 1$ and $\theta^{j,(n)} \in \Theta$. Usually, more than one of the intermediate distributions is of interest.\newline

At $j=0$, $\theta^{0,(n)} \sim p^\Theta_0$ for $n \in \{1,\dots,N\}$ is sampled for an SMC algorithm with $N$ particles. The weights $\{\omega^{0,(n)}\}_{n=1}^N$ are initialised as $\omega^{0,(n)}=N^{-1}$. Thus $p^\Theta_0(\theta) \approx \hat{p}^\Theta_0(\theta) = \sum_{n=1}^N \omega^{0,(n)}\delta_{\theta^{0,(n)}}(\theta)$. Given the approximation $\hat{p}^\Theta_j(\theta)$ of $p^\Theta_j(\theta)$, the weights are updated according to $\omega^{j+1,(n)}\propto \omega^{j,(n)}p^\Theta_{j+1}(\theta^{j,(n)})(p^\Theta_j(\theta^{j,(n)}))^{-1}$ such that $\sum_{n=1}^N \omega^{j+1,(n)} = 1$. The effective sample size (ESS), is then used to measure the degeneracy of the weights, which is defined as $$\text{ESS}(\{\tilde{\omega}^{j,(n)}\}_{n=1}^N) = \frac{(\sum_{n=1}^N (\tilde{\omega}^{j,(n)}))^2}{\sum_{n=1}^N (\tilde{\omega}^{j,(n)})^2},$$ and takes values between $1$ and $N$ for any unnormalised weights $\{\tilde{\omega}^{j,(n)}\}_{n=1}^N$, $\tilde{\omega}^{j,(n)} \geq 0$. As a rule of thumb, as ESS drops below $N/2$, the particles are resampled according to the probabilities $\{\omega^{j+1,(n)}\}_{n=1}^N$ using schemes such as multinomial resampling \citep{resamplingcomparison} and all weights $\{\omega^{j+1,(n)}\}_{n=1}^N$ are reset as $1/N$. Finally, the particles $\{\theta^{j,(n)}\}_{n=1}^N$, after the optional resampling step, are mutated via a Markov (MCMC) kernel $\kappa^{j+1}:\Theta \times \Theta \rightarrow \mathbb{R}_{\geq 0}$ that is $p^\Theta_{j+1}$-stationary to form a new set of particles $\{\theta^{j+1,(n)}\}_{n=1}^N$, and together with $\{\omega^{j+1,(n)}\}_{n=1}^N$ to define $\hat{p}^\Theta_{j+1}$. It can be shown under mild conditions that $\mathbb{E}_{\hat{p}^\Theta_J}[\psi(\theta)]$ converges to $\mathbb{E}_{p^\Theta_J}[\psi(\theta)]$ as $N \rightarrow \infty$ for multivariate function $\psi$ mapping from $\Theta$ \citep{cltsmc2004,smc2006}.\newline 

MCMC kernels often require tuning, and it is unlikely that a set of hyperparameters would work for all intermediate distributions targeted by SMC. Adaptive SMC algorithms introduce heuristics to utilise the particles and the MCMC performance from the previous time step to inform the hyperparameter choices for the current time step \citep{smchmctuning,smcfearnhead}. See Section \ref{sec:smc_mcmcmoves} for more discussions. The overall algorithm is illustrated in Algorithm \ref{alg:smc} in Appendix \ref{apd:algo}.

\subsection{Offline learning}
\label{sec:offlinelearning}
Given a dataset $\mathcal{D}_\tau$, we seek to obtain samples from the posterior distribution $p^*(\bigcdot|\mathcal{D}^{\mathcal{R}}_\tau,\mathcal{D}_\tau^{\mathcal{S},\mathcal{A}})$ for tractable $p^*$, tractable here refers to its likelihood  (see Table \ref{tab:pos_summary}), or $\pabc(\bigcdot|\mathcal{D}^{\bar{r}}_\tau,\mathcal{D}_\tau^{\mathcal{S},\mathcal{A}})$ for degenerate $p^*$. In both cases, the sample space is simply $\Theta$. Assuming that the unnormalised target posterior density is differentiable, except on a set which has zero measure \citep{hmcneal} (and the same holds for $\theta \mapsto Q_\theta(s,a)$ for any $s \in \mathcal{S}$, $a \in \mathcal{A}_s$), we can simply use HMC (potentially with parallel tempering \citep{paralleltemperinggeyer} if the target $\epsilon$ is small in $\pabc$). The gradient of $\theta \mapsto g_{s,a}(\theta)$ for a tabular $Q_\theta$ is given in Appendix \ref{apd:gradq}.

\subsection{Online learning}
\label{sec:onlinelearning}
To implement posterior sampling for exploration, we require samples from the sequence of posterior distributions $p^*(\bigcdot|\mathcal{D}_{t_i})$ at $t_i \in \mathcal{T}$ for tractable $p^*$, or from $\pabco{\epsilon_{t_i}}(\bigcdot|\mathcal{D}_{t_i})$ for a sequence of tolerances $\{\epsilon_{t_i}\}_{t_i \in \mathcal{T}}$ in the degenerate case. In this section, we propose a sequential Monte Carlo framework and focus on the degenerate case. The tractable case can be viewed as a special case of the degenerate case, where the ABC kernel function is replaced by the likelihood function and the tolerances are fixed and implicitly defined by the likelihood function.\newline

A well-tuned sampling algorithm should be able to target each approximated distribution with minimal tolerance. However, as tolerances are low, successive distributions become far apart (under a suitable probability distance metric), leading to SMC weights degeneracy and reduced MCMC effectiveness (see Section \ref{sec:example}) and more MCMC steps in the mutation kernel are required. Additionally, as data arrives, evaluating the full likelihood becomes more computationally expensive. In real-world RL applications, where decisions must be made within time constraints, it is crucial to balance the tolerances with computational cost. These are what make the degenerate case a difficult sampling problem, and the problem of ensuring MCMC effectiveness within each SMC update and relaxing the target distribution accordingly has largely been overlooked.\newline

To control the overall computational cost with respect to dataset size as decisions are made and data arrives, while keeping the overall discrepancy between its approximated posterior of $\pabco{\epsilon_{t_i}}(\bigcdot|\mathcal{D}_{t_i})$ and $p^*(\bigcdot|\mathcal{D}_{t_i})$ low and reasonable, we outline four aspects where the vanilla SMC algorithm needs modifications.

\begin{enumerate}
    \item Annealing scheme between successive target distributions.
    \item Selection of tolerances $\{\epsilon_{t_i}\}_{t_i \in \mathcal{T}}$.
    \item MCMC hyperparameter tuning for each SMC mutation step.
    \item Accounting for the growing cost of likelihood evaluations in MCMC mutations and SMC weight updates as the dataset expands.
\end{enumerate}

Note that we do not aim to solve each of them perfectly. Rather, we acknowledge the current gaps in the current SMC literature in solving this challenging online Bayesian RL problem. Where possible, we provide potential solutions, and any unresolved issues are left for future work.

\subsubsection{Constructing intermediate distributions}
\label{sec:annealscheme}
Firstly, we utilise annealing distributions \citep{chopin2002,smc2006} to bridge the successive distributions. Simplifying the notation slightly, suppose we aim to update $\pabco{\epsilon}(\theta|\mathcal{D})$ to $\pabco{\epsilon^\prime}(\theta|\mathcal{D}^\prime)$ after collecting new data, where $\mathcal{D} \subseteq \mathcal{D}^\prime$ and the tolerance changes from $\epsilon$ to $\epsilon^\prime >0$. Denote the new data as $\tilde{\mathcal{D}} := \mathcal{D}^\prime \setminus \mathcal{D}$. At annealing step $i$, we introduce intermediate distributions where $\mathcal{D}$ has tolerance $\epsilon_i$ and $\tilde{\mathcal{D}}$ has tolerance $\tilde{\epsilon}_i$, denoted as $$p_i^\Theta(\theta) := \pabco{\epsilon_i,\tilde{\epsilon}_i}(\theta|\mathcal{D},\tilde{\mathcal{D}}) \propto \pabco{\epsilon_i}(\theta|\mathcal{D})\prod_{s,a \in \tilde{\mathcal{D}}^{\mathcal{S},\mathcal{A}}} K_{\tilde{\epsilon}_i}(g_{s, a}(\theta), \bar{r}_{s,a}),$$
and transition from $p_{i-1}^\Theta(\theta)$ to $p_{i}^\Theta(\theta)$. Since in each SMC iteration, weights are updated before mutation and ESS is computed using the weights only, the new tolerances $(\epsilon_{i}, \tilde{\epsilon}_{i})$ can be selected so that they are closer to a target tolerance (a smaller tolerance) and the corresponding new ESS is a fraction $\alpha<1$ of the previous ESS at ($\epsilon_{i-1}, \tilde{\epsilon}_{i-1}$) \citep{abcsmc2011,abcsmcessproof} (see Algorithm \ref{alg:smcessscheme} for pseudocode). This controls the discrepancy between successive distributions. Each weight update is then followed by resampling if the ESS is falls below $N/2$, and MCMC mutation steps are implemented regardless of resampling.\newline

An example of a straightforward annealing scheme for $\epsilon > \epsilon^\prime$ is to use the ESS criterion to first assign an initial tolerance $\tilde{\epsilon}_1$ to $\tilde{\mathcal{D}}$, followed by gradually reducing $\tilde{\epsilon}_1$ to $\epsilon$ before uniformly lowering $\epsilon$ to $\epsilon^\prime$. That is, $(i)$ find $\tilde{\epsilon}_1 > \tilde{\epsilon}_2 > \dots > \tilde{\epsilon}_k=\epsilon$ and set $\epsilon_1=\epsilon_2=\dots=\epsilon_k=\epsilon$; $(ii)$ find $\epsilon_{k+1}> \epsilon_{k+2} > \dots > \epsilon_{k+\ell}=\epsilon^\prime$ and set $\tilde{\epsilon}_{k+i} = \epsilon_{k+i}$ for $i \in \{1,\dots,\ell\}$. Hence, if $K_\epsilon$ is a Gaussian kernel, the above annealing scheme can be interpreted as data annealing \citep{chopin2002} implemented through likelihood tempering \citep{abcsmc2011}.\newline

Note that under this scheme, the next tolerance $(\epsilon_{i}, \tilde{\epsilon}_{i})$ given $(\epsilon_{i-1}, \tilde{\epsilon}_{i-1})$ can always be found using the ESS criterion, unless the target tolerance ($\epsilon$ or $\epsilon^\prime$) can be reached with a smaller ESS reduction than $\alpha$. See Appendix \ref{apd:ess} for discussions. In this paper, we omit the discussions on the selection of tolerances when more than one tolerances satisfy the ESS criteria. Instead, we simply use bisection, a commonly used algorithm in the literature (see e.g. \citep{abcsmc2011,smchmctuning}) as a simple algorithm to find one such solution. The pseudocode is presented in Algorithm \ref{alg:smcpseudo2}, excluding the lines marked with an asterisk $(^*)$, and a more detailed version can be found in Algorithm \ref{alg:smcdegenerateupdatedetermined}. The SMC weight update ratios are presented in Table \ref{tab:smcweightupdate}.

\subsubsection{Choices of the tolerances}
\label{sec:tolerances}
Using the notation in \ref{sec:annealscheme} and given a dataset, a strategy to assign a tolerance $\epsilon^\prime$ for the posterior given that dataset, rather than assigning it arbitrarily, is essential for scenarios such as saving computational cost, deciding when to collect new data, ensuring $\tilde{\epsilon}_1$ and $\epsilon$ are not too far apart as more data becomes available. As tolerances decrease, particle rejuvenation (mutation) carried out by MCMC becomes less effective at mitigating weight degeneracy in SMC under a fixed computational budget. Eventually, the accuracy of the weighted particle approximation to the target (approximated) distribution deteriorates more than that of the likelihood approximation. Therefore, it is important to align the tolerances with the effectiveness of the MCMC sampler.\newline

Our primary aim is to devise an approach to ensure that a given tolerance level is reached only if the MCMC mutation kernel remains effective under a fixed number of MCMC moves. Hence, to maintain ``acceptable'' MCMC mixing, we propose to conduct a baseline sanity check, e.g. the Gelman-Rubin-related diagnostic \citep{gelman-rubbin}, to assess how far each particle has moved relative to the initial spread of the particles (See Appendix \ref{apd:mcmceffectiveness} for more discussion of potential metrics). When poor MCMC performance is flagged, the likelihood function is relaxed by gradually increasing the tolerance and rechecking until MCMC effectiveness is restored, so that each step results in the new ESS to be reduced by at most a factor of $\alpha$ while capped at a higher target tolerance (e.g. doubling the original tolerance). Specifically, a simple modification to the scheme in Section \ref{sec:annealscheme} is that if ineffective MCMC occurs when $\tilde{\epsilon}_i > \epsilon_i$, further reduction of $\tilde{\epsilon}_i$ is no longer attempted but $\epsilon_i$ is raised gradually until either MCMC is flagged effective again or if it matches $\tilde{\epsilon}_i$. On the other hand, if $\tilde{\epsilon}_i = \epsilon_i$, the common tolerance is gradually uniformly increased until MCMC performance improves. \newline

In addition, provided the MCMC remains effective as indicated by the baseline check, we propose to stop decreasing the tolerances and advance to the next target distribution when the incremental gain in the approximation accuracy from lowering the tolerances becomes too small. To do this, for a given dataset, we propose to simply compute the empirical expected squared Bellman error of a given set of weighted particles $\{\omega^{(n)},\theta^{(n)}\}_{n=1}^N$, which is defined as 
\begin{equation}
    \sum_{n=1}^N \omega^{(n)}  \sum_{s,a,r \in \mathcal{D}^\prime} \big(Q_{\theta^{(n)}}(s,a) - (r + \mathbb{E} [\max_{a^\prime \in \mathcal{A}_{S^\prime}} Q_{\theta^{(n)}}(S^\prime,a^\prime)])\big)^2
    \label{eq:bellmandefinition}
\end{equation}
for a given dataset $\mathcal{D}^\prime$. This error should be $0$ should there be infinitely many independent particles distributed according to $\hat{p}_\epsilon (\cdot |\mathcal{D}^\prime)$ and there exists a solution in $\Theta$ to the BOEs (for deterministic rewards MDPs). Then, a simple modification to the scheme in Section \ref{sec:annealscheme} is that when the entire dataset $\mathcal{D}^\prime$ shares a common tolerance $\epsilon_i=\tilde{\epsilon}_i$, further tolerance reduction is halted if no significant error improvement is observed over several consecutive steps. Since both ineffective MCMC mixing and likelihood approximation contribute to the empirical Bellman error, in practice, this check serves as both a measure of how well $\hat{p}_\epsilon$ approximates $p^*$ and an additional check to monitor MCMC mixing. \newline

Hence, the main distinction of this method from that in the literature \citep{abcsmc2011} is the harmonisation of the tolerance for the new and old dataset and the reversal of the
tolerance reduction until MCMC is deemed effective again. The above-mentioned modifications are marked with an asterisk $(^*)$ in Algorithm \ref{alg:smcpseudo2}.\newline

As of the writing of this paper, we have not identified any studies that specifically address the problem of selecting the tolerances for a sequence of relaxed degenerate posterior distributions corresponding to a growing dataset within the SMC framework for RL problems. Some existing works that discuss the stopping criteria for decreasing the tolerances when the dataset is fixed include \citet{abcsmc2011}, which proposed keeping the MCMC acceptance rate above a threshold. However, since the acceptance rate can be increased by reducing the step-size, we found this approach to be less reliable. Another approach proposed by \citet{umberto2021smcadaptive} suggested decreasing the tolerances until the estimated supremum of the ratio of two consecutive SMC targeted densities is close to $1$.

\begin{table}[ht]
    \centering
    \begin{tabular}{|c|c|c|c|}
    \hline
        Stage & $p_i^\Theta(\theta) $ & $ p_{i+1}^\Theta(\theta) $ & $p_{i+1}^\Theta(\theta) (p_i^\Theta(\theta))^{-1} $ \\ \hline
        I & $ \pabc(\theta|\mathcal{D})$& $\pabco{\epsilon,\tilde{\epsilon}_1}(\theta|\mathcal{D},\tilde{\mathcal{D}})$  & $\propto \prod_{(s,a,r) \in \tilde{\mathcal{D}}} K_\epsilon(g_{s,a}(\theta),r)$\\ \hline
        II & $ \pabco{\epsilon,\tilde{\epsilon}_i}(\theta|\mathcal{D},\tilde{\mathcal{D}})$ & $ \pabco{\epsilon,\tilde{\epsilon}_{i+1}}(\theta|\mathcal{D},\tilde{\mathcal{D}})$ & $\propto \prod_{(s,a,r) \in \tilde{\mathcal{D}}} K_{\tilde{\epsilon}_{i+1}}(g_{s,a}(\theta),r) (K_{\tilde{\epsilon}_i}(g_{s,a}(\theta),r))^{-1}$\\ \hline
        III/IVb & $ \pabco{\epsilon_i}(\theta|\mathcal{D}^\prime)$& $ \pabco{\epsilon_{i+1}}(\theta|\mathcal{D}^\prime)$ & $\propto \prod_{(s,a,r) \in \mathcal{D}^\prime} K_{\epsilon_{i+1}}(g_{s,a}(\theta),r) (K_{\epsilon_i}(g_{s,a}(\theta),r))^{-1}$\\ \hline
        IVa & $\pabco{\epsilon_i,\tilde{\epsilon}}(\theta|\mathcal{D},\tilde{\mathcal{D}})$ &  $\pabco{\epsilon_{i+1},\tilde{\epsilon}}(\theta|\mathcal{D},\tilde{\mathcal{D}})$ & $\propto \prod_{(s,a,r) \in \mathcal{D}} K_{\epsilon_{i+1}}(g_{s,a}(\theta),r) (K_{\epsilon_i}(g_{s,a}(\theta),r))^{-1}$\\ \hline
    \end{tabular}
    \caption{The density ratio for SMC weight updates for Algorithm \ref{alg:smcpseudo2} and Algorithm \ref{alg:smcdegenerateupdatedetermined}.}
    \label{tab:smcweightupdate}
\end{table}

\begin{tcolorbox}
\begin{myalgorithm}[Pseudocode for updating $\pabco{\epsilon}(\theta|\mathcal{D})$ to $\pabco{\epsilon^\prime}(\theta|\mathcal{D}^\prime)$ with adaptive tolerance choices]
    \label{alg:smcpseudo2}
    \textbf{\begin{itemize}[series=heading, leftmargin=0cm]
        \item[] Initialise
    \end{itemize}}
    \begin{modenumerate}[series=innerlist,label=\arabic*., leftmargin=0.5cm]
    \item Let $\pabco{\epsilon}(\theta|\mathcal{D})$ be approximated by weight-particle pairs $W_0=\{\omega^{0,(n)},\theta^{0,(n)}\}_{n=1}^N$. Set $\epsilon_0 \leftarrow \epsilon$, $\tilde{\epsilon}_0 \leftarrow \infty$. Set $i \leftarrow 0$
    \moditem{$^*$} Set MCMC effectiveness counter $c_m \gets 0$ with maximum $N_m$, Bellman error counter $c_b \gets 0$, $\epsilon^\prime \gets 0$ with maximum $N_b$.
    \end{modenumerate}
    \begin{itemize}[resume=heading, leftmargin=0cm]
    \item[] \textbf{Introduce new data $\tilde{\mathcal{D}} = \mathcal{D}^\prime \setminus \mathcal{D}$ and perform likelihood tempering (Reduce tolerances)}
    \end{itemize}

    \begin{modenumerate}[resume=innerlist,label=\arabic*., leftmargin=0.5cm]
    \item Set $i \leftarrow i + 1$. \label{alg:smc2:steprepeat1}
    \item 
    \underline{Stage I}: \textit{Find tolerance for new data $\tilde{\mathcal{D}}$.}\newline
    If $\tilde{\epsilon}_{i-1} =\infty$, find and set $\tilde{\epsilon}_i$ such that $\epsilon \leq \tilde{\epsilon}_i < \infty$ using ESS rule. Set $\epsilon_i \leftarrow \epsilon_{i-1}$.\newline

    \underline{Stage II:} \textit{Reduce tolerance for new data $\tilde{\mathcal{D}}$ to match tolerance of old data $\mathcal{D}$.}\newline
    If $\tilde{\epsilon}_{i-1} > \epsilon_{i-1}$, find and set $\tilde{\epsilon}_i$ such that $\epsilon \leq \tilde{\epsilon}_i < \tilde{\epsilon}_{i-1}$ using ESS rule. Set $\epsilon_i \leftarrow \epsilon_{i-1}$.\newline

    \underline{Stage III:} \textit{Reduce tolerance for all data $\mathcal{D}^\prime$.}\\
    If $\tilde{\epsilon}_{i-1}=\epsilon_{i-1}$, find and set $\tilde{\epsilon}_i=\epsilon_i$ such that $0 < \epsilon_{i} = \tilde{\epsilon}_i < \epsilon_{i-1}$ using ESS rule.

    \item Use SMC to update $W_i$ from $W_{i-1}$ to approximate $\pabco{\epsilon_i,\tilde{\epsilon}_i}(\bigcdot|\mathcal{D}, \tilde{\mathcal{D}})$.
    \moditem{$^*$} If MCMC remains effective, reset $c_m \gets 0$, else increment $c_m \gets c_m + 1$. If $c_m=N_m$, jump to \ref{alg:smc2:increasetemp}.
    \moditem{$^*$} If $\epsilon_{i-1}=\tilde{\epsilon}_{i-1}$ and if Bellman error did not improve, $c_b \gets c_b +1$, else reset $c_b \gets 0$. If $c_b = N_b$, exit.
    \item If $\epsilon_i \neq \epsilon^\prime$ or $\tilde{\epsilon}_i \neq \epsilon^\prime$, jump to \ref{alg:smc2:steprepeat1}, otherwise, exit.
    \end{modenumerate}

    \begin{itemize}[resume=heading, leftmargin=0cm]
    \item[] \textbf{Increase tolerances until MCMC is effective again}
    \end{itemize}

    \begin{modenumerate}[resume=innerlist,label=\arabic*., leftmargin=0.5cm]
    \moditem{$^*$} Set $i \gets i+1$. \label{alg:smc2:increasetemp}
    \moditem{$^*$}  \underline{Stage IVa}: \textit{Increase tolerance for old data $\mathcal{D}$. (capped at twice the original tolerance)}\newline
    If $\tilde{\epsilon}_{i-1} > \epsilon_{i-1}$, find and set $\epsilon_i$ such that $\epsilon_{i-1} < \epsilon_i \leq \min(2\epsilon_{i-1}, \tilde{\epsilon}_{i-1})$ using ESS rule. Set $\tilde{\epsilon}_{i} \gets \tilde{\epsilon}_{i-1}$.

    \underline{Stage IVb}: \textit{Increase tolerances for all data $\mathcal{D}^\prime$. (capped at twice the original tolerance)}\newline
    If $\tilde{\epsilon}_{i-1} = \epsilon_{i-1}$, find and set $\epsilon_i=\tilde{\epsilon}_i$ such that $\epsilon_{i-1} < \epsilon_{i} = \tilde{\epsilon}_i \leq 2\epsilon_{i-1}$ using ESS rule.
    
    \moditem{$^*$} Use SMC to update $W_i$ from $W_{i-1}$ to approximate $\pabco{\epsilon_i, \tilde{\epsilon}_i}(\bigcdot|\mathcal{D}, \tilde{\mathcal{D}})$.
    
    \moditem{$^*$} If MCMC remains ineffective, jump to \ref{alg:smc2:increasetemp}. If MCMC becomes effective and $\tilde{\epsilon}_{i} > \epsilon_{i}$, jump to \ref{alg:smc2:steprepeat1}, otherwise, exit.
    \end{modenumerate}

\end{myalgorithm}
\end{tcolorbox}

\subsubsection{MCMC hyperparameter tuning}
\label{sec:smc_mcmcmoves}

Each intermediate distribution may require different MCMC kernel hyperparameters to ensure efficient sampling across the parameter space. To tune these hyperparameters, we primarily adopt the strategy from \citet{smchmctuning}, which leverages existing particles and trial runs based on Effective Squared Jumping Distance (ESJD) \citep{esjd,smcfearnhead}. By assuming that consecutive distributions require similar hyperparameters, the adaptation process relies on the performance of the previous run targeting an earlier distribution. Such hyperparameters include the HMC mass matrix, step-size, and the number of Leapfrog integration steps. However, if tolerances are to be raised due to MCMC ineffectiveness, the hyperparameters from the previous run may not serve as a reliable guide for adaptation. In such cases, an additional trial run or a more carefully chosen searching space (lower and upper bounds for hyperparameter searching) may be required for the adaptation step before checking MCMC effectiveness on the current distribution. Pseudocodes are detailed in Algorithm \ref{alg:smcbase} and \ref{alg:smchmcadptmove} in Appendix \ref{apd:algo}. For further discussion on the number of MCMC iterations, see Appendix \ref{apd:mcmceffectiveness}.

\subsubsection{Potential solutions to the unbounded and expanding computational cost of likelihood evaluation as data arrives}
\label{sec:smc_compcost}

Firstly, in the annealing scheme introduced earlier, the number of intermediate tempering distributions between two successive posteriors may vary depending on the new dataset and tolerance levels. In online learning tasks that have a deadline for decision-making, a natural extension is to introduce state-action-dependent tolerances so that the tolerances can be dropped in batches. This allows the sampling algorithm to pause, derive a policy from the most recent weighted particles to interact with the MDP, and resume at the next posterior update time. The resulting posterior then reflects our belief regarding the uncertainty of the unknown parameter given the data while accounting for the computational constraints. We leave the practical implementation of this approach for future work.\newline

Next, the computational cost of each MCMC step scales linearly with dataset size due to the likelihood evaluations, causing each successive posterior update to take longer as data arrives. A solution is to use Stochastic Gradient MCMC (SGMCMC) \citep{sgld_teh,sgmcmc_complete_recipe,tdsmc} instead, which leverages the conditional independent likelihood to estimate the log-likelihood gradient unbiasedly via data sub-sampling with a dataset-independent cost and removes the accept-reject step. SGMCMC can be viewed as discretising a Stochastic Differential Equation (SDE) with a stationary distribution that matches the target distribution, using a decaying step-size.\newline 

Finally, the overall computational cost of the SMC algorithm is proportional to the cumulative number of tolerances applied to each data instance across the full dataset. Thus, the cost is high if the tolerances for each data instance are updated frequently. To alleviate the cost, one could discretise a pre-defined tolerance interval and trade off computational cost for increased memory usage by storing the likelihood values at a set of tolerances. Alternatively, each weight update could be approximated by data sub-sampling \citep{tdsmc}, though it would generally introduce bias.\newline

Theoretical and empirical evaluation of these inference approximation methods is beyond the scope of this paper and is left for future work.

\subsubsection{Final algorithm}
The overall algorithm for interacting with the MDP is presented in Algorithm \ref{alg:onlinedegen1}.

\begin{tcolorbox}
\begin{myalgorithm}[Pseudocode for online learning with SMC]
\label{alg:onlinedegen1}
    \textbf{\begin{itemize}[series=heading, leftmargin=0cm]
        \item[] Initialise
    \end{itemize}}
\begin{enumerate}[series=innerlist,label=\arabic*., leftmargin=0.5cm]
    \item Sample $\theta^{(n)} \sim p^\Theta(\bigcdot)$ for $n=1,\dots,N$ and set $W=\{N^{-1},\theta^{(n)}\}$ as the weight-particle pairs to approximate $p^\Theta(\bigcdot)$.
    \item Initialise episode counter $e \gets 0$, time counter $t \gets 0$, episode time counter $t_e \gets 0$, and dataset $\mathcal{D}_{t} \gets \emptyset$. Set maximum number of episodes $E$.
\end{enumerate}
    \textbf{\begin{itemize}[resume=heading, leftmargin=0cm]
    \item[] One episode
    \end{itemize}}
\begin{enumerate}[resume=innerlist,label=\arabic*., leftmargin=0.5cm]           \item Sample $\theta^{t_e}$ from weight-particle pairs $W$.\label{alg:online:initialstate}
    \item Sample initial-state $s_t \in \rho(\bigcdot)$. 
    \item Select action $a_t \in \argmax_{a \in \mathcal{A}_{s}} Q_{\theta^{t_e}}(s_t,a)$. \label{alg:online:samplestate}
    \item Observe $r_t \sim p^R(\bigcdot|s_t,a_t)$, $s_{t+1
} \sim p^S(\bigcdot|s_t,a_t)$.
    \item Set $\mathcal{D}_{t+1} \gets \mathcal{D}_t \cup \{(s_t,a_t,r_t,s_{t+1})\}$ for deterministic rewards MDPs; Append ${(s_t,a_t,r_t,s_{t+1})}$ to $\mathcal{D}_t$ to obtain $\mathcal{D}_{t+1}$ for stochastic rewards MDPs.
    \item If $s_{t+1} \notin \mathcal{S}^g$, increment $t \gets t+1$ and jump to \ref{alg:online:samplestate}.
\end{enumerate}
    \textbf{\begin{itemize}[resume=heading,leftmargin=0cm]
    \item[] Update posterior
    \end{itemize}}
\begin{enumerate}[resume=innerlist,label=\arabic*., leftmargin=0.5cm]
    \item Increment $e \gets e+1$, $t \gets t+1$ and set $t_e \gets t$. Select new tolerance $\epsilon_{t_e}$ (omitted for stochastic rewards MDPs) and update weight-particle pairs $W$ to approximate $\pabco{\epsilon_{t_e}}(\bigcdot|\mathcal{D}_{t_e})$ using Algorithm \ref{alg:smcpseudo2}.
    \item If $e< E$, jump to \ref{alg:online:initialstate}.
\end{enumerate}
\end{myalgorithm}
\end{tcolorbox}

%% file: sections/experiment.tex
\section{Experimental study}
\label{sec:experiment}

In this section, we demonstrate the efficacy of our algorithm by comparing it with other exploration reinforcement learning algorithms. Our goal is to highlight the ability of our method to learn effectively, even in challenging environments.\newline

We apply Algorithm \ref{alg:onlinedegen1} with the sampling algorithm described in Algorithm \ref{alg:smcpseudo2} to a well-established benchmark environment known as Deep Sea \citep{randomisedvaluefunction}, which is designed to evaluate exploration strategies in reinforcement learning. This environment presents a challenging setting that requires effective exploration due to its deceptive reward structure and increasing complexity with greater depth values in the Deep Sea environment.\newline

We have chosen this particular benchmark problem because it facilitates direct comparisons with various state-of-the-art reinforcement learning methods. Specifically, we compare our approach to posterior sampling for reinforcement learning (PSRL) \citep{psrlinfhorizon}, a Bayesian exploration strategy that balances exploration and exploitation by sampling from a posterior distribution over models, and bootstrapped deep Q-Networks (BDQN) \citep{bootstraposband}, which leverages randomised value functions to encourage exploration. By evaluating our algorithm in this environment, we aim to assess its performance relative to these well-established methods and provide insights into its exploration efficiency.

\subsection{Experiment setup}
\begin{figure}[ht]
    \centering
    \includegraphics[width=0.5\linewidth]{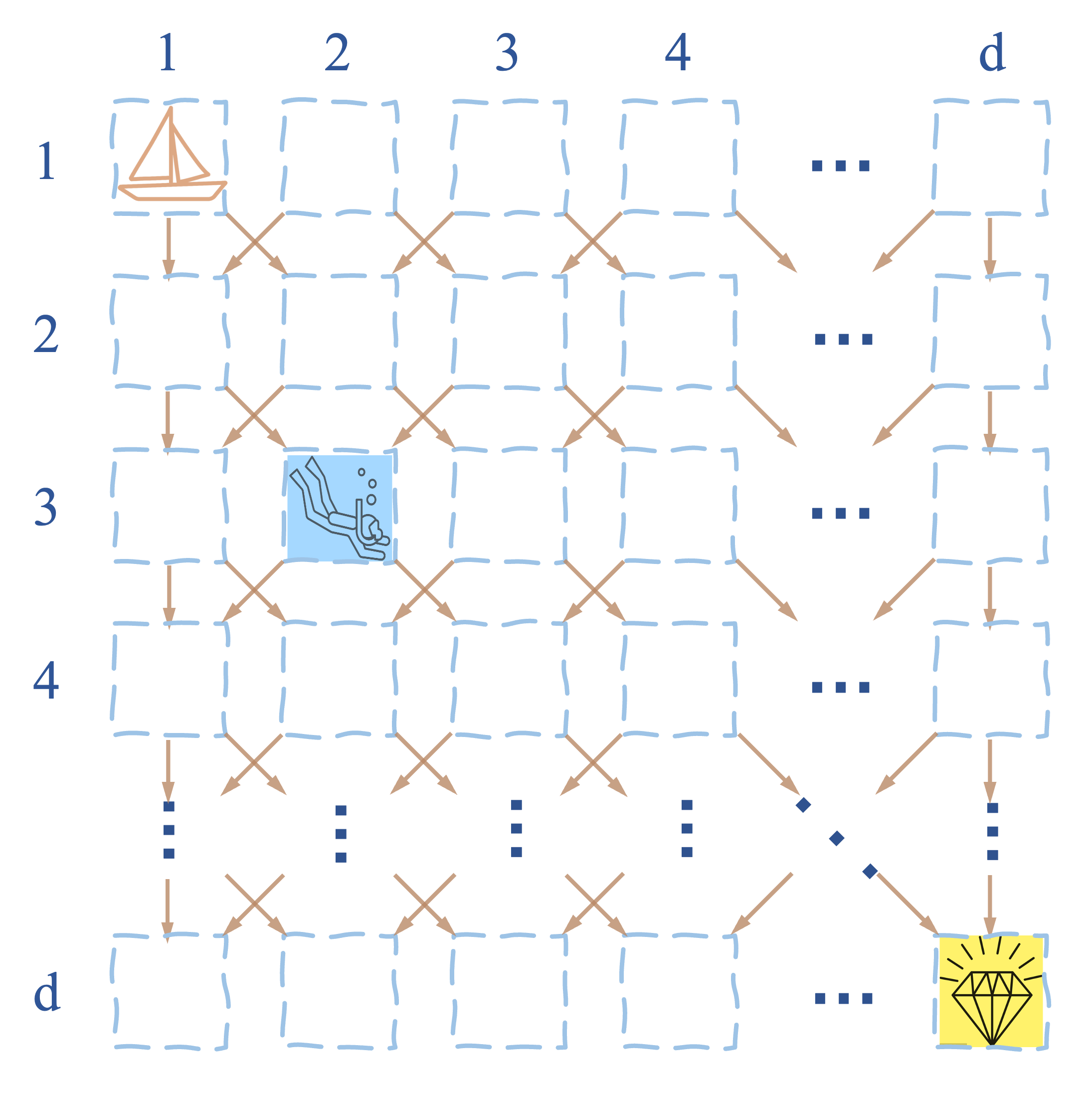}
    \caption{Deep Sea illustration \citep{randomisedvaluefunction}.}
    \label{fig:deep-sea}
\end{figure}

The Deep Sea problem, which is illustrated in Figure {\ref{fig:deep-sea}}, is a finite-horizon MDP with deterministic transitions. The state space is comprised of cells as illustrated in the figure. The diver descends through these cells until they reach the bottom level, and the episode terminates. The goal states are thus all the cells in the bottom level. The action space is $\mathcal{A} = \{0, 1\}$, and action 0 causes the diver to descend one level and then to the adjacent cell on the right. Action 1 moves the diver to the cell one level below on the left. The right or left part of the movement is contingent on not exiting the domain. The state is a two-dimensional vector with the row and column number of the cell. The order of the size of the state space is $d^2$, where $d$ denotes depth. Every episode has the same termination time $T=d-1$. However, in our examples, we accentuate the exploration challenge by always initialising the diver at the top-left cell, or grid state $(0, 0)$, where the diver left her boat, so that $|\mathcal{S}| = d(d+1)/2$. The bottom far right cell contains the treasure, and the reward earned for visiting this cell is $R=1$. Furthermore, moving down and left earns a reward of $R_d = {1}/{100d}$, while down and right a reward $R_d=-{1}/{100d}$. All the rewards in this example are deterministic. As we explain next, these choices for rewards, negative for going right and positive for moving left, are again to accentuate exploration difficulty.\newline

For this deterministic MDP, the optimal cumulative reward for one episode is to keep taking action $0$ at all steps to be able to visit the bottom cell of the far right and collect the extra reward for the treasure, which will offset the penalties for moving right en route. Note, though, that knowledge of this best policy is not exploited in our data-driven Bayesian learning framework in Algorithm \ref{alg:onlinedegen1}. In the figures below, due to the relaxation of the posterior, we refer to our method as approximate Bayesian reinforcement learning (ABRL). This algorithm interacts with the environment, over episodes, to gather data for learning the optimal policy. Thus, until our diver actually visits the bottom right cell and collects the data point (or extra reward) for the treasure, the likelihood will not inform the optimality of visiting this bottom right cell beyond what the chosen prior distribution expresses.  A poorly conceived exploration and exploitation strategy may thus always guide the diver away from visiting this cell with the treasure, since positive rewards are earned for all collected data points for moves left, as opposed to negative rewards for moving right. For example, purely dithering strategies like epsilon-greedy have been shown to take an exponentially long number of episodes, in depth $d$, to explore and reach the goal \citep{suttonbartobook}.\newline

We represent each $Q^{\ast}(s,a)$ with its own scalar parameter as in Definition \ref{def:tabular} with the vector $\theta \in \mathbb{R}^{d_\Theta}$. The prior distribution is $p^\Theta(\theta) = \mathcal{N}(\theta;0,4^2I)$ and the posterior being learned is the challenging degenerate distribution $p^*$ in Table \ref{tab:pos_summary}. The SMC algorithm uses $N = 20$ particles for the following Deep Sea depths, $d=1, 2, ..., 15$; and $N = 100$ particles for depths $d= 16, ..., 40$. 
We use the Gelman-Rubin diagnostic \citep{gelman-rubbin} for flagging tolerance values at which the MCMC is no longer effective, which will then trigger the revision of the tolerance as detailed in Section \ref{sec:tolerances}.

\subsection{Experiment result}
The following metrics are used to demonstrate the performance of our method as a function of the problem size $d$. The first metric is the cumulative regret over $E$ episodes,
\begin{equation}
    \text{Regret}(d, E) := \sum_{e=1}^E (V^{\ast}(s^e_0) - \sum_{t=0}^{d-2} r(s^e_t, a^e_t)),
    \label{eq:regret}
\end{equation}
where $(s_0^e,a_0^e,\dots,s_{d-2}^e,a_{d-2}^e,s_{d-1}^e)$ is the observed sequence of the state action pairs in episode $e$ (with $s_{d-1}^e \in \mathcal{S}^g$ being a goal state). The second metric is the learning time \citep{randomisedvaluefunction}, which is the first episode where the average regret drops below 0.5. This is to investigate the performance of our algorithm as the problem size grows:
\begin{equation}
    \text{Learning time}(d) := \min\left\{ E > 1 \ \bigg| \ \frac{\text{Regret}(d, E)}{E} \leq 0.5 \right\}.
    \label{eq:learning_time}
\end{equation}
For both metrics, our method is evaluated against PSRL and BDQN. \citet{bootstraposband, randomisedvaluefunction} showed that the PSRL is the strongest performer among a selection of competing methods and thus serves as a suitable strong baseline in our numerical evaluations.

\begin{figure}[H]
    \centering
    \includegraphics{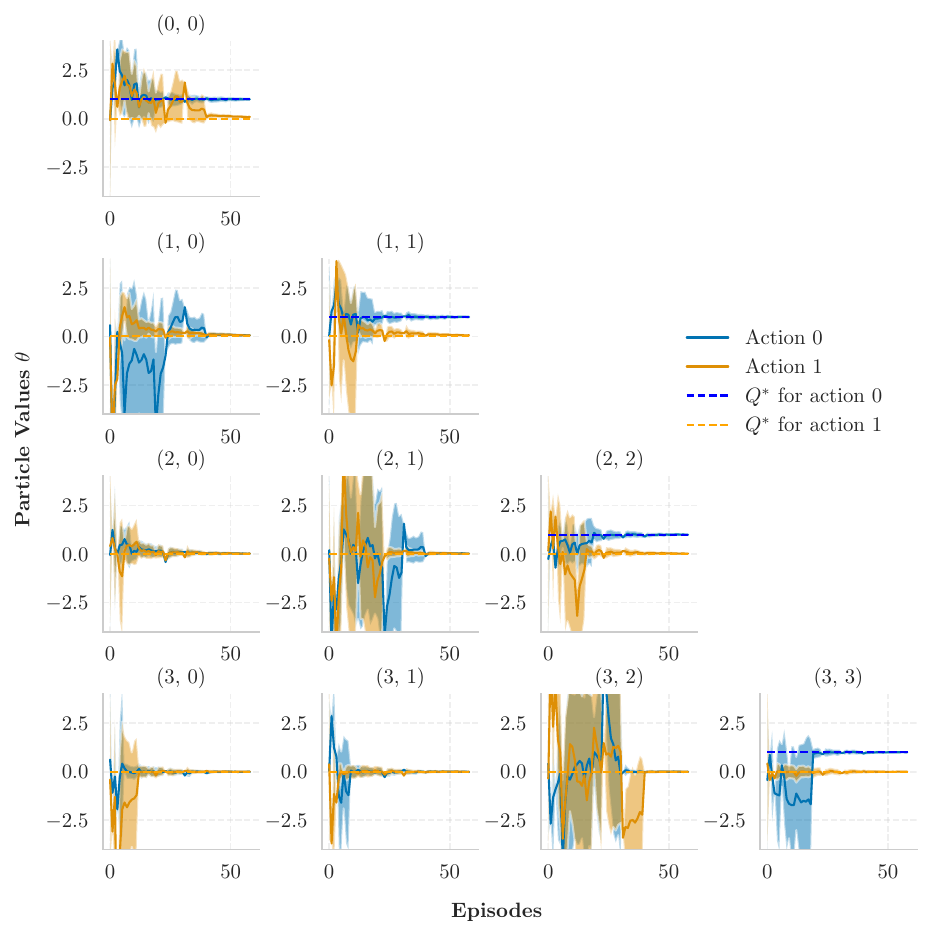}
    \caption{Posterior samples for Algorithm \ref{alg:smcpseudo2}  for a $5\times 5$ Deap Sea problem. Each sub-figure illustrates the posterior samples $Q_\theta(s,a)$ for $a = 0$ and $a = 1$  for a particular cell $s$. Cell arrangement is shown in Figure \ref{fig:deep-sea}. The spread of the data is characterised by the empirical standard deviation, represented by the shaded region around the mean (solid lines).}
    \label{fig:sample_distribution}
\end{figure}

Figure \ref{fig:sample_distribution} presents the particle positions that approximate the posterior distributions defined in stage III in Table \ref{tab:smcweightupdate} for every episode, plotted against the episode number, for a $d=5$ Deep Sea problem with 10 particles. The solution to the BOEs is learned for each state-action pair via its own $\theta$-component. For example, the top-most cell shows the posterior samples of $Q^*$ for $s=(0,0)$, and $a=0$ and $a=1$. It can be clearly seen that the posterior changes as the episodes progress and these changes are step-like due to the appearance of data for state-action pairs not previously observed;  either data for the specific $(s, a)$ is collected or for any other pairs that $(s, a)$  communicates with. Equally, we see uncertainty in the posterior for larger $d$ being manifest in states with smaller $d$ that communicate with it. Finally, learning is quicker for states closer to the goal states, which are states at depth $d=5$.\newline

\begin{figure}[H]
    \centering
    \includegraphics{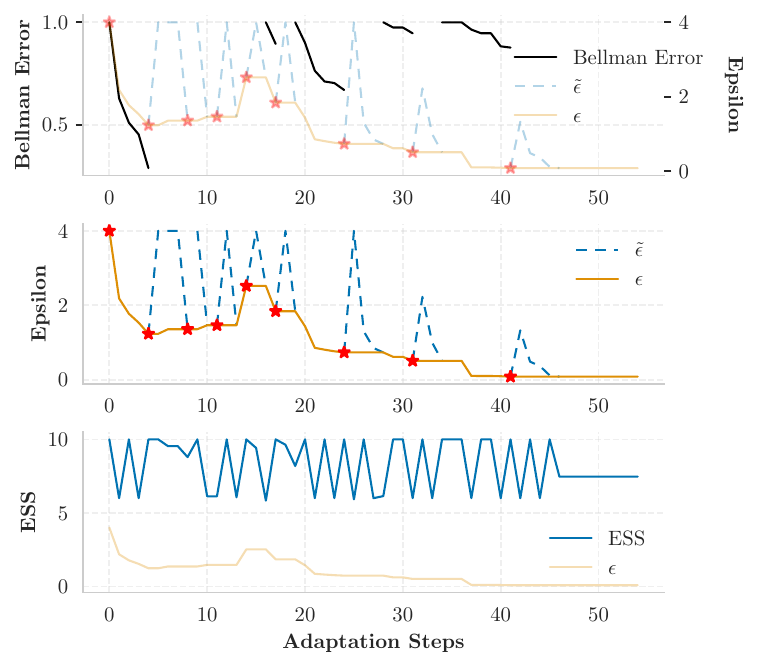}
    \caption{Training details with the Bellman error and the ESS in parallel with the tolerance $\epsilon$ in the solid line and $\tilde{\epsilon}$ in dotted lines; the final tolerance for each adaptation period is marked with $\star$.}
    \label{fig:epsilon}
\end{figure}

Figure \ref{fig:epsilon} illustrates a single realisation of the sequential adaptation process of Algorithm \ref{alg:smcpseudo2} for the $5\times5$ Deep Sea example by presenting several signals concurrently. The progression of the tolerances $(\epsilon,\tilde{\epsilon})$ is illustrated in the middle panel. As explained in Section \ref{sec:annealscheme}, new data are assimilated into the posterior with its own tolerance $\tilde{\epsilon}$ (dotted lines), which may initially be large in order to maintain a target ESS level; recall the old data's tolerance value is $\epsilon$ (the solid line). This is then followed by the gradual reduction of $\tilde{\epsilon}$ to arrive at a common tolerance value for the posterior with all the data (new and old). The common tolerance may be larger than that of the previous posterior tolerance if MCMC is ineffective for the enlarged dataset at the previous posterior's tolerance value. Otherwise, if the empirical Bellman error, as defined in Equation \ref{eq:bellmandefinition}, decreases, the common tolerance is adjusted downward. The $\epsilon$ at the end of each adaptation period is marked with $\star$, after which new data is introduced; in top figure, the recording of the normalised Bellman error is triggered immediately after the tolerance $\tilde{\epsilon}$ of the new data matches the tolerance $\epsilon$ of the previous data and stops when no further improvement is identified. The bottom figure displays the ESS dropping due to the change of tolerance values, and increasing due to the resampling step in SMC.\newline

\begin{figure}[ht]
    \centering
    \begin{subfigure}[t]{0.49\textwidth}
        \centering
        \includegraphics{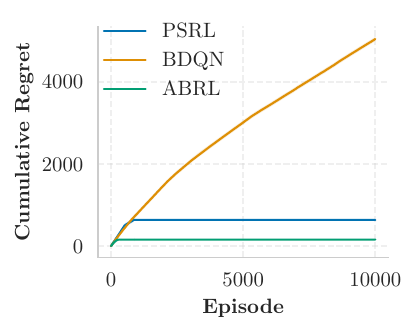}
        \caption{Comparison of the cumulative regret for PSRL, BDQN, and ABRL methods in a $10 \times 10$ Deep Sea environment.}
        \label{fig:regret_compare}
    \end{subfigure}
    \hfill
    \begin{subfigure}[t]{0.49\textwidth}
        \centering
        \includegraphics{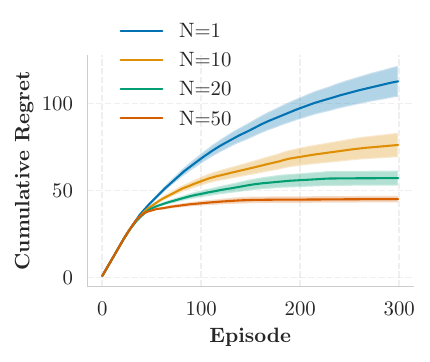}
        \caption{Comparison of the cumulative regret with different number of particles in a $10 \times 10$ Deep Sea environment.}
    \label{fig:compare_n}
    \end{subfigure}
    \caption{}
\end{figure}

 Figure \ref{fig:regret_compare} compares the cumulative regret ABRL, PSRL\footnote{We have implemented accelerated PSRL following \citet{randomisedvaluefunction} to be run in the deterministic environment where each of the observations $(s, a, r, s^\prime)$ was repeated 10 times in the dataset.} and boostrapped DQN (BDQN), averaged over 3 random runs. Both BDQN and ABRL use $20$ ensembles/particles.  It can be seen that ABRL achieves a much lower cumulative regret level and converges after roughly 200 episodes. A smaller cumulative regret implies more rapid exploration, while its levelling off implies convergence at the best policy. (If the best policy was not found, the regret would increase.) BDQN's regret shoots off while PSRL's regret flatlines. PSRL appears to initially struggle to explore toward the treasure.\newline

Figure \ref{fig:compare_n} shows the effect of the number of particles on the cumulative regret, averaged over 50 random runs. It is clear that the regret levels off earlier with more particles, and the cumulative regret has less variability over different runs.\newline

\begin{figure}[H]
    \centering
    \includegraphics{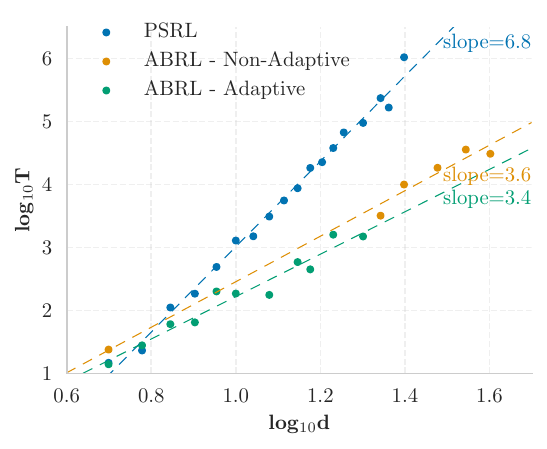}
    \caption{Learning time with increasing problem sizes (log scale)}
    \label{fig:LearningT}
\end{figure}

Finally, we contrast the performance of PSRL and ABRL with increasing Deep Sea problem sizes in Figure \ref{fig:LearningT}. Each experiment is repeated with $5$ random seeds. To save on run time, we use an adaptive algorithm illustrated in Algorithm \ref{alg:smcpseudo2} with fewer particles for smaller problem sizes, and turn off part of the adaptation (referred to as Non-Adaptive in the figure) for larger problems (d $\geq$ 20), and increase the number of particles to $100$. In high-dimensional settings, we identify through pilot runs a sufficiently small target tolerance $\epsilon_{target}$ for all data instances that enables the MCMC chains to explore effectively. During training, we progressively reduce $\epsilon$ until it reaches $\epsilon_{target}$. As shown in the figure, PSRL struggled in the Deep Sea environment, with its learning times scale as $\mathcal{O}(d^{6.8})$. In contrast, ABRL scales as $\mathcal{O}(d^{3.4})$ with the adaptive algorithm, and $\mathcal{O}(d^{3.6})$ with the non-adaptive algorithm.

%% file: sections/conclude.tex
\section{Discussion and conclusion}
\label{sec:conclude}

We introduced a Bayesian framework to construct posterior beliefs for the optimal action-value function $Q^*$. It uses a parametric model for $Q^\ast$, and through a suitably defined likelihood, it sequentially enforces the BOEs as data become available.  The updated belief progressively constrains the prior to the manifold on which $Q^*$ lies. Compared to other MSTDE-based approaches, our framework does not rely on unrealistic or ad-hoc assumptions, thus offering improved interpretability and, potentially, better theoretical properties, such as asymptotic posterior consistency.\newline

Likelihood functions for $Q^{\ast}$ are introduced for both deterministic and stochastic rewards, where the latter assumes additive zero-mean noise. However, to facilitate computational inference for deterministic rewards, we introduced a controllable relaxation of the likelihood. The Bellman operator becomes intractable for a large or infinite state space, which is akin to the so-called double-sampling problem in existing MSBE-based methods. To avoid this problem, a Monte Carlo approximation was proposed, but its implementation is left for future work. A potential research direction is to extend our framework using a generalised Bayes approach \citep{generalisedbayes} where a loss-based likelihood replaces the standard likelihood to handle the intractable Bellman expectations, or more complex reward distributions, while maintaining coherent Bayesian inference over $Q^*$.\newline

We have shown that posterior sampling for exploration--in the literature often implemented by adhering to a greedy policy derived from a posterior sample of $Q^{\ast}$--is equivalent to sampling from the posterior distribution over the set of optimal deterministic policies. This establishes a direct link to Thompson sampling in multi-arm bandit problems, where an optimal MDP policy is analogous to an optimal arm. Although exploration via posterior sampling will (under suitable conditions) lead to the optimal policy as the posterior contracts, it does not guarantee optimality of the exploration; for example, as measured by the cumulated regret. In future work, it would be interesting to establish theoretical regret bounds for our approach.\newline

For a tabular paramerisation of $Q^{\ast}$, we have highlighted the lack of identifiability of the likelihood for $Q^{\ast}$ when the MDP is only partially explored. This issue is exacerbated when $Q^*$ is undiscounted and the MDP admits improper policies that result in non-goal recurrent states. In which case, as shown in Theorem \ref{thm:loopthm}, lack of identifiability can persist over an unbounded region of the parameter space, even when all state-action pairs have been visited at least once. Thus, it is important that the prior is chosen to explicitly exclude improper policies for Bayesian consistency. However, designing such priors for $Q^{\ast}$ is an open problem. In the deterministic rewards setting, we have also demonstrated the need for small enough tolerances to ensure an accurate posterior over optimal policies, which highlights the approximation error and sampling efficiency trade-off. \newline

To address the sampling challenges in the case of deterministic rewards, we introduced an adaptive annealing scheme for SMC that aims to maintain the MCMC's effectiveness while avoiding excessive relaxation of the sequence of approximate target posterior distributions. Our experimental results on the Deep Sea benchmark offer promising evidence of our framework's efficacy in exploration and learning through sampling. As extensions, one could explore non-linear parametrisations of $Q^*$ for larger-scale problems; or truly linear-time complexity implementations of sequential particle-based algorithms as discussed in Section \ref{sec:smc_compcost}.\newline

Although we have provided a practical solution for sequentially selecting the tolerances of the posterior distributions, open challenges remain. Future avenues for improvement include the use of intermediate MCMC samples \citep{wastefreesmc}, or the modification of the way MCMC is run across particles for more reliable convergence diagnostics \citep{multichainsmcmc_diagnostics}. MCMC methods for manifolds may also improve sampling efficiency in such posterior landscapes \citep{manifoldlifting}. In addition, techniques such as delayed acceptance MCMC and surrogate likelihood methods \citep{anthonylee_smc_surrogate} could also further reduce computational costs.\newline

In conclusion, this work demonstrates the benefits of posterior sampling for uncertainty quantification and exploration in MDPs. It could serve as a baseline for future comparisons with alternative approximation techniques for modelling and inference.  For example, different implementations that prioritise greater scalability, efficiency, and domain-specific knowledge integration.\newline

%% file: sections/appendix.tex
\section{Proofs}
 \subsection{Bellman optimality equation uniqueness for \texorpdfstring{$Q^*$}{Q*}}
  \label{apd:proofbellmanuniqueness}

While certain forms of Lemma \ref{lem:qvuniqueness} are widely accepted in the literature, a complete proof is not readily available. For completeness and clarity, we provide a proof to ensure the result is well established under the assumptions of interest.\newline
  
 \begin{proof}[Proof of Lemma \ref{lem:qvuniqueness}]

The idea is to use an augmented MDP suggested in \citep{bertsekas2019} to prove the result. We show that if $\mathcal{M}$ satisfies the assumptions of the lemma, so does the augmented MDP. We then apply Theorem \ref{thmvunique} on the augmented MDP, which implies the result of the lemma.\newline
 
Firstly, we present the augmented MDP suggested in \citep{bertsekas2019}. Consider an augmented MDP $\mathcal{M}^\prime=\{\tilde{\mathcal{S}}, \tilde{\mathcal{A}}, p^{\tilde{S}}, p^{\tilde{R}}, \tilde{\rho}\}$ such that $\tilde{\mathcal{S}} := (\bigcup_{s \in \mathcal{S}} \{s\} \times \mathcal{A}_s) \cup (\mathcal{S} \times \{a_\emptyset\})$ for a new action $a_\emptyset$. For any $\tilde{s}, \tilde{s}^\prime \in \tilde{\mathcal{S}}$, we denote the decomposition $\tilde{s}:=(\tilde{s}^0, \tilde{s}^1)$, $\tilde{s}^\prime := ({\tilde{s}}^{\prime 0}, {\tilde{s}}^{\prime1})$, where $\tilde{s}^0,\tilde{s}^{\prime 0} \in \mathcal{S}$, $\tilde{s}^1 \in \mathcal{A}_{\tilde{s}^0} \cup \{a_\emptyset\}$ and $\tilde{s}^{\prime 1} \in \mathcal{A}_{\tilde{s}^{\prime 0}} \cup \{a_\emptyset\}$. Furthermore,\newline
 
if $(\tilde{s}^0, \tilde{s}^1)=(s,a)$, $s \in \mathcal{S}\setminus \{s^g\}$, $a \in \mathcal{A}_s$, $$\tilde{\mathcal{A}}_{\tilde{s}} = \{a_\emptyset\}, \qquad p^{\tilde{S}}(\tilde{s}^\prime|\tilde{s},a_\emptyset)=p^S({\tilde{s}^{\prime0}}|s,a)\delta_{a_\emptyset}(\tilde{s}^{\prime1}), \qquad p^{\tilde{R}}(r|\tilde{s}, a_\emptyset) = p^R(r|s,a);$$
 
if $(\tilde{s}^0, \tilde{s}^1) = (s,a_\emptyset)$, $s \in \mathcal{S}$,
$$\tilde{\mathcal{A}}_{\tilde{s}} = \mathcal{A}_s,\qquad \text{for any } a \in \mathcal{A}_s \text{: }p^{\tilde{S}}(\tilde{s}^\prime|\tilde{s},a) = \delta_{(s,a)}(\tilde{s}^{\prime}),\qquad p^{\tilde{R}}(r|\tilde{s}, a) = \delta_0(r);$$

if $(\tilde{s}^0, \tilde{s}^1) = (s^g,a^g)$,
$$\tilde{\mathcal{A}}_{\tilde{s}}=\{a_\emptyset\},\qquad p^{\tilde{S}}(\tilde{s}^\prime|\tilde{s},a_\emptyset)=p^S(\tilde{s}^{\prime0}|s^g,a^g)\delta_{a^g}(\tilde{s}^{\prime1}),\qquad p^{\tilde{R}}(r|\tilde{s},a_\emptyset) = p^R(r|s^g,a^g) = \delta_0(r).$$

Now, we show that if $\mathcal{M}$ has a unique absorbing state-action pair, so does the augmented MDP.\newline

\begin{lemma}
    $\tilde{s}=(s^g,a^g)$ is the unique absorbing state of $\mathcal{M}^\prime$
\end{lemma}
\begin{proof}
    Let $\tilde{s}=(\tilde{s}^0, \tilde{s}^1) \in \tilde{\mathcal{S}}$ with action $\tilde{a} \in \tilde{\mathcal{A}}_{\tilde{s}}$
    \begin{itemize}[align=left]
        \item[Step 1:] Show that $\tilde{s}=(s^g,a^g)$, $\tilde{a}=a_\emptyset$ is an absorbing state-action pair.

        Firstly, $a_\emptyset $ is the only element in $\tilde{\mathcal{A}}_{\tilde{s}}$. Also,  
        $p^{\tilde{S}}((s^g,a^g)|(s^g,a^g),a_\emptyset) = p^S(s^g|s^g,a^g) \delta_{a^g}(a^g) = \delta_{a^g}(a^g)$, and, 
        $p^{\tilde{R}}(r|(s^g,a^g),a_\emptyset) = \delta_0(r)$. $((s^g,a^g), a_\emptyset)$ is therefore an absorbing state-action pair.

        \item[Step 2:] Uniqueness.
        \begin{itemize}[align=left]
        \item[Case 1:] Suppose $\tilde{s}^0 \neq s^g$, $\tilde{s}^1=a_\emptyset$.
        
        As $\tilde{a} \in \mathcal{A}_{\tilde{s}^0}$, $\tilde{a} \neq a_\emptyset$. Then,
        $p^{\tilde{S}}(\tilde{s}|\tilde{s},\tilde{a}) = \delta_{(\tilde{s}^0, \tilde{a})}((\tilde{s}^0, a_\emptyset)) = 0$. Hence, it is not absorbing.

        \item[Case 2:] Suppose $\tilde{s}^0 \neq s^g, \tilde{s}^1 \neq a_\emptyset$.

        $\tilde{a} = a_\emptyset$, and $p^{\tilde{S}}(\tilde{s}|\tilde{s},\tilde{a}) = p^S(\tilde{s}^0|\tilde{s}^0, \tilde{s}^1)\delta_{a_\emptyset}(\tilde{s}^1)=0$. Hence, it is not absorbing.

        \item[Case 3:] Suppose $\tilde{s}^0=s^g$.
        
        $\tilde{s}^1=a^g$ as $s^g$ is absorbing in $\mathcal{M}$. Hence, $\tilde{a}=a_\emptyset$. This is an absorbing state by Step 1.
        \end{itemize}
    \end{itemize}
    \label{lem:goaluniqueness}
\end{proof}

Next, we show that if $\mathcal{M}$ satisfies Assumption \ref{ass:boeunique}, so does $\mathcal{M}^\prime$.\newline

For any policy $\pi \in \Pi$, define the policy $\varpi(\pi):= \tilde{\pi}$ on $\mathcal{M}^\prime$ such that if $(\tilde{s}^0,\tilde{s}^1) = (s,a)$, $s \in \mathcal{S}$, $a \in \mathcal{A}_s$, $\tilde{\pi}(\tilde{a}|\tilde{s}) := \delta_{a_\emptyset}(\tilde{a})$. If $(\tilde{s}^0,\tilde{s}^1)=(s, a_\emptyset)$, $s \in \mathcal{S}$, then $\tilde{\pi}(\tilde{a}|\tilde{s}) := \pi(\tilde{a}|\tilde{s}^0)$. Let $\tilde{\Pi}=\{\tilde{\pi}:\tilde{\mathcal{S}} \rightarrow \mathscr{P}(\tilde{\mathcal{A}})|\forall \tilde{s} \in \tilde{\mathcal{S}},\,\, \text{supp}(\tilde{\pi}(\bigcdot|\tilde{s})) = \tilde{\mathcal{A}}_{\tilde{s}}\}$. It is clear that $\{\varpi(\pi)\}_{\pi \in \Pi} = \tilde{\Pi}$. Furthermore, suppose $\Pi^d = \{\pi \in \Pi|\pi \text{ is a Dirac measure}\}$ and $\tilde{\Pi}^d = \{\tilde{\pi} \in \tilde{\Pi}|\tilde{\pi} \text{ is a Dirac measure}\}$, it is also easy to see that $\{\varpi(\pi)\}_{\pi \in \Pi^d} = \tilde{\Pi}^d$.\newline

\begin{lemma}
    Suppose $\pi \in \Pi^d$. If $\pi$ is proper, $\tilde{\pi}=\varpi(\pi)$ is proper.
    \label{lem:proper}
\end{lemma}

\begin{proof}
    Let $\mathcal{M}^\prime$ evolves as $\tilde{\mathcal{S}}_0=\tilde{s}_0, \tilde{\mathcal{A}}_0 = \tilde{a}_0, \tilde{\mathcal{S}}_1=\tilde{s}_1, \tilde{\mathcal{A}}_1 = \tilde{a}_1, \dots$, then, the transition dynamics up to $\tau+1$ can be defined via the density:
    \begin{equation}
p^{\tilde{\pi}}(\tilde{s}_{1:\tau+1}, \tilde{a}_{0:\tau}|\tilde{s}_0) = \prod_{t=0}^{\tau} \tilde{\pi}(\tilde{a}_t|\tilde{s}_t) p^{\tilde{S}}(\tilde{s}_{t+1}|\tilde{s}_t,\tilde{a}_t).
    \end{equation}
    Assume $\tilde{s}_{2\tau+1} \neq (s^g,a^g)$. If $\tilde{s}_0^1=a_\emptyset$,
    \begin{equation*}
    p^{\tilde{\pi}}(\tilde{s}_{1:2\tau}, \tilde{a}_{0:2\tau-1}|\tilde{s}_0) = \prod_{t=0}^{\tau-1} \pi(\tilde{a}_{2t}|\tilde{s}_{2t}^0)\delta_{(\tilde{s}_{2t}^0,\tilde{a}_{2t})}(\tilde{s}_{2t+1}) \delta_{a_\emptyset}(\tilde{a}_{2t+1}) p^S(\tilde{s}_{2t+2}^0|\tilde{s}_{2t+1}^0, \tilde{s}_{2t+1}^1) \delta_{a_\emptyset}(\tilde{s}_{2t+2}^1), \\
    \end{equation*}
    \begin{align*}
        p^{\tilde{\pi}}(\tilde{s}_{2\tau}|\tilde{s}_0) &= \int p^{\tilde{\pi}}(\tilde{s}_{1:2\tau}, \tilde{a}_{0:2\tau-1}|\tilde{s}_0) \mathrm{d} \tilde{a}_{0:2\tau-1} \mathrm{d} \tilde{s}_{1:2\tau-1} \\
        &= \int \prod_{t=0}^{\tau-1} \pi(\tilde{a}_{2t}|\tilde{s}^0_{2t}) p^S(\tilde{s}^0_{2t+2}|\tilde{s}^0_{2t},\tilde{a}_{2t}) \mathrm{d} \{\tilde{s}^0_{2t}\}_{t=1}^{\tau-1}\mathrm{d} \{\tilde{a}_{2t}\}_{t=0}^{\tau-1} \delta_{a_\emptyset}(\tilde{s}^1_{2\tau})\\
        &= p^\pi(\tilde{s}^0_{2\tau}|\tilde{s}^0_0)\delta_{a_\emptyset}(\tilde{s}^1_{2\tau}),
    \end{align*}

and if $\tilde{s}_0^1 \in \mathcal{A}_{\tilde{s}_0^0}$,
\begin{align*}
    p^{\tilde{\pi}}(\tilde{s}_{1:2\tau+1},\tilde{a}_{0:2\tau}|\tilde{s}_0^0) =& \delta_{a_\emptyset}(\tilde{a}_0) p^S(\tilde{s}_1^0|\tilde{s}_0^0,\tilde{s}_0^1)\delta_{a_\emptyset}(\tilde{s}_1^1) \\
    \times& \prod_{t=1}^\tau \pi(\tilde{a}_{2t-1}|\tilde{s}_{2t-1}^0) \delta_{(\tilde{s}_{2t-1}^0,\tilde{a}_{2t-1})}(\tilde{s}_{2t}) \delta_{a_\emptyset}(\tilde{a}_{2t})p^S(\tilde{s}_{2t+1}^0|\tilde{s}_{2t}^0,\tilde{s}_{2t}^1)\delta_{a_\emptyset}(\tilde{s}_{2t+1}^1),
\end{align*}
\begin{align*}
    &p^{\tilde{\pi}}(\tilde{s}_{2\tau+1}|\tilde{s}_0) \\
    =& \int p^{\tilde{\pi}}(\tilde{s}_{1:2\tau+1},\tilde{a}_{0:2\tau}|\tilde{s}_0^0) \mathrm{d} \tilde{a}_{0:2\tau} \mathrm{d} \tilde{s}_{1:2\tau} \\
    =& \int p^S(\tilde{s}_1^0|\tilde{s}_0^0,\tilde{s}_0^1) \prod_{t=1}^\tau \pi(\tilde{a}_{2t-1}|\tilde{s}_{2t-1}^0)p^S(\tilde{s}_{2t+1}^0|\tilde{s}^0_{2t-1},\tilde{a}_{2t-1}) \mathrm{d} \{\tilde{s}^0_{2t-1}\}_{t=1}^\tau \mathrm{d} \{\tilde{a}_{2t-1}\}_{t=1}^\tau \delta_{a_\emptyset}(\tilde{s}_{2t+1}^1) \\
    =& \int p^\pi(\tilde{s}^0_{2\tau+1}|\tilde{s}^0_1) p^S(\tilde{s}^0_1|\tilde{s}^0_0,\tilde{s}^1_0) \mathrm{d} \tilde{s}_1^0 \delta_{a_\emptyset}(\tilde{s}_{2t+1}^1).
\end{align*}

Now, as $\pi$ is proper, and the MDP is stationary, taking $\tau \rightarrow \infty$ concludes that $p^{\tilde{\pi}}(\tilde{s}_{\tau}^0 \neq s^g|\tilde{s}_0) \rightarrow 0$ in both scenarios, thus proving that $\tilde{\pi}$ is proper.
\end{proof}

Suppose $\tilde{\pi} \in \tilde{\Pi}^d$ is improper. As there exists $\pi \in \Pi^d$ such that $\tilde{\pi} = \varpi(\pi)$, this implies that $\pi$ is improper by Lemma \ref{lem:proper}.\newline

Let $\tilde{s}^1 = a_\emptyset$, then
\begin{align} \mathbb{E}^{\tilde{\pi}} \Bigg[ \sum_{t=0}^{2\tau} \tilde{R}_t \Big| \tilde{S}_0=\tilde{s} \Bigg] &= \mathbb{E}^{\tilde{\pi}} \Bigg[ \sum_{t=0}^{\tau} \tilde{R}_{2t}(\tilde{S}_{2t},\tilde{A}_{2t}) \Big| \tilde{S}_0=\tilde{s} \Bigg] + \mathbb{E}^{\tilde{\pi}} \Bigg[ \sum_{t=0}^{\tau-1} \tilde{R}_{2t+1}(\tilde{S}_{2t+1},\tilde{A}_{2t+1}) \Big| \tilde{S}_0=\tilde{s} \Bigg] \nonumber \\
&= \mathbb{E}^{\pi} \Bigg[ \sum_{t=0}^{\tau-1} R_{2t+1}(\tilde{S}_{2t+1}^0, \tilde{S}^1_{2t+1}) \Big| \tilde{S}_0^0=\tilde{s}^0 \Bigg]. \label{eqn:lemmvtilde}
\end{align}

Let $V^{\pi,\mathcal{M}}(s) := \lim_{\tau \rightarrow \infty} \mathbb{E}^\pi[\sum_{t=0}^\tau R_t|S_0=s]$ and $V^{\tilde{\pi},\mathcal{M}^\prime}(\tilde{s}):= \lim_{\tau \rightarrow \infty} \mathbb{E}^{\tilde{\pi}}[\sum_{t=0}^\tau \tilde{R}_t|\tilde{S}_0=\tilde{s}]$ for any $s \in \mathcal{S}$, $\tilde{s} \in \tilde{\mathcal{S}}$, $\pi \in \Pi$, $\tilde{\pi} \in \tilde{\Pi}$. As $\pi$ is improper, pick $\tilde{s}^0 \in \mathcal{S}$ such that $V^{\pi,\mathcal{M}}(\tilde{s}^0) = \infty$. Then, taking $\tau \rightarrow \infty$ in Equation \ref{eqn:lemmvtilde} gives $V^{\tilde{\pi},\mathcal{M}^\prime}((\tilde{s}^0,\tilde{s}^1)) = \infty$. Therefore, $\mathcal{M}^\prime$ satisfies the conditions of Theorem \ref{thmvunique}.\newline

Let the reward be deterministic and denote $\tilde{r}(\tilde{s},\tilde{a}):=\tilde{R}_t|\tilde{S}_t=\tilde{s}, \tilde{A}_t=\tilde{a}$. Now, we can apply the result of Theorem \ref{thmvunique} to $\mathcal{M}^\prime$, which allows us to rewrite the uniqueness equation to match the form of the BOEs on $Q^*$ and thereby show the uniqueness of $Q^*$.\newline

By Theorem \ref{thmvunique} on $\mathcal{M}^\prime$, we have
$$     \mathcal{B}^{*,\mathcal{M}^\prime}(V^{*,\mathcal{M}^\prime})(\tilde{s}) :=  \max_{\tilde{a} \in \mathcal{\tilde{A}}_{\tilde{s}}} \tilde{r}_0(\tilde{s},\tilde{a})+ \sum_{\tilde{s}^\prime \in \mathcal{S}} V^{*,\mathcal{M}^\prime}(\tilde{s}^\prime) p^{\tilde{S}}(\tilde{s}^\prime|\tilde{s},\tilde{a}) = V^{*,\mathcal{M}^\prime}(\tilde{s})$$ for all $\tilde{s} \in \mathcal{S}$, and it is the unique fixed point of $\mathcal{B}^{*,\mathcal{M}^\prime}$ under $\{V:\mathcal{\tilde{S}} \rightarrow \mathbb{R}|V((s^g,a^g)) = 0\}$, where $V^{*,\mathcal{M}^\prime}(\tilde{s}) = \sup_{\tilde{\pi} \in \tilde{\Pi}} V^{\tilde{\pi},\mathcal{M}^\prime}(\tilde{s})$.\newline

If $\tilde{s}=(s,a_\emptyset)$, $s \in \mathcal{S}$,
\begin{equation}V^{*,\mathcal{M}^\prime}(\tilde{s}) = \max_{\tilde{a} \in \mathcal{\tilde{A}}_{\tilde{s}}} \sum_{\tilde{s}^\prime \in \mathcal{S}} V^{*,\mathcal{M}^\prime}(\tilde{s}^\prime) p^{\tilde{S}}(\tilde{s}^\prime|\tilde{s},\tilde{a}) = \max_{\tilde{a} \in \mathcal{\tilde{A}}_{\tilde{s}}} V^{*,\mathcal{M}^\prime}((\tilde{s}^0,\tilde{a})), \label{eqn:QproofVstar1} \end{equation}

as $\tilde{r}(\tilde{s},\tilde{a})=0$ for any $\tilde{a} \in \tilde{\mathcal{A}}_{\tilde{s}}$ and $p^{\tilde{S}}(\tilde{s}^\prime|\tilde{s},a)=\delta_{(s,\tilde{a})}(\tilde{s}^\prime)$.\newline

If $\tilde{s}=(s,a)$, $s \in \mathcal{S}$, $a \in \mathcal{A}_s$, which is the input of interest,
\begin{align}
    V^{*,\mathcal{M}^\prime}(\tilde{s}) &= \max_{\tilde{a}\in \mathcal{A}_{\tilde{s}}} r(\tilde{s}^0,\tilde{s}^1) + \sum_{\tilde{s}^{\prime0} \in \mathcal{S}} V^{*,\mathcal{M}^\prime}((\tilde{s}^{\prime0},a_\emptyset)) p^S(\tilde{s}^{\prime0}|\tilde{s}^0,\tilde{s}^1) \nonumber \\
    &= r(\tilde{s}^0,\tilde{s}^1) + \sum_{\tilde{s}^{\prime0} \in \mathcal{S}} \max_{\tilde{a} \in \mathcal{\tilde{A}}_{\tilde{s}}} V^{*,\mathcal{M}^\prime}((\tilde{s}^0,\tilde{a})) p^S(\tilde{s}^{\prime0}|\tilde{s}^0,\tilde{s}^1), \label{eqn:QproofVstar2}
\end{align}
where the last equality comes from Equation \ref{eqn:QproofVstar1} above.\newline

Thus, under $V^{*,\mathcal{M}^\prime}((s^g,a^g)) = 0$, there is a unique solution that satisfies the equations in Equation \ref{eqn:QproofVstar1} and Equation \ref{eqn:QproofVstar2}. As the inputs of the two set of equations are disjoint, it implies that there exists a unique solution to the set of equations in Equation \ref{eqn:QproofVstar2}. Setting $Q^*(s,a) := V^{*,\mathcal{M}^\prime}((s,a))$ for any $s \in \mathcal{S}$, $a \in \mathcal{A}_s$ finishes the proof.

 \end{proof}

 \subsection{Theoretical form of posterior under tabular \texorpdfstring{$Q_\theta$}{Qθ} and Gaussian likelihood}
 \label{apd:proofpsrl}

\begin{proof}[Proof of Theorem \ref{thm:psrl}]
Firstly, rewrite $$p(\theta \in E^*|\mathcal{D}) = \frac{\int_{E^*}(p(\theta,r_{1:n}|s_{1:n},a_{1:n})\mathrm{d}\theta}{p(r_{1:n}|s_{1:n},a_{1:n})}$$
and note that $$E^\ell = \Theta \cap \bigcap\limits_{\substack{s^\prime \in \mathcal{S}^{\prime\mathcal{D}} \\ s^\prime \notin \mathcal{S}^g}}^n
\Big\{\theta \in \Theta| \theta_{\nu(s^\prime,\ell(s^\prime))} = \max\limits_{a^\prime \in \mathcal{A}_{s^\prime}} \theta_{\nu(s^\prime,a^\prime)} \Big\}.$$\newline

Denote the prior $p^\Theta(\theta) := \prod_{i=1}^{d_{\Theta}} \mathcal{N}(\theta_i; 0, \sigma^2)$. It is easy to see that $\bigcup\limits_{\ell \in \mathcal{S}^{\prime\mathcal{S}} \rightarrow \mathcal{A}^{\prime\mathcal{D}}} E^\ell = \Theta$, and for $\ell,\ell^\prime \in \ell^{\prime\mathcal{D}}$ such that $\ell \neq \ell^\prime$, $E^\ell \cap E^{\ell^\prime} = \emptyset$ $p^\Theta$-a.s. Also, let $E^\Theta = \{\theta \in \Theta| \theta_i \neq \theta_j$ for all $i,j \in \{1,\dots,d_{\Theta}\}, i \neq j\}$. It is clear that $p^\Theta(E^\Theta)=1$.\newline

With the partition $\{E^\ell\}_{\ell \in \ell^{\prime\mathcal{D}}}$ of $\Theta$, we can now rewrite $p(\theta,r_{1:n},\theta \in E^\Theta|s_{1:n},a_{1:n})$ in the following form:
\begin{align*}
&p(\theta,r_{1:n},\theta \in E^\Theta|s_{1:n},a_{1:n}) \\
=& \Bigg[ \prod_{i=1}^n \sum_{\ell \in \ell^{\prime\mathcal{D}}} \mathcal{N}\Big(r_i;\theta_{\nu(s_i,a_i)}-\sum_{s_i^\prime \in \mathcal{S}} p^S(s_i^\prime|s_i,a_i)\max_{a_i^\prime \in \mathcal{A}_{s_i^\prime}}\theta_{\nu(s_i^\prime,a_i^\prime)},\epsilon^2\Big) \mathbbm{1}(\theta \in E^{\ell})\Bigg] p^\Theta(\theta) \mathbbm{1}(\theta \in E^\Theta)\\
=& \sum_{\ell \in \ell^{\prime\mathcal{D}}} \Big[ \prod_{i=1}^n \mathcal{N}(r_i;\theta_{\nu(s_i,a_i)}- \sum_{s_i^\prime \in \mathcal{S}} p^S(s_i^\prime|s_i,a_i)\max_{a_i^\prime \in \mathcal{A}_{s_i^\prime}}\theta_{\nu(s_i^\prime,a_i^\prime)}, \epsilon^2) \mathbbm{1}(\theta \in E^{\ell}) \Big] p^\Theta(\theta)\mathbbm{1}(\theta \in E^\Theta) \\
=& \sum_{\ell \in \ell^{\prime\mathcal{D}}} [p(\theta,r_{1:n},\theta \in E^{\ell}|s_{1:n},a_{1:n})]\mathbbm{1}(\theta \in E^\Theta).
\end{align*}

We can now rewrite the numerator of $p(\theta \in E^*|\mathcal{D})$ using the partition:
\begin{align*}
    \int_{E^*} p(\theta,r_{1:n}|s_{1:n},a_{1:n})\mathrm{d} \theta &= \int_{E^*} p(\theta,r_{1:n}, \theta \in E^\Theta|s_{1:n},a_{1:n}) + p(\theta,r_{1:n}, \theta \in {E^\Theta}^c|s_{1:n},a_{1:n})\mathrm{d} \theta \\
    &= \int_{E^*} \sum_{\ell \in \ell^{\prime\mathcal{D}}} [p(\theta,r_{1:n},\theta \in E^{\ell})]\mathbbm{1}(\theta \in E^\Theta) \mathrm{d}\theta \\
    &= \sum_{\ell \in \ell^{\prime\mathcal{D}}} \int_{E^* \cap E^\ell \cap E^\Theta} p(\theta,r_{1:n}|s_{1:n},a_{1:n}) \mathrm{d} \theta \\
    &= \sum_{\ell \in \ell^{\prime\mathcal{D}}} \int_{E^* \cap E^\ell} p(\theta,r_{1:n}|s_{1:n},a_{1:n}) \mathrm{d} \theta, \\
\end{align*}

and similarly, for the denominator,
\begin{equation*}
   p(r_{1:n}|s_{1:n},a_{1:n})  = \sum_{\ell \in \ell^{\prime\mathcal{D}}} \int_{E^\ell \cap E^\Theta} p(\theta,r_{1:n}|s_{1:n},a_{1:n}) \mathrm{d} \theta = \sum_{\ell \in \ell^{\prime\mathcal{D}}} \int_{E^\ell} p(\theta,r_{1:n}|s_{1:n},a_{1:n}) \mathrm{d} \theta.
\end{equation*}

We now define auxiliary distributions that can help us to evaluate the integrals.\newline

Consider an auxiliary joint distribution $p^{\ell}$ as follows:
$$p^{\ell}(\theta,r_{1:n}) := \prod_{i=1}^n \mathcal{N}(r_i;\theta_{\nu(s_i,a_i)} - 
\sum_{s_i^\prime \in \mathcal{S}} p^S(s_i^\prime|s_i,a_i) \theta_{\nu(s_i^\prime,\ell(s_i^\prime))},\epsilon^2) p^\Theta(\theta),$$

with the conditional distribution $r_{1:n}|\theta;\ell \sim \mathcal{N}(r_{1:n}; B^\ell \theta, \epsilon^2 I)$.\newline

Thus, $(\theta,r_{1:n})$ are jointly Gaussian under $p^\ell$, i.e.
\begin{equation}
\begin{pmatrix} \theta \\ r_{1:n} \end{pmatrix} \sim \mathcal{N}\Bigg(0, \begin{pmatrix} \sigma^2I_{d_{\Theta}} & \sigma^2{B^{\ell}}^T \\ \sigma^2 B^\ell & \sigma^2 B^\ell {B^\ell}^T + \epsilon^2I_n\end{pmatrix}\Bigg).
\label{eqn:jointgaussian:apdproofpsrl}
\end{equation}
Furthermore, the posterior is of the form:
$$\theta|r_{1:n} \sim \mathcal{N}(\sigma^2 {B^\ell}^T(\sigma^2 B^\ell {B^\ell}^T + \epsilon^2 I_n)^{-1} r_{1:n}, \sigma^2 I_{d_{\Theta}} - \sigma^4 {B^\ell}^T(\sigma^2 B^\ell {B^\ell}^T + \epsilon^2 I_n)^{-1} B^\ell).$$

By construction, $$\int_{E^* \cap E^\ell \cap E^\Theta} p(\theta,r_{1:n}|s_{1:n},a_{1:n}) \mathrm{d}\theta = \int_{E^* \cap E^\ell} p^\ell(\theta,r_{1:n}) \mathrm{d} \theta = p^\ell(r_{1:n}) p^\ell(\theta \in E^* \cap E^\ell|r_{1:n}),$$

and likewise,
$$\int_{E^\ell \cap E^\Theta} p(\theta,r_{1:n}|s_{1:n},a_{1:n}) \mathrm{d}\theta = p^\ell(r_{1:n}) p^\ell(\theta \in E^\ell|r_{1:n}),$$

While the marginal $p^\ell(r_{1:n})$ can be read off from the joint Gaussian model above in Equation \ref{eqn:jointgaussian:apdproofpsrl},   $p^\ell(\theta \in E^\ell|r_{1:n})$ can be evaluated by observing that it is simply a multivariate Gaussian cumulative distribution function, which can be approximated by suitable Monte-Carlo methods if $E^\ell \cap E^* \neq \emptyset$, otherwise it is simply zero. The same argument holds for $p^\ell(\theta \in E^*  \cap E^\ell|r_{1:n})$ for simple $E^*$, such as 
$E^*=\bigcap_{s \in \mathcal{S}}\bigcap_{\substack{a \in \mathcal{A}_{s} \\ a \neq \mu(s)}} \{\theta \in \Theta| \theta_{\nu(s,a)} - \theta_{\nu(s,\mu(s))} \leq 0 \}$ for computing the probabilities of optimal actions. Hence, we can now evaluate $\int_{E^*} p(\theta,r_{1:n}|s_{1:n},a_{1:n})\mathrm{d} \theta$ and $ p(r_{1:n}|s_{1:n},a_{1:n})$ and, hence, $p(\theta \in E^*|s_{1:n},a_{1:n})$. \newline

Thus, the overall form of the posterior of interest is:
\begin{equation}
p(\theta \in E^*|\mathcal{D}) = \frac{\sum_{\ell \in \ell^{\prime\mathcal{D}}} p^\ell(r_{1:n}) p^\ell(\theta \in E^* \cap E^\ell|r_{1:n})}{\sum_{\ell \in \ell^{\prime\mathcal{D}}} p^\ell(r_{1:n}) p^\ell(\theta \in E^\ell|r_{1:n})}.
\end{equation}

 \end{proof}

 \subsection{Unidentifiable likelihood for MDPs which contain non-goal recurrent states}
 \label{apd:nondecaylkhproof}
\begin{proof}[Proof of Theorem \ref{thm:loopthm}]

We divide the proof into several steps, some of which are not strictly necessary for the proof, but are presented to provide additional insights into MDPs with a non-goal recurrent state $s^r$.\newline

We first show that a greedy policy that results in $s^r$ is improper. Therefore, the existence of $s^r$ implies the existence of an improper policy in a finite state-space MDP. The converse is straightforward and is therefore omitted. \newline

\underline{\textit{Non-goal recurrent state implies improper policy}}\newline
Let $\pi_\theta$ be the greedy policy. There exists $s_0^r \in \text{supp}(\rho)$, $t_r \in \mathbb{Z}_{\geq 0}$ such that $p^{\pi_\theta}(S_{t_r}=s^r|S_0=s_0^r) > 0$ and $p^{\pi_\theta}(S_t=s^r \text{ for some } t \in \mathbb{Z}_{\geq 1}|S_0=s^r)=1$. Since $\mathcal{S}$ is finite, the Markov chain starting at $s^r$ with transition $p^{\pi_\theta}(S_{t+1}=s^\prime|S_t=s)=p^S(S_{t+1}=s^\prime|S_t=s, A_t=\pi_\theta(s))$ for $s,s^\prime \in \mathcal{S}$ satisfies $\lim_{t \rightarrow \infty} p^{\pi_\theta}(S_t=s^r|S_{t_r}=s^r) > 0$ if the limit exists, or $\exists e > 0$ such that for any $t_l>t_r$, $\exists t > t_l$ such that $p^{\pi_\theta}(S_t=s^r|S_{t_r}=s^r) > e$ \citep{markovchain}. Since for $t > t_r$,
$$ p^{\pi_\theta}(S_t \notin \mathcal{S}^g|S_0=s_0^r) = \sum_{s \in \mathcal{S}} p^{\pi_\theta}(S_t \notin S^g|S_{t_r}=s) p(S_{t_r}=s|S_0=s_0^r),$$
and that $s^r \notin \mathcal{S}^g$, this implies $\lim_{t \rightarrow \infty} p^{\pi_\theta}(S_t \notin \mathcal{S}^g|S_0=s^r_0) < 1$ if $\lim_{t \rightarrow \infty} p^{\pi_\theta}(S_t=s^r|S_{t_r}=s^r)$ exists, otherwise, it is also clear that $\lim_{t \rightarrow \infty} p^{\pi_\theta}(S_t \notin \mathcal{S}^g|S_0=s^r) < 1$.  Therefore, $\pi_\theta$ is improper. Note that this implies that $\pi_\theta$ is not optimal.\newline

We now define some additional notations for decision rules deployed from time $1$ onwards after moving away from $S_0=s_0 \in \mathcal{S}$ taking action $a_0 \in \mathcal{A}_{s_0}$.\newline

\underline{\textit{Notations}}\newline
Let $\tilde{\pi}_t:\mathcal{S} \rightarrow \mathcal{A}$ be a decision rule at time $t \in \mathbb{Z}_{\geq 1}$ such that $\tilde{\pi}_t(s_t) \in \mathcal{A}_{s_t}$ for any $S_t=s_t \in \mathcal{S}$ encountered at time $t$. Define $\tilde{\pi}: \mathbb{Z}_{\geq 1} \times \mathcal{S} \rightarrow \mathcal{A}$ such that $\tilde{\pi}(t,s)=\tilde{\pi}_t(s)$, the set of all such $\tilde{\pi}$ as $\tilde{\Pi}$, and for any $\tau \in \mathbb{Z}_{\geq 1}$, 
$$
p^{\tilde{\pi}}(s_{1:\tau}, a_{1:\tau}|S_0=s_0, A_0=a_0)= \prod_{t=1}^{\tau}[p^S(s_t|s_{t-1},a_{t-1}) \mathbbm{1}(a_t \in \tilde{\pi}(t,s_t))].$$
The notations are defined similarly for their marginal and conditional probabilities. Note that it is not necessary to assume policy stationarity in this proof.\newline

We are now ready to present the main body of the proof, where we give the choice of $u$, and the subset of $\theta$ that can lead to likelihood invariance. Note that although refining these definitions is possible specific to the dataset, the definitions presented below are applicable to all possible datasets $\mathcal{D}$ for simplicity.\newline

\underline{\textit{Main body of proof}}\newline
Let the set of state-action pairs that can lead to $s^r$ be

\begin{align*}
\mathcal{C}^r = \{(s,a)|\forall s \in \mathcal{S}, a \in \mathcal{A}_s,\,\, &\exists \tilde{\pi}_t:\mathcal{S} \rightarrow \mathcal{A}, \,\, t \in \mathcal{Z}_{\geq 1} \\
&\text{ such that } p^{\tilde{\pi}_t}(S_t=s^r \text{ for some } t \in \mathcal{Z}_{\geq 1} | S_0=s, A_0=a) >0 \}.
\end{align*}

Let $u \in [0,1]^{d_{\Theta}}$ such that $$u_i=\max_{\tilde{\pi} \in \tilde{\Pi}} p^{\tilde{\pi}}(S_t=s^r \text{ for some } t \in \mathbb{Z}_{\geq 1}|(S_0,A_0) = \nu^{-1}(i)),$$

and let 
$$\mathcal{O} = \{\theta \in \Theta | \forall s \in \mathcal{S} \text{ such that } \exists a \in \mathcal{A}_s, (s,a) \in \mathcal{C}^r, \argmax_{a^\prime \in \mathcal{A}_s} \theta_{\nu(s,a^\prime)} \cap \argmax_{a^\prime \in \mathcal{A}_s} u_{\nu(s,a^\prime)} \neq \emptyset\}.$$  This is the set of $\theta \in \Theta$ in which at any state that can reach $s^r$, a derived greedy policy of $\theta$ always pick actions that maximises the probability of eventually leading to $s^r$, i.e. an action $a \in \mathcal{A}_s$ maximising $u_{\nu(s,\cdot)}$ maximises $\theta_{\nu(s,\cdot)}$. Because this condition is to be satisfied independently for each $s \in \mathcal{S}$, it is clear that if $\Theta$ is taken as $\mathbb{R}^{d_\Theta}$, the Lebesgue measure on $\mathbb{R}^{d_\Theta}$ of $\mathcal{O}$ is infinite. \newline

Finally, choose
\begin{equation*}
c_\theta = \begin{cases} \max\limits_{s \in \mathcal{S}} \max\limits_{a \in \argmax\limits_{a \in \mathcal{A}_s} \theta_{\nu(s,a)}}\max\limits_{\substack{\bar{a} \in \mathcal{A}_s \\ \bar{a} \neq a}} \frac{\theta_{\nu(s,\bar{a})} - \theta_{\nu(s,a)}}{u_{\nu(s,a)}-u_{\nu(s,\bar{a})}} &  \parbox[t]{8.5cm}{$\exists s \in \mathcal{S} \text{ such that } |\mathcal{A}_s| \geq 2, \text{ and } \exists a,a^\prime \in \mathcal{A}_s \text{ such that } u_{\nu(s,a)} \neq u_{\nu(s,a^\prime)}$}\\ 
-\infty & \text{ otherwise },\\
\end{cases}
\end{equation*}
which is non-positive due to the definition of $\mathcal{O}$.\newline

The likelihood function has the form:
$$ L(\theta|\mathcal{D}) = \prod_{(s,a,r) \in \mathcal{D}} \mathcal{N}(r;\theta_{\nu(s,a)} - \sum_{s^\prime \in \mathcal{S}} p^S(s^\prime|s,a) \max_{a^\prime \in \mathcal{A}_{s^\prime}} \theta_{\nu(s^\prime,a^\prime)}, \epsilon^2).$$

We now have the following lemma, which shows that the likelihood is invariant with these choices of $u$, $\mathcal{O}$ and $c_\theta$.

\begin{lemma}
\label{lem:simpleex}
For any $c > c_\theta$, $\theta \in \mathcal{O}$, 
\begin{equation}
(\theta + cu)_{\nu(s,a)} - \sum_{s^\prime \in \mathcal{S}} p^S(s^\prime|s,a) \max_{a^\prime \in \mathcal{A}_{s^\prime}} (\theta + cu)_{\nu(s^\prime, a^\prime)} = \theta_{\nu(s,a)} - \sum_{s^\prime \in \mathcal{S}} p^S(s^\prime|s,a) \max_{a^\prime \in \mathcal{A}_{s^\prime}} \theta_{\nu(s^\prime, a^\prime)}.
\label{eqn:lemma:loopthm_alg}
\end{equation}
\label{lem:loopthm_alg}
\end{lemma}

\begin{proof}
Firstly, we show that for any $s \in \mathcal{S}$, 
\begin{equation}
\max_{a \in \mathcal{A}_s} (\theta + cu)_{\nu(s,a)} = \max_{a \in \mathcal{A}_s} \theta_{\nu(s,a)} + c \max_{a \in \mathcal{A}_s} u_{\nu(s,a)} \label{eqn:lem:step1}.
\end{equation}

Given that $\theta \in \mathcal{O}$,\newline
\underline{Case 1}: For $s \in \mathcal{S}$ such that $\exists a \in \mathcal{A}_s$ and $(s,a) \in \mathcal{C}^r$, let $a \in \argmax_{a^\prime \in \mathcal{A}_s} \theta_{\nu(s,a^\prime)} \cap \argmax_{a^\prime \in \mathcal{A}_s} u_{\nu(s,a^\prime)}$.\newline

If $|\mathcal{A}_s| \geq 2$, and if $\exists a^\prime \in \mathcal{A}_s$ such that $u_{\nu(s,a)} \neq u_{\nu(s,a^\prime)}$, then 
\begin{align*}c_\theta \geq \frac{\theta_{\nu(s,a^\prime)} - \theta_{\nu(s,a)}}{u_{\nu(s,a)} - u_{\nu(s,a^\prime)}} \,\, &\Rightarrow \,\, c_\theta(u_{\nu(s,a)} - u_{\nu(s,a^\prime)}) \geq \theta_{\nu(s,a^\prime)} - \theta_{\nu(s,a)} \\
&\Rightarrow \,\, \theta_{\nu(s,a)} + c_\theta u_{\nu(s,a)} \geq \theta_{\nu(s,a^\prime)} + c_\theta u_{\nu(s,a^\prime)}.
\end{align*}

Now, if $c > c_\theta$, $c(u_{\nu(s,a)} - u_{\nu(s,a^\prime)}) > c_\theta(u_{\nu(s,a)} - u_{\nu(s,a^\prime)})$, which implies that 
$$(\theta+cu)_{\nu(s,a)} - (\theta + cu)_{\nu(s,a^\prime)} \geq (\theta + c_\theta u)_{\nu(s,a)} - (\theta + c_\theta u)_{\nu(s,a^\prime)} \geq 0.$$

On the other hand, if $u_{\nu(s,a)} = u_{\nu(s,a^\prime)}$ $\forall a^\prime \in \mathcal{A}_s$, it is clear that $\theta_{\nu(s,a)} + c u_{\nu(s,a)} \geq \theta_{\nu(s,a^\prime)} + c u_{\nu(s,a^\prime)}$ for any $c \in \mathbb{R}$.\newline

Hence, in either scenario, $(\theta + cu)_{\nu(s,a)} \geq (\theta + cu)_{\nu(s,a^\prime)}$ for any $c > c_\theta$. \newline

Therefore, $a$ maximises $(\theta + cu)_{\nu(s,a)}$, and 
$$\max_{a^\prime \in \mathcal{A}_s} (\theta + cu)_{\nu(s,a^\prime)} = (\theta + cu)_{\nu(s,a)}  = \max_{a^\prime \in \mathcal{A}_s} \theta_{\nu(s,a^\prime)} + c \max_{a^\prime \in \mathcal{A}_s} u_{\nu(s,a^\prime)}.$$
Hence, Equation \ref{eqn:lem:step1} holds.\newline

\underline{Case 2}: For $s \in \mathcal{S}$ which $\not\exists a \in \mathcal{A}_s$ such that $(s,a) \in \mathcal{C}^r$, $u_{\nu(s,a)} = 0$ $\forall a \in \mathcal{A}_s$. Hence, 
Equation \ref{eqn:lem:step1} also holds.\newline

Thus, by Equation \ref{eqn:lem:step1}, showing Equation \ref{eqn:lemma:loopthm_alg} is equivalent to showing 
$$u_{\nu(s,a)} = \sum_{s^\prime \in \mathcal{S}} p^S(s^\prime|s,a) \max_{a^\prime \in \mathcal{A}_{s^\prime}} u_{\nu(s^\prime, a^\prime)}$$ for all $s \in \mathcal{S}, a \in \mathcal{A}_s$.\newline

To show this, note that 
\begin{align*}
u_{\nu(s,a)} =& \max_{\tilde{\pi} \in \tilde{\Pi}} p^{\tilde{\pi}} (S_t=s^r \text{ for some } t \in \mathbb{Z}_{\geq 1} | S_0=s, A_0=a) \\
=& \max_{\tilde{\pi} \in \tilde{\Pi}} \sum_{s^\prime \in \mathcal{S}} p^{\tilde{\pi}}(S_t=s^r \text{ for some } t \in \mathcal{Z}_{\geq 1} | S_1=s^\prime, A_1 = \tilde{\pi}(1,s^\prime)) p^S(s^\prime|s,a) \\
=& \sum_{s^\prime \in \mathcal{S}} \max_{a^\prime \in \mathcal{A}_{s^\prime}} \max_{\substack{\tilde{\pi} \in \tilde{\Pi}\\ \tilde{\pi}(1,s^\prime) = a^\prime}} p^{\tilde{\pi}}(S_t=s^r \text{ for some } t \in \mathcal{Z}_{\geq 1} | S_1=s^\prime, A_1 = a^\prime) p^S(s^\prime|s,a) \\
=& \sum_{s^\prime \in \mathcal{S}} \max_{a^\prime \in \mathcal{A}_{s^\prime}} \max_{\substack{\tilde{\pi} \in \tilde{\Pi}\\ \tilde{\pi}(1,s^\prime) = a^\prime}} \big(p^{\tilde{\pi}}(S_1=s^r|S_1=s^\prime, A_1=a^\prime) + p^{\tilde{\pi}}(S_1 \neq s^r |S_1=s^\prime, A_1=a^\prime) \\
& \cdot p^{\tilde{\pi}}(S_t=s^r \text{ for some } t \in \mathcal{Z}_{\geq 2} | S_1=s^\prime, A_1 =a^\prime)\big) p^S(s^\prime|s,a) \\
=& \sum_{s^\prime \in \mathcal{S}} \big(\mathbbm{1}(s^r \in s^\prime) + (1-\mathbbm{1}(s^r \in s^\prime)) \max_{a^\prime \in \mathcal{A}_{s^\prime}} u_{\nu(s^\prime, a^\prime)}\big) p^S(s^\prime|s,a).
\end{align*}
Since $\mathbbm{1}(s^r \in s^\prime) \max\limits_{a^\prime \in \mathcal{A}_{s^\prime}} u_{\nu(s^\prime, a^\prime)} = \begin{cases} 1& \text{if } s^\prime = s^r \\ 0 & \text{otherwise} \end{cases}$,  we have the result.

\end{proof}
Therefore, 
$$    L(\theta|\mathcal{D}) = \prod_{(s,a,r) \in \mathcal{D}} \mathcal{N}(r;(\theta + cu)_{\nu(s,a)} - \sum_{s^\prime \in \mathcal{S}} p^S(s^\prime|s,a) \max_{a^\prime \in \mathcal{A}_{s^\prime}} (\theta + cu)_{\nu(s^\prime, a^\prime)}, \epsilon^2) = L(\theta+cu|\mathcal{D}) $$
for $\theta \in \mathcal{O}$ and for all $c > c_\theta$.\newline

\underline{\textit{Final Remark}}\newline
Finally, to remark, and as a sanity check, if $\theta \in \mathcal{O}$ and $s \in \mathcal{S}$ such that $\exists a \in \mathcal{A}_s$ satisfying  $(s,a) \in \mathcal{C}^r$, and $c > c_\theta$, $\argmax_{a \in \mathcal{A}_s} (\theta + cu)_{\nu(s,a)} \cap \argmax_{a \in \mathcal{A}_s} u_{\nu(s,a)} \neq \emptyset$ by the proof of Lemma \ref{lem:loopthm_alg}. This implies that $\theta + cu \in \mathcal{O}$. \end{proof}

\subsection{Derivation of the posterior probability for exploration in the simple 5D MDP example}
\label{apd:5dexampleproof}

\begin{proof}[Proof of Lemma \ref{lem:5dmdp}]

    Recall that $\mathcal{D}=\{(s^1,a^1,r^1),(s^1,a^2,r^2),(s^2,a^1,r^3),(s^3,a^2,r^4)\}$. We have $\mathcal{S}^g=\{s^4,s^5\}$, $\nu(s^1,a^1)=1$, $\nu(s^1,a^2)=2$, $\nu(s^2,a^1)=3$, $\nu(s^3,a^2)=4$, $\nu(s^4,a^g)=5$ and $\nu(s^5,a^g)=6$ and  $r_{1:4}=(r^1,r^2,r^3,r^4)^T$. The likelihood function is
    $$p(\theta,r_{1:4}|s_{2:5},a_{2:5}) = \mathcal{N}(r^1;\theta_1-\theta_3,\epsilon^2)\mathcal{N}(r^2;\theta_2-\theta_4,\epsilon^2)\mathcal{N}(r^3;\theta_3,\epsilon^2)\mathcal{N}(r^4;\theta_4,\epsilon^2)\mathcal{N}(\theta;0,\sigma^2I).$$
    
    Next, we have $\mathcal{S}^{\prime\mathcal{D}} = \{s^2,s^3,s^4,s^5\}$. Since all states except for $s^1$ have one admissible action only respectively, $\ell^{\prime \mathcal{D}}=\{\ell\}$ where $\ell(s^2)=a^1$, $\ell(s^3)=a^2$, $\ell(s^4)=a^g$ and $\ell(s^5)=a^g$. Hence, we have $E^\ell = \Theta$. Now, we can simply apply the conjugate posterior result in Theorem \ref{thm:psrl}. \newline

    Apply the definition of the theorem gives
    $$B^\ell = \begin{pmatrix}
        1 & 0 & -1 & 0\\
        0 & 1 & 0 & -1 \\
        0 & 0 & 1 & 0 \\
        0 & 0 & 0 & 1 \\
    \end{pmatrix},\,\, \Gamma^\ell = \frac{1}{\sigma^4 + 3\sigma^2\epsilon^2 + \epsilon^4} \begin{pmatrix}
        \sigma^2 + \epsilon^2 & 0 & \sigma^2 & 0 \\
        0 & \sigma^2 + \epsilon^2 & 0 & \sigma^2 \\
        \sigma^2 & 0 & 2\sigma^2 + \epsilon^2 & 0 \\
        0 & \sigma^2 & 0 & 2\sigma^2 + \epsilon^2
    \end{pmatrix}.$$
    
    Hence, we have
    $$\mu^\ell_{\theta|r} = \frac{\sigma^2}{\sigma^4 + 3\sigma^2\epsilon^2 + \epsilon^4} \begin{pmatrix}
        r^1 ( \sigma^2 + \epsilon^2) + r^3 \sigma^2 \\
        r^2 (\sigma^2 + \epsilon^2) + r^4 \sigma^2 \\
        r^3 (\sigma^2 + \epsilon^2) - r^1 \epsilon^2 \\
        r^4 (\sigma^2 + \epsilon^2) - r^2 \epsilon^2
    \end{pmatrix},$$
and
$$
\Sigma^\ell_{\theta|r} = \frac{\sigma^2\epsilon^2}{\sigma^4 + 3\sigma^2\epsilon^2 + \epsilon^4} \begin{pmatrix}
    2\sigma^2 + \epsilon^2 & 0 & \sigma^2 & 0 \\
    0 & 2\sigma^2 + \epsilon^2 & 0 & \sigma^2 \\
    \sigma^2 & 0 & \sigma^2 + \epsilon^2 & 0 \\
    0 & \sigma^2 & 0 & \sigma^2 + \epsilon^2
\end{pmatrix}.
$$

Therefore, $\theta|\mathcal{D} \sim \mathcal{N}(\mu_{\theta|r}^\ell,\Sigma_{\theta|r}^\ell)$.

Let $E^* = \{\theta|\theta_2 - \theta_1 < 0\}$. Then, 
\begin{align*}
\pabc(\theta \in E^*|\mathcal{D}) &= \Phi\Bigg(-\frac{\sigma((r^2+r^4-r^1-r^3)\sigma^2+(r^2-r^1)\epsilon^2)}{\epsilon\sqrt{2(2\sigma^2+\epsilon^2)(\sigma^4 + 3\sigma^2\epsilon^2 + \epsilon^4)}}\Bigg) \\
&= \Phi \Bigg(\frac{d\epsilon^2-c}{\epsilon/\sigma \sqrt{2(2\sigma^2+\epsilon^2)(\sigma^4 + 3\sigma^2\epsilon^2 + \epsilon^4)}}\Bigg) \\
&= \Phi \Bigg(\frac{kd-c}{\sigma\sqrt{2k(2+k)(k^2+3k+1)}} \Bigg).
\end{align*}
\end{proof}

\section{More discussions on Thompson sampling for MABs and posterior sampling for MDPs}
\label{apd:psrl_equivalence}
\subsection{Tie-breaking rules for Thompson sampling for MABs}
\label{apd:psrl_mab}

When $\argmax_k \bar{r}_k(\theta)$ is not $p(\bigcdot|\mathcal{D}_\tau)$-almost-surely unique, the integral given in Equation \ref{eq:ts_mab} $\int_{\Theta} \mathbbm{1}\Big(k^\star \in \argmax_k \bar{r}_k(\theta)\Big) p(\theta|\mathcal{D}_{\tau})\mathrm{d}\theta$, which is the marginal posterior probability that $A_{\tau+1} = k^*$ is optimal,  does not define a probability mass function. This is because the subset of $\theta$ that contains more than one optimal action has a non-zero probability mass under $p(\cdot|\mathcal{D}_\tau)$, and this implies that optimality does not partition the parameter space $\Theta$. Since for every $\theta$, there exists at least one optimal action, in such scenarios, a tie-breaking rule is needed so that we can define a probability mass function on an action that ``is optimal and is selected''.\newline

Let the marginally probability that arm $k^\star$ is selected at action $A_{\tau+1}$ given $\mathcal{D}_\tau$ be
$$\mathbb{P}(A_{\tau+1}=k^\star|\mathcal{D}_\tau) = \int \mathbb{P}(A_{\tau+1}=k^\star|\theta)p(\theta|\mathcal{D}_\tau) \mathrm{d}\theta.$$

An arm $k^\star$ is chosen given $\theta$ if it is optimal given $\theta$ as well as being selected by the tie-breaking rule. That is,

\begin{align*}
\mathbb{P}(A_{\tau+1}=k^\star|\theta) &= \mathbb{P}(A_{\tau+1}=k^\star,k^\star \in \argmax_{k} \bar{r}_k(\theta)|\theta) \\
&= \mathbb{P}(A_{\tau+1}=k^\star|k^\star \in \argmax_{k} \bar{r}_k(\theta),\theta) \mathbbm{1}\Big(k^* \in \argmax_{k} \bar{r}_k(\theta)\Big),
\end{align*}

where $\mathbb{P}(A_{\tau+1}=k^\star|k^\star \in \argmax_{k} \bar{r}_k(\theta),\theta)$ represents the tie-breaking rule. A simple tie-breaking rule can be defined as
$$\mathbb{P}(A_{\tau+1}=k^\star|k^\star \in \argmax_{k} \bar{r}_k(\theta),\theta) = \frac{1}{\sum_{k=1}^K \mathbbm{1}(\bar{r}_k(\theta) = \bar{r}_{k^\star}(\theta))},$$ which chooses optimal actions uniformly given $\theta$.

\subsection{The optimal policy interpretation of posterior sampling for MDPs with tie-breaking rules}
\label{apd:psrl_mdp}

First of all, assume that $\argmax_{a \in \mathcal{A}_{s}}Q_\theta(s,a)$ is $p^\Theta(\bigcdot|\mathcal{D}_{t_i-1})$-almost-surely unique. To see why sampling a $\theta \sim p^\Theta(\cdot|\mathcal{D}_{t_i-1})$ and acting greedily to $Q_\theta$ during time steps $t \in \{t_i,\dots,t_{i+1}-1\}$ is equivalent to sampling a deterministic policy $M=\mu$ according to the probability $\mathbb{P}(M=\mu|\mathcal{D}_{t_i-1}) = p^\Theta(\{\theta| \forall s \in \mathcal{S}, \mu(s) \in \argmax_{a \in \mathcal{A}_s} Q_\theta(s,a)\}|\mathcal{D}_{t_i-1})$ and acting according to $\mu$ between $t \in \{t_i,\dots,t_{i+1}-1\}$, an informal argument is that in the former case, the sampled $\theta$ induces the optimal policy $\mu^\prime$ such that $\mu^\prime(s) \in \argmax_{a \in \mathcal{A}_{s}}Q_\theta(s,a)$ for all $s \in \mathcal{S}$, which has the same marginal distribution as $\mu$ in the latter case. The former approach then effectively follows $\mu^\prime$, which implies that both approaches result in the same marginal distribution to the subsequent observations of the MDP.\newline

We now verify the equivalence formally and take into account any tie-breaking rules that may be needed when there is a subset of $\theta$ with non-zero probability mass under $p(\cdot|\mathcal{D}_{t_i-1})$ where $\argmax_{a \in \mathcal{A}_s} Q_\theta(s,a)$ is not unique for some $s \in \mathcal{S}$.\newline

When a tie-breaking rule is required, we can extend the approach to consider non-stationary or stochastic policies within $t \in \{t_i,\dots,t_{i+1}-1\}$. Specifically, let $\Pi_d := \{\mu:\mathcal{S} \rightarrow \mathcal{A} | \mu(s) \in \mathcal{A}_s \,\, \forall s \in \mathcal{S} \}$. Let $M_{t_i:t_{i+1}}  := \{M_t\}_{t=t_i}^{t_{i+1}-1}$ be the random variable of a sequence of decision rules, where each $M_t$ is a random variable on $\Pi_d$. We define the marginal probability of selecting a sequence of decision rules $M_{t_i:t_{i+1}-1} = \mu_{t_i:t_{i+1}-1}:= \{\mu_t\}_{t=t_i}^{t_{i+1}-1}$ given $\mathcal{D}_{t_i-1}$, where $\mu_t \in \Pi_d$ for $t \in \{t_i,\dots,t_{i+1}-1\}$, as

$$\!\begin{multlined}[t][0cm]
 \mathbb{P}(M_{t_i:t_{i+1}-1} = \mu_{t_i:t_{i+1}-1}|\mathcal{D}_{t_i-1}):= \\
 \int \mathbb{P}(M_{t_i:t_{i+1}-1}=\mu_{t_i:t_{i+1}-1} | \theta, \mu_{t_i:t_{i+1}-1} \text{ is optimal})\mathbb{P}(\mu_{t_i:t_{i+1}-1}\text{ is optimal}|\theta) p^\Theta(\theta|\mathcal{D}_{t_i-1}) \mathrm{d} \theta,
\end{multlined}$$

where the tie-breaking rule is defined by
\begin{align*}
\!\begin{multlined}[t][0cm]
\mathbb{P}(M_{t_i:t_{i+1}-1} = \mu_{t_i:t_{i+1}-1}| \theta, \mu_{t_i:t_{i+1}-1} \text{ is optimal}) =\\
\prod_{t=t_i}^{t_{i+1}-1} \prod_{s_t^\prime \in \mathcal{S}} \mathbb{P}(M_t(s_t^\prime) = \mu_t(s_t^\prime))|\mu_t(s_t^\prime) \in \argmax_{a \in \mathcal{A}_{s_t^\prime}} Q_\theta(s_t^\prime,a), \theta),
\end{multlined}
\end{align*}
and the definition of optimality for the set of policies implies
$$\mathbb{P}(\mu_{t_i:t_{i+1}-1} \text{ is optimal}|\theta) = \prod_{t=t_i}^{t_{i+1}-1} \prod_{s_t^\prime \in \mathcal{S}} \mathbbm{1}\Big(\mu_t(s_t^\prime) \in \argmax\limits_{a \in \mathcal{A}_{s_t^\prime}} Q_\theta(s_t^\prime,a)\Big).$$

Now, the density of observing the state action sequence $a_{t_i},s_{t_i+1},a_{t_i+1},\dots,s_{t_{i+1}-1},a_{t_{i+1}-1},s_{t_{i+1}+1}$ starting from $s_{t_i}$ and following a sampled policy from $\mathbb{P}(M_{t_i:t_{i+1}-1} = \mu_{t_i:t_{i+1}-1}|\mathcal{D}_{t_i-1})$ is given by
\begin{align*}
&p^{\Pi_d}(s_{t_i+1:t_{i+1}},a_{t_i:t_{i+1}-1}|\mathcal{D}_{t_i-1},s_{t_i})\\
=& \sum_{\mu_{t_i:t_{i+1}-1}} \Big[\prod_{t=t_i}^{t_{i+1}-1} p^S(s_{t+1}|s_t,a_t) \mathbbm{1}(a_t \in \mu_t(s_t)) \Big]\mathbb{P}(M_{t_i:t_{i+1}-1} = \mu_{t_i:t_{i+1}-1} |\mathcal{D}_{t_i-1}) \\
=& \!\begin{multlined}[t][16cm]
\int p^\Theta(\theta|\mathcal{D}_{t_i-1}) \sum_{\mu_{t_i:t_{i+1}-1}} \Bigg[  \Big[ \prod_{t=t_i}^{t_{i+1}-1} p^S(s_{t+1}|s_t,a_t) \mathbbm{1}(a_t \in \mu_t(s_t)) \times \\
\times \Big(\prod_{s_t^\prime \in \mathcal{S}} \mathbb{P}(M_t(s_t^\prime) = \mu_t(s_t^\prime) |\mu_t(s_t^\prime) \in \argmax_{a \in \mathcal{A}_{s_t^\prime}} Q_\theta(s_t^\prime,a), \theta)\mathbbm{1}\Big(\mu_t(s_t^\prime) \in \argmax\limits_{a \in \mathcal{A}_{s_t^\prime}} Q_\theta(s_t^\prime,a\Big)\Big)\Big]\Bigg] \mathrm{d}\theta
\end{multlined}\\
=& \!\begin{multlined}[t][16cm] \int  \Big[ \prod_{t=t_i}^{t_{i+1}-1} p^S(s_{t+1}|s_t,a_t)\mathbbm{1}\Big(a_t \in \argmax\limits_{a\in \mathcal{A}_{s_t}} Q_\theta(s_t,a)\Big) \times\\ \times \mathbb{P}(M_t(s_t) = a_t|a_t \in \argmax_{a \in \mathcal{A}_{s_t}} Q_\theta(s_t,a), \theta) \Big] p^\Theta(\theta|\mathcal{D}_{t_i-1})\mathrm{d}\theta, \end{multlined}\\
\end{align*}
where the final expression is the marginal density of observing the same state action sequence by sampling $\theta \sim p^\Theta(\bigcdot|\mathcal{D}_{t_i-1})$ and acting greedily according to $Q_\theta$ subject to the tie-breaking probabilities.\newline

For choices of the tie-breaking rule, one may simply set, for any $s \in \mathcal{S}$, $a \in \mathcal{A}_s$
$$\mathbb{P}(M_t(s) = a |a \in \argmax_{a^\prime \in \mathcal{A}_{s}} Q_\theta(s,a^\prime), \theta) = \frac{1}{\sum_{a^\prime \in \mathcal{A}_{s}} \mathbbm{1}(Q_\theta(s,a^\prime) = Q_\theta(s,a))}.$$

This corresponds to randomly choosing an optimal action at any time step at any state where more than one action appears to be optimal. Alternatively, for consistency, one may define, for any $s \in \mathcal{S}$, $a \in \mathcal{A}_s$,
\begin{align*}
    \mathbb{P}(M_{t_i}(s) = a|a \in \argmax_{a^\prime \in \mathcal{A}_{s}} Q_\theta(s,a^\prime), \theta) &= \frac{1}{\sum_{a^\prime \in \mathcal{A}_{s}} \mathbbm{1}(Q_\theta(s,a^\prime) = Q_\theta(s,a))} \\
    \mathbb{P}(M_t(s) = a|a \in \argmax_{a^\prime \in \mathcal{A}_{s}} Q_\theta(s,a^\prime), \theta) &= \mathbbm{1}(a=\mu_{t_i}(s))
\end{align*}
for any $t \in \{t_i+1,\dots,t_{i+1}-1\}$.\newline

Therefore, subject to the tie-breaking rules, the practical implementation represented by the latter density can be interpreted as deploying a policy under the posterior distribution of it being optimal for $t_{i+1}-t_i$ steps. For any MDPs with a continuous reward distribution and, as a result, a continuous prior on $\Theta$, though, tie-breaking rules are not required in practice.\newline

\section{More discussions on prior choices}

\label{apd:prior}

In this section, we focus on MDPs (with finite $|\mathcal{A}|$) that have non-goal recurrent states, and we specifically discuss the case where the expected rewards are negative except for the goal state-action pair. An example of such MDPs is one where each action incurs a negative reward as a time penalty, except when the goal is reached. We first show how the support of the prior should be defined to incorporate this information. We then propose a simple relaxation which can be interpreted as only integrating the assumption of the negative rewards but disregarding the structure of $p^S$ at the cost of a prior for easier inference. Priors for other forms of MDPs are left for future studies.

\begin{assumption}
    $\mathbb{E}[R(s,a)] < c$ for any $s \in \mathcal{S} \setminus \mathcal{S}^g$, $a \in \mathcal{A}_s$, where $c < 0$. \label{ass:boundedreward}
\end{assumption}

For any MDP $\mathcal{M}$ unknown up to $p^R$, let 
\begin{align*}
\mathcal{Q}^{\mathcal{M}}  := \{Q \in \{\mathcal{S} \otimes \mathcal{A} \rightarrow \mathbb{R}\}|&  Q(s,a) - \sum_{s^\prime \in \mathcal{S}} \max_{a^\prime \in \mathcal{A}_{s^\prime}}Q(s^\prime, a^\prime) p^S(s^\prime|s,a) < c  \,\, \forall s \in \mathcal{S} \setminus \mathcal{S}^g, a \in \mathcal{A}_s, \\
&Q(s^g, a^g)=0 \,\, \forall s^g \in \mathcal{S}^g, a^g \in \mathcal{A}_{s^g}\}.
\end{align*}
We now show that it is sufficient to restrict the support of prior so that the induced prior of $Q_\theta$ is supported on and only on $\mathcal{Q}^\mathcal{M}$ when the transition dynamics are known and Assumptions \ref{ass:boeunique} and \ref{ass:boundedreward} hold.

\begin{proposition}
    Let $\mathcal{M}$ be a MDP known up to its reward distribution $p^R$ and that Assumption \ref{ass:boeunique} holds. For any $Q^* \in \mathcal{Q}^{\mathcal{M}}$, there exists a $p^R$ satisfying Assumption \ref{ass:boundedreward} such that the $\mathcal{M}$ paired with $p^R$ has $Q^*$ as its optimal action value function and satisfies Assumption \ref{ass:boeunique}. Conversely, for any reward distribution $p^R$ paired with $\mathcal{M}$ that satisfies Assumption \ref{ass:boeunique} and \ref{ass:boundedreward}, the corresponding optimal action value function $Q^* \in \mathcal{Q}^{\mathcal{M}}$. \label{prop:strictprior}
\end{proposition}
\begin{proof}
For any $Q \in \mathcal{Q}^{\mathcal{M}}$, define a deterministic rewards function $r$ such that $r(s,a) := Q(s,a) - \sum_{s^\prime \in \mathcal{S}} \max_{a^\prime \in \mathcal{A}_{s^\prime}}Q(s^\prime, a^\prime) p^S(s^\prime|s,a)$ for any $s \in \mathcal{S}$, $a \in \mathcal{A}_s$. By definition of $\mathcal{Q}^{\mathcal{M}}$, $ r(s,a) < c$ for all $s \in \mathcal{S} \setminus \mathcal{S}^g$, $a \in \mathcal{A}_s$. $Q$ is therefore the optimal action value function of $\mathcal{M}$ with $p^R$ with reward function $r$ by the uniqueness of solution of BOEs. Furthermore, if there exists non-goal recurrent states, the corresponding improper policy must incur negative infinite rewards, and hence the resulting $\mathcal{M}$ with $p^R$ satisfies Assumption \ref{ass:boeunique}.\newline

Conversely, as the $Q^*$ of $\mathcal{M}$ paired with $p^R$ which satisfies Assumptions \ref{ass:boeunique} and \ref{ass:boundedreward} is the unique solution of the BOEs, $\mathbb{E}[R(s,a)] = Q^*(s,a) - \sum_{s^\prime \in \mathcal{S}} \max_{a^\prime \in \mathcal{A}_{s^\prime}}Q^*(s^\prime, a^\prime) p^S(s^\prime|s,a) < c$ for any $s \in \mathcal{S}^g$ by Assumption \ref{ass:boundedreward}. Hence, $Q^* \in \mathcal{Q}^{\mathcal{M}}$ by definition.
\end{proof}

Hence, when a MDP $\mathcal{M}$ is known up to its reward function and that it satisfies Assumptions \ref{ass:boeunique} and \ref{ass:boundedreward}, $\mathcal{Q}^{\mathcal{M}}$ is the set of all possible optimal action value function for $\mathcal{M}$. When $p^S$ is partially available via interactions with the environment, we can define a data-dependent prior of the form $p^\Theta(\theta|\mathcal{D}_\tau^{\mathcal{S},\mathcal{A}})$ with support $\{\theta \in \Theta| Q_\theta \in \{Q \in \{\mathcal{S} \otimes \mathcal{A} \rightarrow \mathbb{R}\}| Q(s,a) - \sum_{s^\prime \in \mathcal{S}} \max_{a^\prime \in \mathcal{A}_{s^\prime}}Q(s^\prime, a^\prime) p^S(s^\prime|s,a) < c  \,\, \forall s,a \in \mathcal{D}_\tau^{\mathcal{S},\mathcal{A}} \text{ such that } s \notin \mathcal{S}^g, \text{ 
 and } \forall s^g \in \mathcal{S}^g, a^g \in \mathcal{A}_{s^g}, Q(s^g, a^g)=0 \}\}$.\newline

However, as the constraint is typically non-convex for non-trivial parametric classes of $Q_\theta$, the resulting posterior would therefore have a non-convex support, making conventional MCMC methods inefficient \citep{hmcneal}. Furthermore, $p^S$ may not be analytically available. A simple relaxation of the constraint would be to consider 
$$\mathcal{Q}^\prime := \{Q \in \{\mathcal{S} \otimes \mathcal{A} \rightarrow \mathbb{R}\}| Q(s,a) < c \,\, \forall s \in \mathcal{S} \setminus \mathcal{S}^g, a \in \mathcal{A}_s,\,\, Q(s^g,a^g)=0 \,\, \forall s^g \in \mathcal{S}^g, a^g \in \mathcal{A}_{s^g}\}.$$
It is clear that $\mathcal{Q}^{\mathcal{M}} \subseteq \mathcal{Q}^\prime$. The following proposition shows that when $p^S$ is also unknown, but $\mathcal{A}$ is known, the support of possible $Q^*$ of such MDPs that satisfies Assumptions \ref{ass:boeunique} and \ref{ass:boundedreward} is $\mathcal{Q}^\prime$.
\begin{proposition}
    Let $\mathcal{M}$ be a MDP known up to its reward distribution $p^R$ and transition distribution $p^S$ and Assumption \ref{ass:boeunique} holds. For any $Q^* \in \mathcal{Q}^\prime$, there exists a $p^S$ and $p^R$ satisfying Assumption \ref{ass:boundedreward} such that $\mathcal{M}$ paired with $p^R$ and $p^S$ has $Q^*$ as its optimal action value function and satisfies Assumption \ref{ass:boeunique}. Conversely, for any $p^R$ and $p^S$ paired with $\mathcal{M}$ that satisfies Assumption \ref{ass:boundedreward}, the corresponding optimal action value function $Q^* \in \mathcal{Q}^{\prime}$. \label{prop:looseprior}
\end{proposition}

\begin{proof}
For any $Q \in \mathcal{Q}^\prime$, define $p^S$ such that $p^S(s^\prime|s,a) := \delta_{s^g}(s^\prime)$ for an $s^g \in \mathcal{S}^g$ and a determinstic reward function $r$ such that $r(s,a):=Q(s,a) < c$ for any $s \in \mathcal{S} \setminus \mathcal{S}^g$, $a \in \mathcal{A}_s$ for some constant $c$. Then, $p^R$ satisfies Assumption \ref{ass:boundedreward} and the MDP therefore satisfies Assumption \ref{ass:boeunique}. Hence, it is clear that $Q$ is the optimal action value function of $\mathcal{M}$ with $p^R$ and $p^S$.\newline

Conversely, for any $p^R$ and $p^S$ satisfying Assumption \ref{ass:boundedreward}, its optimal action value function $Q^* \in \tilde{\mathcal{Q}}^\mathcal{M}$ because all expected rewards are smaller than $c<0$ except for the reward of the goal state-action pairs.
\end{proof}

Hence, this provides a justification to enforce the simpler support $\tilde{\mathcal{Q}}^\prime$ on $Q^*$ when the prior information of $p^S$ is unknown or neglected. Alternatively, one can define a prior with soft constraint on the set $\mathcal{Q}^\prime \setminus \mathcal{Q}^{\mathcal{M}}$, by penalising any $Q \in \mathcal{Q}^\prime \setminus \mathcal{Q}^{\mathcal{M}}$ 
under some suitable distance function, e.g. compute $\min_{\hat{Q} \in \mathcal{Q}^{\mathcal{M}}} d(Q,\hat{Q})$ for some distance $d$, and incorporate the distance into the prior function. Other forms of support are also possible depending on the prior knowledge of $\mathcal{M}$.\newline

\section{More discussions on the sampling methods}

\subsection{Algorithms}
\label{apd:algo}

\begin{algorithm}[ht!]
\caption{HMC (with potential stopping criteria)}
\label{alg:hmc}
\KwIn{number of Leapfrog steps $L$, number of samples $M$, initial sample $\theta_0$, Hamiltonian function $H$ with density $p^\Theta$ and mass matrix $C$, step-size $\delta$, an early $\text{StoppingCriteria}$}
\KwOut{samples of $p^\Theta$ - $\{\theta_m\}_{m=1}^M$}
\For{$m \gets 1$ \KwTo $M$}{
Sample $p_m \sim \mathcal{N}(0,C)$.\\
Set $\tilde{p}_m \gets p_m$, $\tilde{\theta}_m \gets \theta_{m-1}$.\\
\For{$\ell \gets 1$ \KwTo $L$}{
Set $\tilde{p}_m^{\delta/2} \gets \tilde{p}_m + \frac{\delta}{2} \nabla_{\theta} \log p^\Theta(\theta)\Big|_{\tilde{\theta}_m}$.\\
Set $\tilde{\theta}_{m} \gets \tilde{\theta}_m + \delta C^{-1}\tilde{p}_m^{\delta/2}$.\\
Set $\tilde{p}_{m} \gets \tilde{p}_{m}^{\delta/2} + \frac{\delta}{2} \nabla_{\theta} \log p^\Theta(\theta) \Big|_{\tilde{\theta}_{m}}$.
}
Set $\theta_{m} \gets \tilde{\theta}_{m}$ with probability $\min(1,\exp(H((\theta_{m-1}^T,p_m^T)^T) - H((\tilde{\theta}_{m}^T, \tilde{p}_{m}^T)^T))$, otherwise set $\theta_{m} \gets \theta_{m-1}$.\\
\If{$\text{StoppingCriteria}(\{\theta_k\}_{k=1}^{m})$ is reached}{
Set $M \gets m$ and break.
}
}
\end{algorithm}

\begin{algorithm}[ht!]
\caption{SMC for Static Problems with Adaptive HMC Kernel}\label{alg:smc}
\KwIn{unnormalised densities of $\{p^\Theta_j(\cdot)\}_{j=1}^J$ and initial density $p^\Theta_0(\cdot)$, maximum number of HMC steps $M$, initial HMC step-size upper bound $\delta^\star$, initial number of Leapfrog steps upper bound $L^\star$, number of particles $N$, other non-adaptable hyperparameters for HMC}
\KwOut{particle-weight pairs $\{\omega^{j,(n)},\theta^{j,(n)}\}_{j=1,n=1}^{J,N}$ to approximate $p_j^\Theta(\theta)$ as $\sum_{n=1}^N \omega^{j,(n)}\delta_{\theta^{j,(n)}}(\theta)$}
Draw $\theta^{0,(n)} \sim p^\Theta_0(\cdot)$ independently, and set $\omega^{0,(j)} \gets N^{-1}$ for $n \in \{1,\dots N\}$.\\
  \For{$j \gets 1$ \KwTo $J$}{
    $\{\omega^{j,(n)},\theta^{j,(n)}\}_{n=1}^N, \delta^\star, L^\star \gets \text{SMCOneStep}(\{\omega^{j-1,(n)}.\theta^{j-1,(n)}\}_{n=1}^N, \delta^\star, L^\star)$ with Algorithm \ref{alg:smcbase} to update from $p_{j-1}^\Theta(\cdot)$ to $p_j^\Theta(\cdot)$.} 
\end{algorithm}

\begin{algorithm}[ht!]
\caption{SMC for Static Problems with Adaptive HMC Kernel - One Update from $p_{j-1}^\Theta$ to $p_j^\Theta$ (SMCOneStep)}\label{alg:smcbase}
\KwIn{unnormalised densities of $p_{j-1}^\Theta$, $p_{j}^\Theta$, weight-particle pairs $\{\omega^{j-1,(n)}, \theta^{j-1,(n)}\}_{n=1}^N$ approximation of $p_{j-1}^\Theta(\cdot)$, maximum number of HMC steps $M$, initial HMC step-size upper bound $\delta^\star$, initial number of Leapfrog steps upper bound $L^\star$}
\KwOut{weight-particle pairs $\{\omega^{j,(n)}, \theta^{j,(n)}\}_{n=1}^N$ approximation of $p_{j}^\Theta(\cdot)$, updated $\delta^\star$ and $L^\star$}

Set $\omega^{j,(n)} \gets \omega^{j-1,(n)} \frac{p^\Theta_j(\theta^{j-1,(n)})}{p^\Theta_{j-1}(\theta^{j-1,(n)})}$ for $n \in \{1,\dots,N\}$.\\
Set $\omega^{j,(n)} \gets \frac{\omega^{j,(n)}}{\sum_{n=1}^N \omega^{j,(n)}}$ for $n \in \{1,\dots N\}$.\\
\uIf{$\mathrm{ESS}(\{\omega^{j,(n)}\}_{n=1}^N) < N/2$}{Resample $\{\theta^{j-1,(n)}\}_{n=1}^N$ according to the probabilities $\{\omega^{j,(n)}\}_{n=1}^N$ as $\{\bar{\theta}^{j,(n)}\}_{n=1}^N$.\\
Set $\omega^{j,(n)} \gets N^{-1}$ for $n \in \{1,\dots,N\}$.}
\uElse{Set $\bar{\theta}^{j,(n)} \gets \theta^{j-1,(n)}$ for $n \in \{1,\dots,N\}$.}
 $\delta^\star, L^\star, C_j, \{h^{j,(n)}\}_{n=1}^N \gets \text{AdaptKernel}(\delta^\star, L^\star, \{\omega^{j-1,(n)},\theta^{j-1,(n)}\}_{n=1}^N, \{\bar{\theta}^{j,(n)}\}_{n=1}^N, p_j^\Theta)$  using Algorithm \ref{alg:smchmcadptmove}. \\
Draw $\theta^{j,(n)} \sim \kappa^{j}_{h^{j,(n)},C_j}(\bar{\theta}^{j,(n)},\bigcdot)$ for $n \in \{1,\dots,N\}$, where $\kappa^{j}_{h^{j,(n)},C_j}$ is a $p^\Theta_j$-stationary Markov kernel consisting of a maximum of $M$ HMC steps, with step-size $\delta^{j,(n)}$, $L^{j,(n)}$ Leapfrog steps where $(\delta^{j,(n)},L^{j,(n)})=h^{j,(n)}$, and mass matrix $C_j$.
\end{algorithm}

\begin{algorithm}
    \caption{HMC Kernel Adaptation - A Slight Modification of \citep{smchmctuning} (AdaptKernel)}
    \label{alg:smchmcadptmove}
    \KwIn{previous weight-particle approximation $\sum_{n=1}^N \omega^{j,(n)}\delta_{\theta^{j,(n)}}(\theta)$ at time $j$, current particles before MCMC moves $\{\bar{\theta}^{j,(n)}\}_{n=1}^N$, upper bound for step-size, $\delta^\star$, upper bound for number of Leapfrog steps $L^\star$}
    \KwOut{$\delta^\star$, $L^\star$, $C_j$, $\{h^{j,(n)}\}_{n=1}^N$}
    Set $C_{j} \gets \text{diag}(\text{Var}(\{w^{j,(n)},\theta^{j,(n)}\}_{n=1}^N))^{-1}$ ($\text{Var}(\cdot)$ computes the empirical variance of the weighted particles).\\
    \For{$n \gets 1$ \KwTo $N$}{
    Sample $\tilde{\delta}^{j,(n)} \sim \mathcal{U}[0,\delta^\star]$ ($\mathcal{U}$ denotes the uniform distribution).\\
    Sample $\tilde{L}^{j,(n)} \sim \mathcal{U}\{1,\dots,L^\star\}$.\\
    Sample $p \sim \mathcal{N}(0,C_j)$.\\
    Compute $((\tilde{\theta}^{j,(n)})^T,\tilde{p}^T)^T = \hat{\Psi}_{\tilde{L}^{j,(n)},\tilde{\delta}^{j,(n)}}^{C_j,p_{j}^\Theta}(((\bar{\theta}^{j,(n)})^T,p^T))$, i.e. $\tilde{L}^{j,(n)}$ Leapfrog steps with step-size $\tilde{\delta}^{j,(n)}$ with mass matrix $C_j$ targeting $p^\Theta_j$.\\
    Set $\zeta^{j,(n)} \gets H(((\bar{\theta}^{j,(n)})^T,p^T)^T) - H(((\tilde{\theta}^{j,(n)})^T,\tilde{p}^T)^T)$.\\
    Set $\Lambda^{j,(n)} \gets \frac{(\tilde{\theta}^{j,(n)} - \bar{\theta}^{j,(n)})^T C_j^{-1} (\tilde{\theta}^{j,(n)} - \bar{\theta}^{j,(n)})}{\tilde{L}^{j,(n)}} \min(1, \exp(\zeta^{j,(n)}))$.
    }
    Set $\alpha^\star \gets \min_{\alpha \in \mathbb{R}}(\sum_{n=1}^N ||\zeta^{j,(n)}|-\alpha (\tilde{\delta}^{j,(n)})^2|)$.\\
    Set $\delta^\star \gets \max\Big(\sqrt{\frac{|\log(0.9)|}{\alpha^\star}},\max\big(\big\{\tilde{\delta}^{j,(n)}\big| |\zeta^{j,(n)}| < |\log(0.9)| \text{ for } n \in \{1,\dots,N\}\big\}\big)\Big)$.\\
    Sample and set $h^{j,(n)} := (\delta^{j,(n)}, L^{j,(n)}) \sim \sum_{n=1}^N \frac{\Lambda^{j,(n)}}{\sum_{k=1}^N \Lambda^{j,(k)}} \mathbbm{1}((\tilde{\delta}^{j,(n)},\tilde{L}^{j,(n)}=\bigcdot)$ for $n \in \{1,\dots,N\}$.\\
    \uIf{$N^{-1}\sum_{n=1}^N \mathbbm{1}(L^{j,(n)} \in P_{80}(\{\tilde{L}^{j,(k)}\}_{k=1}^N)) > 0.5$ ($P_{l}$ denotes the first $l^{\text{th}}$ percentile)}{Set $L^\star\gets L^\star+5$.}
    \uElseIf{$N^{-1}\sum_{n=1}^N \mathbbm{1}(\tilde{L}^{j,(n)} \in P_{20}(\{\tilde{L}^{j,(k)}\}_{k=1}^N)) > 0.5$ \text{and} $L^\star>5$}{Set $L^\star\gets L^\star-5$.}
\end{algorithm}

\begin{algorithm}
    \caption{ESS Adaptation Scheme (ESSAdapt)}
    \label{alg:smcessscheme}
    \KwIn{target tolerance $\epsilon^\prime$, unnormalised current density $p^\Theta_i(\cdot)$, with approximation $\hat{p}^\Theta_i(\theta) \approx \sum_{n=1}^N \omega^{i,(n)} \delta_{\theta^{i,(n)}}(\theta)$, unnormalised target density as a function of tolerance $ \bar{\epsilon} \mapsto p^\Theta_{\bar{\epsilon}}(\theta)$, ESS reduction factor $\alpha$}
    \KwOut{new tolerance $\epsilon_{i+1}$}
    Set $E \gets \mathrm{ESS}(\{\omega^{i,(n)}\}_{n=1}^N)$.\\
    \uIf{$\mathrm{ESS}\Big(\Big\{\omega^{i,(n)} \frac{p^\Theta_{\epsilon^\prime}(\theta^{i,(n)})}{p^\Theta_i(\theta^{i,(n)})}\Big\}_{n=1}^N\Big) \geq \alpha E$}{Set $\epsilon_{i+1} \gets \epsilon^\prime$.}
    \uElse{Set $\epsilon_{i+1} \gets$  a solution of $f(\bar{\epsilon}) = \text{ESS}\Big(\Big\{ \omega^{i,(n)} \frac{p^\Theta_{\bar{\epsilon}}(\theta^{i,(n)})}{p^\Theta_i(\theta^{i,(n)})}\Big\}_{n=1}^N\Big) - \alpha E = 0$ using bisection.}
\end{algorithm}

\begin{algorithm}
\caption{Simple SMC Updates from $\pabco{\epsilon}(\theta|\mathcal{D})$ to $\pabco{\epsilon^\prime}(\theta|\mathcal{D}^\prime)$ ($\mathcal{D} \subset \mathcal{D}^\prime$) with Adaptive ESS Schedule and $\epsilon^\prime < \epsilon$}
\label{alg:smcdegenerateupdatedetermined}
\KwIn{$\mathcal{D}$, $\mathcal{D}^\prime$, $\epsilon$, $\epsilon^\prime$, weight-particle pairs approximating $\pabco{\epsilon}(\theta|\mathcal{D})$ of the form $\hat{p}_\epsilon(\theta|\mathcal{D}) \approx \sum_{n=1}^N \omega^{(n)} \delta_{\theta^{(n)}}(\theta)$, $\theta^{(n)} \in \Theta$, $0 \leq \omega^{(n)} \leq 1$ for $n \in \{1,\dots,N\}$, $\sum_{n=1}^N \omega^{(n)} = 1$, ESS reduction factor $\alpha$, maximum number of HMC steps $M$, initial upper bound for HMC step-size $\delta^\star$, initial upper bound for number of Leapfrog steps $L^\star$, other non-adaptable hyperparameters for HMC}
\KwOut{weight-particle pairs $\{\omega^{(n)},\theta^{(n)}\}_{n=1}^N$ approximation of $\pabco{\epsilon^\prime}(\theta|\mathcal{D}^\prime)$}

Set $\tilde{\mathcal{D}} \gets \mathcal{D}^\prime \setminus \mathcal{D}$.\\
Set $j \gets 1$.\\
From the current distribution $\pabc(\theta|\mathcal{D})$ and its approximation $\sum_{n=1}^N \omega^{(n)} \delta_{\theta^{(n)}}(\theta)$, find $\tilde{\epsilon}_1$ using Algorithm \ref{alg:smcessscheme} (ESSAdapt) with reduction factor $\alpha$ to target $\bar{\epsilon} \mapsto \pabco{\epsilon,\bar{\epsilon}}(\theta|\mathcal{D},\tilde{\mathcal{D}})$, with target tolerance $\epsilon$.\\
Move and update weight-particle pairs $\{\omega^{(n)},\theta^{(n)}\}_{n=1}^N$ from $\pabc(\theta|\mathcal{D})$ to approximate $\pabco{\epsilon,\tilde{\epsilon}_1}(\theta|\mathcal{D},\tilde{\mathcal{D}})$ using Algorithm \ref{alg:smcbase} (SMCOneStep) with $M$ maximum HMC steps and hyperparameters $\delta^\star, L^\star$. Update $\delta^\star, L^\star$.\\
\While{$\tilde{\epsilon}_j > \epsilon$}{
Set $j \gets j + 1$.\\
From the current distribution $\pabco{\epsilon,\tilde{\epsilon}_{j-1}}(\theta|\mathcal{D},\tilde{\mathcal{D}})$ and its approximation $\sum_{n=1}^N \omega^{(n)} \delta_{\theta^{(n)}}(\theta)$, find $\tilde{\epsilon}_j$ using Algorithm \ref{alg:smcessscheme} (ESSAdapt) with reduction factor $\alpha$ to target $\bar{\epsilon} \mapsto \pabco{\epsilon,\bar{\epsilon}}(\theta|\mathcal{D},\tilde{\mathcal{D}})$, with target tolerance $\epsilon$.\\
Move weight-particle pairs $\{\omega^{(n)},\theta^{(n)}\}_{n=1}^N$ from $\pabco{\epsilon,\tilde{\epsilon}_{j-1}}(\theta|\mathcal{D},\tilde{\mathcal{D}})$ to approximate $\pabco{\epsilon,\tilde{\epsilon}_j}(\theta|\mathcal{D},\tilde{\mathcal{D}})$ using Algorithm \ref{alg:smcbase} (SMCOneStep) with $M$ maximum HMC steps and hyperparameters $\delta^\star, L^\star$. Update $\delta^\star, L^\star$.
}
Set $\epsilon_i \leftarrow \epsilon$ for $i \in \{1,\dots,j\}$.\\
Set $k \leftarrow j$.\\
\While{$\epsilon_k> \epsilon^\prime$}{
Set $k \gets k + 1$.\\
From the current distribution $\pabco{\epsilon_{k-1}}(\theta|\mathcal{D}^\prime)$ and its approximation $\sum_{n=1}^N \omega^{(n)} \delta_{\theta^{(n)}}(\theta)$, find $\epsilon_k$ using Algorithm \ref{alg:smcessscheme} (ESSAdapt) with reduction factor $\alpha$ to target $\bar{\epsilon} \mapsto \pabco{\bar{\epsilon}}(\theta|\mathcal{D}^\prime)$, with target tolerance $\epsilon^\prime$.\\
Move weight-particle pairs $\{\omega^{(n)},\theta^{(n)}\}_{n=1}^N$ from $\pabco{\epsilon_{k-1}}(\theta|\mathcal{D}^\prime)$ to approximate $\pabco{\epsilon_{k}}(\theta|\mathcal{D}^\prime)$ using Algorithm \ref{alg:smcbase} (SMCOneStep) with $M$ maximum HMC steps and hyperparameters $\delta^\star, L^\star$. Update $\delta^\star, L^\star$.
}
Set $\tilde{\epsilon}_{i} \leftarrow \epsilon_i$ for $i \in \{j+1,\dots,k\}$.
\end{algorithm}

\clearpage

\subsection{Choice of mass matrix of HMC}
\label{apd:massmatrix}

To find a suitable mass matrix $C$, the idea is that if $p^\Theta(\bigcdot)$ is approximately Gaussian with covariance matrix $\Sigma = L L^T$, one may construct a Hamiltonian dynamics on the transformed variable $\theta^\prime = L^{-1}\theta$, such that the Hamiltonian is $H^\prime(({\theta^\prime}^T,{p^\prime}^T)^T) = -\log p^\Theta(L\theta^\prime) + {p^\prime}^T p^\prime /2$. This corresponds to sampling from an approximated $2d_\Theta$ dimensional isotropic Gaussian random variable. In fact, an equivalent construction would be to set the mass matrix $C= (LL^T)^{-1}=\Sigma^{-1}$ with $H((\theta^T,p^T)^T)=-\log p^\Theta(\theta) + p^T C^{-1}p/2$, which is essentially the Hamiltonian dynamics on $\theta$ and the transformed variable $p^\prime = L^Tp$ with mass matrix $I$. This choice of the mass matrix $C$, therefore, allows us to use HMC as if we are targeting a distribution with covariance $I$ \citep{hmcneal}. To see this, using the notations of Algorithm \ref{alg:hmc}, the discretised Hamiltonian dynamics associated with $H^\prime$ using the Leapfrog integrator is:
\begin{align*}
    L^Tp_t^{\delta/2} &= {p^\prime_t}^{\delta/2} = p^\prime_t + \frac{\delta}{2}\nabla_{\theta} \log p^\Theta(L\theta)\Big|_{\theta=\theta^\prime_t} = L^T(p_t+\frac{\delta}{2} \nabla_{\theta} \log p^\Theta(\theta) \Big|_{\theta=\theta_t})\\
    L^{-1}\theta_t^{\delta/2} &= {\theta^\prime_t}^{\delta/2} = \theta^\prime_t + \delta  {p_t^\prime}^{\delta/2} = L^{-1}(\theta_t + \delta LL^T p_t^{\delta/2}) = L^{-1}(\theta_t + \delta C^{-1} p_t^{\delta/2}).
\end{align*}

\subsection{Existence of solutions to the ESS adaptive criterion of SMC}
\label{apd:ess}

There are three different ways in which the ESS adaptive criterion is used to find the successive tolerance as described in Algorithm \ref{alg:smcpseudo2}:
\begin{itemize}[align=left]
    \item[Stage I:] $\pabc(\theta|\mathcal{D}) \rightarrow \pabco{\epsilon,\tilde{\epsilon}_1}(\theta|\mathcal{D},\tilde{\mathcal{D}})$, where $\tilde{\epsilon}_1 \geq \epsilon$.
    \item[Stage II:] $\pabco{\epsilon,\tilde{\epsilon}_i}(\theta|\mathcal{D},\tilde{\mathcal{D}}) \rightarrow \pabco{\epsilon,\tilde{\epsilon}_{i+1}}(\theta|\mathcal{D},\tilde{\mathcal{D}})$ where $\tilde{\epsilon}_i \geq \tilde{\epsilon}_{i+1} \geq \epsilon$.
    \item[Stage III:] $\pabco{\epsilon_i}(\theta|\mathcal{D}^\prime) \rightarrow \pabco{\epsilon_{i+1}}(\theta|\mathcal{D}^\prime)$ where $\epsilon_{i} > \epsilon_{i+1} \geq \epsilon^\prime$.
    \item[Stage IVa:] $\pabco{\epsilon_i, \tilde{\epsilon}}(\theta|\mathcal{D},\tilde{\mathcal{D}}) \rightarrow \pabco{\epsilon_{i+1}, \tilde{\epsilon}}(\theta|\mathcal{D},\tilde{\mathcal{D}})$ where $\epsilon_i \leq \epsilon_{i+1} \leq \tilde{\epsilon}$.
    \item[Stage IVb:] $\pabco{\epsilon_i}(\theta|\mathcal{D}^\prime) \rightarrow \pabco{\epsilon_{i+1}}(\theta|\mathcal{D}^\prime)$ where $\epsilon_{i} < \epsilon_{i+1} \leq \epsilon^\prime$.
\end{itemize}
For simplicity, assume that $K_\epsilon$ is a Gaussian kernel with variance $\epsilon^2$.\newline

Stage I: Assume $W=\{\omega^{(n)}, \theta^{(n)}\}_{n=1}^N$ approximates $\pabc(\theta|\mathcal{D})$. Let $\omega^{(n)\prime}(\tilde{\epsilon}_1)$ be the weight update for particle $n$ to approximate $\pabco{\epsilon,\tilde{\epsilon}_1}(\theta|\mathcal{D},\tilde{\mathcal{D}})$ following Table \ref{tab:smcweightupdate} for any $\tilde{\epsilon}_1>0$. Then, 
$$\text{ESS}(\{\omega^{(n)\prime}(\tilde{\epsilon}_1)\}_{n=1}^N)= \frac{\Big[\sum_{n=1}^N \omega^{(n)} \exp\Big(-\frac{1}{2\tilde{\epsilon}_1^2}\big(\sum_{(s,a,r) \in \tilde{\mathcal{D}}}(r-g_{s,a}(\theta^{(n)}))^2\big)\Big)\Big]^2}{\sum_{n=1}^N (\omega^{(n)})^2 \exp\Big(-\frac{1}{\tilde{\epsilon}_1^2}\big(\sum_{(s,a,r) \in \tilde{\mathcal{D}}}(r-g_{s,a}(\theta^{(n)}))^2\big)\Big)},$$
and it is easy to see that $\lim_{\tilde{\epsilon}_1 \rightarrow 0}\text{ESS}(\{\omega^{(n)\prime}(\tilde{\epsilon}_1)\}_{n=1}^N) = 1$ and $\lim_{\tilde{\epsilon}_1 \rightarrow \infty}\text{ESS}(\{\omega^{(n)\prime}(\tilde{\epsilon}_1)\}_{n=1}^N) = \frac{[\sum_{n=1}^N \omega^{(n)}]^2}{\sum_{n=1}^N (\omega^{(n)})^2} = \text{ESS}(\{\omega^{(n)}\}_{n=1}^N)$. Hence, for any $\alpha \in (1/\text{ESS}(\{\omega^{(n)}\}_{n=1}^N), 1)$, there exists an $\tilde{\epsilon}_1 > 0$ such that $\text{ESS}(\{\omega^{(n)\prime}(\tilde{\epsilon}_1)\}_{n=1}^N) = \alpha \text{ESS}(\{\omega^{(n)}\}_{n=1}^N)$ by continuity of the ESS function with respect to $\tilde{\epsilon}_1$. In particular, if $\text{ESS}(\{\omega^{(n)\prime}(\epsilon)\}_{n=1}^N) < \alpha \text{ESS}(\{\omega^{(n)}\}_{n=1}^N)$, there exists an $\tilde{\epsilon}_1$ such that $\epsilon \leq \tilde{\epsilon}_1$ and $\text{ESS}(\{\omega^{(n)\prime}(\tilde{\epsilon}_1)\}_{n=1}^N) = \alpha \text{ESS}(\{\omega^{(n)}\}_{n=1}^N)$ by continuity. \newline

Stage II: Assume $W=\{\omega^{(n)},\theta^{(n)}\}_{n=1}^N$ approximates $\pabco{\epsilon,\tilde{\epsilon}_i}(\theta|\mathcal{D},\tilde{\mathcal{D}})$. Let $\omega^{(n)\prime}(\tilde{\epsilon}_{i+1})$ be the weight update for particle $n$ to approximate $\pabco{\epsilon,\tilde{\epsilon}_{i+1}}(\theta|\mathcal{D},\tilde{\mathcal{D}})$ following Table \ref{tab:smcweightupdate} for any $\tilde{\epsilon}_{i+1} > 0$. Then, $$\text{ESS}(\{\omega^{(n)\prime}(\tilde{\epsilon}_{i+1})\}_{n=1}^N) = \frac{\Big[\sum_{n=1}^N \omega^{(n)} \exp\Big(-(\frac{1}{2\tilde{\epsilon}_{i+1}^2}-\frac{1}{2\tilde{\epsilon}_i^2})\big(\sum_{(s,a,r) \in \tilde{\mathcal{D}}}(r-g_{s,a}(\theta^{(n)}))^2\big)\Big)\Big]^2}{\sum_{n=1}^N (\omega^{(n)})^2 \exp\Big(-(\frac{1}{2\tilde{\epsilon}_{i+1}^2}-\frac{1}{2\tilde{\epsilon}_i^2})\big(\sum_{(s,a,r) \in \tilde{\mathcal{D}}}(r-g_{s,a}(\theta^{(n)}))^2\big)\Big)}.$$

and $\text{ESS}(\{\omega^{(n)\prime}(\tilde{\epsilon}_i)\}_{n=1}^N) = \text{ESS}(\{\omega^{(n)}\}_{n=1}^N)$ and $\text{ESS}(\{\omega^{(n)\prime}(0)\}_{n=1}^N)=1$. Hence, for any $\alpha \in (1/\text{ESS}(\{\omega^{(n)}\}_{n=1}^N), 1)$, there exists an $\tilde{\epsilon}_{i+1} < \tilde{\epsilon}_i$ with $\text{ESS}(\{\omega^{(n)\prime}(\tilde{\epsilon}_{i+1})\}_{n=1}^N) = \alpha \text{ESS}(\{\omega^{(n)}\}_{n=1}^N)$ by continuity with respect to $\tilde{\epsilon}_{i+1}$. In particular, if $\text{ESS}(\{\omega^{(n)\prime}(\epsilon)\}_{n=1}^N) < \alpha \text{ESS}(\{\omega^{(n)}\}_{n=1}^N)$, there exists an $\tilde{\epsilon}_{i+1}$ such that $\epsilon \leq \tilde{\epsilon}_{i+1} \leq \tilde{\epsilon}_i$ and $\text{ESS}(\{\omega^{(n)\prime}(\tilde{\epsilon}_{i+1})\}_{n=1}^N) = \alpha \text{ESS}(\{\omega^{(n)}\}_{n=1}^N)$ by continuity.\newline

Stage III: Using the same argument as Stage II, with $W=\{\omega^{(n)},\theta^{(n)}\}_{n=1}^N$ approximating $\pabco{\epsilon_i}(\theta|\mathcal{D}^\prime)$, and let $\omega^{(n)\prime}(\epsilon_{i+1})$ be the weight to update for $\pabco{\epsilon_{i+1}}(\theta|\mathcal{D}^\prime)$ following Table \ref{tab:smcweightupdate} for any $\epsilon_{i+1}>0$. Then, 
$$\text{ESS}(\{\omega^{(n)\prime}(\epsilon_{i+1})\}_{n=1}^N) = \frac{\Big[\sum_{n=1}^N \omega^{(n)} \exp\Big(-(\frac{1}{2\epsilon_{i+1}^2}-\frac{1}{2\epsilon_{i}^2})\big(\sum_{(s,a,r) \in \mathcal{D}^\prime}(r-g_{s,a}(\theta^{(n)}))^2\big)\Big)\Big]^2}{\sum_{n=1}^N (\omega^{(n)})^2 \exp\Big(-(\frac{1}{2\epsilon_{i+1}^2}-\frac{1}{2\epsilon_i^2})\big(\sum_{(s,a,r) \in \mathcal{D}^\prime}(r-g_{s,a}(\theta^{(n)}))^2\big)\Big)}.$$

Hence, for the same reason, for any $\alpha \in (1/\text{ESS}(\{\omega^{(n)}\}_{n=1}^N), 1)$, there exists an $\epsilon_{i+1} < \epsilon_i$ with $\text{ESS}(\{\omega^{(n)\prime}(\epsilon_{i+1})\}_{n=1}^N) = \alpha \text{ESS}(\{\omega^{(n)}\}_{n=1}^N)$. And, if $\text{ESS}(\{\omega^{(n)\prime}(\epsilon^\prime)\}_{n=1}^N) < \alpha \text{ESS}(\{\omega^{(n)}\}_{n=1}^N)$ and $\epsilon^\prime < \epsilon_i$, there exists an $\epsilon_{i+1}$ with $\epsilon^\prime \leq \epsilon_{i+1}\leq \epsilon_i$ and $\text{ESS}(\{\omega^{(n)\prime}(\epsilon_{i+1})\}_{n=1}^N) = \alpha \text{ESS}(\{\omega^{(n)}\}_{n=1}^N)$.\newline

Stage IVa: With the same argument as Stage II, with $W=\{\omega^{(n)},\theta^{(n)}\}_{n=1}^N$ approximating $\pabco{\epsilon_i, \tilde{\epsilon}}(\theta|\mathcal{D},\tilde{\mathcal{D}})$. Let $\omega^{(n)\prime}(\epsilon_{i+1})$ be the weight update for particle $n$ to approximate $\pabco{\epsilon_{i+1}, \tilde{\epsilon}}(\theta|\mathcal{D},\tilde{\mathcal{D}})$ following Table \ref{tab:smcweightupdate} for any $\epsilon_{i+1} > 0$. Then, if $\text{ESS}(\{\omega^{(n)\prime}(\tilde{\epsilon})\}_{n=1}^N) < \alpha \text{ESS}(\{\omega^{(n)}\}_{n=1}^N)$ and $\tilde{\epsilon} > \epsilon_i$, there exists an $\epsilon_{i+1}$ such that $\epsilon_i \leq \epsilon_{i+1} \leq  \tilde{\epsilon}$ and $\text{ESS}(\{\omega^{(n)\prime}(\epsilon_{i+1})\}_{n=1}^N) = \alpha \text{ESS}(\{\omega^{(n)}\}_{n=1}^N)$ by continuity. However, note that there may not exist an $\tilde{\epsilon} > 0$ such that $\tilde{\epsilon} > \epsilon_i$ and $\text{ESS}(\{\omega^{(n)\prime}(\tilde{\epsilon})\}_{n=1}^N) < \alpha \text{ESS}(\{\omega^{(n)}\}_{n=1}^N)$.\newline

Stage IVb: With the same argument as Stage III,  if $\text{ESS}(\{\omega^{(n)\prime}(\epsilon^\prime)\}_{n=1}^N) < \alpha \text{ESS}(\{\omega^{(n)}\}_{n=1}^N)$ and $\epsilon^\prime > \epsilon_i$, there exists an $\epsilon_{i+1}$ such that $\epsilon_i \leq \epsilon_{i+1} \leq  \epsilon^\prime$ and $\text{ESS}(\{\omega^{(n)\prime}(\epsilon_{i+1})\}_{n=1}^N) = \alpha \text{ESS}(\{\omega^{(n)}\}_{n=1}^N)$ by continuity. However, note that there may not exist an $\epsilon^\prime > 0$ such that $\epsilon^\prime > \epsilon_i$ and $\text{ESS}(\{\omega^{(n)\prime}(\epsilon^\prime)\}_{n=1}^N) < \alpha \text{ESS}(\{\omega^{(n)}\}_{n=1}^N)$.

\subsection{MCMC effectiveness checks for the degenerate case}
\label{apd:mcmceffectiveness}

As discussed in Section \ref{sec:example_landscape}, the difficult posterior landscapes when the tolerances are small and the dataset is incomplete make HMC samplers difficult to propose moves to high-probability regions. A low step-size is therefore required to maintain a reasonable acceptance rate, and as a result, longer MCMC chains are necessary to compensate for the small step-size and move the particles efficiently. However, the number of MCMC iterations per SMC step is usually set as a fixed number with a predetermined terminal tolerance target\citep{chopin2002,abcsmc2011,tdsmc}. Given a fixed computational budget, we argue that monitoring whether MCMC remains effective for each SMC step is essential for the following two reasons and applications: (1) to ensure that the maximum number of MCMC iterations assigned to each SMC step with an appropriate MCMC step-size is sufficient to move the particles in order to mitigate for SMC weight degeneracy, with the option to terminate the MCMC early and proceed to the next target distribution if the MCMC performance is deemed ``satisfactory''; and (2) to address scenarios where the maximum number of MCMC iterations is insufficient by relaxing the target distribution, e.g. increasing the tolerance.\newline

Evaluating MCMC mixing within an SMC framework remains an open research question, however, because the maximum number of MCMC moves within an SMC step is typically small, making conventional MCMC convergence tests prone to high variance and therefore unreliable. Existing solutions to monitor MCMC effectiveness include \citet{smckantas}, which proposed estimating the lag-$M$ correlation as the chain length $M$ increases; \citet{anthonylee_smc_surrogate}, which monitors the ESJD of the MCMC moves and continues until an ESJD-based particle diversification criterion is met.\newline

In this paper, we propose to simply treat the particles as equally weighted independent MCMC chains and monitor both the within-chain variance ($W_i$) and between-chain variance ($B_i$) for each dimension $i$, where
$$B_i := \frac{1}{N-1} \sum_{n=1}^N (\theta^{(n),\bigcdot}_i - \theta^{(\bigcdot),\bigcdot}_i)^2, \quad W_i := \frac{1}{N} \sum_{n=1}^N \frac{1}{M-1} \sum_{m=1}^M (\theta^{(n),m}_i - \theta^{(n),\bigcdot}_i)^2,$$
$$\theta_i^{(n),\bigcdot} := M^{-1}\sum_{m=1}^M \theta^{(n),m}_i, \quad \theta_i^{(\bigcdot),\bigcdot} := (MN)^{-1}\sum_{m=1}^M \sum_{n=1}^N \theta^{(n),m}_i,$$
for $N$ number of $M$-length chains of particles $\{\theta^{(n),m}\}_{m=1}^M$, $n \in \{1,\dots,N\}$. A well-mixed chain should have $B_i \approx M_i$ for most dimensions $i \in \{1,\dots,d_{\Theta}\}$. Motivated by the Gelman-Rubin criterion \cite{gelman-rubbin,gelman-rubin-vats}, we compute the following Gelman-Rubin statistic:

$$\hat{\sigma}_i^2 = \frac{\frac{M-1}{M} W_i + B_i}{W_i}.$$

When the majority of the dimensions meet the criteria $\hat{\sigma}_i^2$ to be less than some pre-specified threshold, we consider the MCMC mixing to be satisfactory (and not ineffective). Note that while the Gelman-Rubin statistic is commonly used to test MCMC convergence from an arbitrary initial distribution and is known for being conservative \citep{mcmcconvergence_forhowmanyiter}, in our case, MCMC is primarily employed within the SMC framework for jittering purposes. Therefore, we do not run the chains for as long as is typically required for a Gelman-Rubin convergence test. Instead, we interpret the statistic as a ratio of between-chain to within-chain average squared moved distances from the mean to ensure the chains are not stuck and have moved adequately, where the threshold set is informed by the Gelman-Rubin interpretation. We found this approach to work well in assessing mixing empirically. Note that the hyperparameter tuning strategy we adopted \citep{smchmctuning} suggested to stop the MCMC early by monitoring the decay of the product of lag-1 correlations for the transformation $\theta_i + \theta_i^2$ across the MCMC iterations for each dimension $i$. However, we did not adopt this approach as the connection between lag-1 correlations and MCMC mixing is not straightforward.\newline

\section{A tabular model-based approach for small state-space}
\label{apd:tabmodelbased}
As discussed in Section \ref{sec:intractllh}, the intractability of the expectation $\mathbb{E}[\max\limits_{a^\prime \in \mathcal{A}_{S_1}} Q_\theta(S_1,a^\prime)|S_0=s,A_0=a]$ for small discrete state space $\mathcal{S}$ stems from the inaccessibility of the analytical form of $p^S$. Hence, an approach to evaluate the expectation is to model the transition probabilities and integrate it into the Bayesian framework along with the $Q_\theta$ function.\newline

For any $s \in \mathcal{S} \setminus \mathcal{S}^g$, $a \in \mathcal{A}_s$, let $\eta_{s,a} \in \{x \in \mathbb{R}^{|\mathcal{S}|}: \sum_{i=1}^{|\mathcal{S}|} x_i = 1\}$ be random variables, and let $\eta_{s,a} \sim \text{Dirichlet}(\alpha_{s,a})$, $\alpha_{s,a} \in \{x \in \mathbb{R}^{|\mathcal{S}|} : x_i > 0\}$ be the hyperparameters. Furthermore, define $\eta := \{\eta_{s,a}\}_{s \in \mathcal{S} \setminus \mathcal{S}^g, a \in \mathcal{A}_s}$, $\alpha := \{\alpha_{s,a}\}_{s \in \mathcal{S} \setminus \mathcal{S}^g, a \in \mathcal{A}_s}$ and let $\xi:\mathcal{S} \rightarrow \{1,\dots,|\mathcal{S}|\}$ be an indexing bijection. Then, we can construct the following generative model.
\begin{align*}
    &p^\pi(s_{1:\tau+1},a_{0:\tau},r_{0:\tau},\theta,\phi,\eta|s_0,a_0,\alpha) \\
    =& \prod_{t=0}^\tau p(r_t|s_t,a_t,\theta,\phi,\eta)p(s_{t+1}|s_t,a_t,\eta) \pi_t(a_t|s_t) p^\Theta(\theta)p^\Phi(\phi)p^E(\eta|\alpha)
\end{align*}
following some policy $\pi_t$ at time $t$, and
\small
\begin{align*}
p(s_{t+1}|s_t,a_t,\eta) &:= \eta_{s_t,a_t,\xi(s_{t+1})}\\
p(r_t|s_t,a_t,\theta,\phi,\eta) &:= \sigma(\phi)^{-1} p^H \Bigg(\sigma(\phi)^{-1} \Big(r_t- \Big(Q_\theta(s_t,a_t) - \sum_{s^\prime \in \mathcal{S}} \max_{a^\prime \in \mathcal{A}_{s^\prime}}Q_\theta(s^\prime,a^\prime) \eta_{s_t,a_t,\xi(s^\prime)}\Big)\Big)\Bigg|s_t,a_t\Bigg) \\
p^E(\eta|\alpha) &:= \prod_{s \in \mathcal{S} \setminus \mathcal{S}^g}\prod_{a \in \mathcal{A}_s} \frac{\Gamma(\sum_{i=1}^{|\mathcal{S}|} \alpha_{s,a,i})} {\prod_{i=1}^{|S|}\Gamma(\alpha_{s,a,i})} \prod_{i=1}^{|\mathcal{S}|} \eta_{s,a,i}^{\alpha_{s,a,i}-1} =  \prod_{s \in \mathcal{S} \setminus \mathcal{S}^g}\prod_{a \in \mathcal{A}_s} \text{Dirichlet}(\eta_{s,a};\alpha_{s,a}).\\
\end{align*}
\normalsize

Note that the rewards are mutually conditionally independent given the corresponding state-action pairs and the unknown variables $\theta,\phi,\eta$. The posterior of interest becomes,
\begin{align*}
&p(\theta,\phi,\eta|s_{0:\tau+1},a_{0:\tau},r_{0:\tau}, \alpha) \propto p(r_{0:\tau},\theta,\phi,\eta|s_{0:\tau+1},a_{0:\tau},\alpha) \\
\propto& \Big[ \prod_{t=0}^\tau p(r_t|s_t,a_t,\theta,\phi,\eta) \Big] p(\theta, \phi,\eta|s_{0:\tau+1},a_{0:\tau}, \alpha) \\
=& \Big[ \prod_{t=0}^\tau p(r_t|s_t,a_t,\theta,\phi,\eta) \Big] p(\eta|s_{0:\tau+1},a_{0:\tau}, \alpha) p^\Theta(\theta) p^\Phi(\phi) \\
=& \Big[ \prod_{t=0}^\tau p(r_t|s_t,a_t,\theta,\phi,\eta) \Big] \prod_{s \in \mathcal{S} \setminus \mathcal{S}^g} \prod_{a \in \mathcal{A}_s} \text{Dirichlet}(\eta_{s,a};\alpha_{s,a}+c^\tau_{s,a}) p^\Theta(\theta) p^\Phi(\phi),
\end{align*}
where $c^\tau_{s,a} \in \mathbb{R}^{|\mathcal{S}|}$, 
$$c^\tau_{s,a,i} = \sum_{t=0}^{\tau} \mathbbm{1}((s,a,\xi^{-1}(i)) = (s_t,a_t,s_{t+1})).$$

As in previous sections, let $\phi$ be a part of $\theta$ for simplicity. To perform posterior sampling with such model-based models for decision-making, given a density estimate $\hat{p}(\theta,\eta|s_{0:\tau+1},a_{0:\tau},r_{0:\tau},\alpha)$ of $p(\theta, \eta|s_{0:\tau+1},a_{0:\tau},r_{0:\tau},\alpha)$, the posterior probability that policy $\mu$ is optimal is simply
\begin{align*}
&p(\{\theta|\forall s \in \mathcal{S}, \mu(s) \in \argmax_{a\in \mathcal{A}_s} Q_\theta(s,a)\}|s_{0:\tau+1},a_{0:\tau},r_{0:\tau},\alpha) \\
=& \int \prod_{s \in \mathcal{S}}\mathbbm{1}\Big(\mu(s) \in \argmax_{a \in \mathcal{A}_s} Q_\theta(s,a)\Big)\hat{p}(\theta|s_{0:\tau+1},a_{0:\tau},r_{0:\tau},\eta) \hat{p}(\eta|s_{0:\tau+1},a_{0:\tau},r_{0:\tau},\alpha) \mathrm{d}\theta \mathrm{d} \eta
.\end{align*}

Thus, an optimal policy can be sampled by first sampling a transition probability function from the posterior transition model followed by sampling an optimal $Q_\theta$ given the transition model.

\section{Miscellaneous}

\subsection{Justifications for using Gaussian kernel for deterministic rewards MDP when \texorpdfstring{$Q^*$}{Q*} does not lie within the parametric class of \texorpdfstring{$Q_\theta$}{Qθ}}
\label{apd:abcemptyset}

Let $\mathcal{D}=\{(s, a, r)| s \in \mathcal{S}, a \in \mathcal{A}_s, r =R(s,a)\}$ be a given dataset, where $R$ is the deterministic reward function, and define the likelihood as 
$$L(\theta|\mathcal{D};\epsilon) = \prod_{(s_i,a_i,r_i) \in \mathcal{D}} \mathcal{N}\Big(r_i;\theta_{\nu(s_i,a_i)} - \sum_{s_i^\prime \in \mathcal{S}} p^S(s_i^\prime|s_i,a_i) \max_{a_i^\prime \in \mathcal{A}_{s_i^\prime}} \theta_{\nu(s_i^\prime,a_i^\prime)},\epsilon^2\Big),$$ and let $p^\Theta$ be the prior over $\Theta$. For simplicity, we assume the following condition on $p^\Theta$:

\begin{assumption}
    For any $\epsilon > 0$, there exists a unique $\theta^* \in \Theta$ such that $p^\Theta(\theta^*)L(\theta^*|\mathcal{D}; \epsilon) = \sup_{\theta \in \Theta} p^\Theta(\theta)L(\theta|\mathcal{D};\epsilon)$. Define the neighbourhood $O_\gamma = \{\theta \in \Theta\big| ||\theta^*-\theta||_2 < \gamma\}$. Then, there exists a $\gamma_1 > 0$ such that for all $0 < \gamma < \gamma_1$, $p^\Theta(O_\gamma) > 0$, and $\inf_{\theta \in O_{\gamma_1}} L(\theta|\mathcal{D};\epsilon)p^\Theta(\theta) > L(\theta^\prime|\mathcal{D};\epsilon)p^\Theta(\theta^\prime)$ for all $\theta^\prime \notin O_{\gamma_1}$.
    \label{ass:prior}
\end{assumption}
We provide a sketch proof for the following statement, which shows that the posterior concentrates on $\theta^*$ as $\epsilon \to 0$:\newline

If $p^\Theta$ satisfies Assumption \ref{ass:prior}, then for any open set $A \subseteq \Theta$,
$$\lim_{\epsilon \rightarrow 0} \hat{p}_\epsilon(A|\mathcal{D}) = \lim_{\epsilon \rightarrow 0} \frac{\int_{\theta \in A}L(\theta|\mathcal{D};\epsilon) p^\Theta(\theta) \mathrm{d} \theta}{\int_{\theta \in \Theta} L(\theta|\mathcal{D};\epsilon) p^\Theta(\theta) \mathrm{d} \theta} = \mathbbm{1}(\theta^* \in A).$$

\begin{proof}[Sketch Proof]
Let
    $$\ell(\theta) := \sum_{(s_i,a_i,r_i) \in \mathcal{D}} -\frac{1}{2} \Big( r_i - \big(\theta_{\nu(s_i,a_i)} - \sum_{s_i^\prime \in \mathcal{S}} p^S(s_i^\prime|s_i,a_i) \max_{a_i^\prime \in \mathcal{A}_{s_i^\prime}} \theta_{\nu(s_i^\prime,a_i^\prime)}\big)\Big)^2,$$
    and $\ell^* := \sup_{\theta \in \Theta} \ell(\theta)$.\newline
    
    Next, suppose $\theta^* \in A$ and for any $\delta^\prime > 0$, define 
    $$B_{\delta^\prime} := \{\theta \in \Theta | \ell(\theta) \geq \ell^* - \delta^\prime\}.$$
    
    By the assumption of $p^\Theta$ and continuity of $\ell(\theta)$, there exists a $\delta > 0$ such that $B_\delta \subseteq A$ and for any $\epsilon > 0$, $\hat{p}_\epsilon(B_\delta|\mathcal{D}) > 0$ and  $\hat{p}_\epsilon(A|\mathcal{D}) > 0$.\newline
    
    Now, we have
    \begin{align*}
    \frac{\hat{p}_\epsilon(A^c|\mathcal{D})}{\hat{p}_\epsilon(A|\mathcal{D})} \leq \frac{\hat{p}_\epsilon(B_\delta^c|\mathcal{D})}{\hat{p}_\epsilon(A|\mathcal{D})} \leq \frac{\hat{p}_\epsilon(B_\delta^c|\mathcal{D})}{\hat{p}_\epsilon(B_\delta|\mathcal{D})} &= \frac{\int_{B_\delta^c} \exp(\ell(\theta))^{1/\epsilon^2}p^\Theta(\theta) \mathrm{d}\theta}{\int_{B_\delta} \exp(\ell(\theta))^{1/\epsilon^2}p^\Theta(\theta) \mathrm{d}\theta} \\
    &=  \frac{\int_{B_\delta^c} \exp(\ell(\theta) - (\ell^*-\delta/2))^{1/\epsilon^2}p^\Theta(\theta) \mathrm{d}\theta}{\int_{B_\delta} \exp(\ell(\theta)-(\ell^*-\delta/2))^{1/\epsilon^2}p^\Theta(\theta) \mathrm{d}\theta}\\
    &\leq \frac{\int_{B_\delta^c} \exp(\ell(\theta) - (\ell^*-\delta/2))^{1/\epsilon^2}p^\Theta(\theta) \mathrm{d}\theta}{\int_{B_{\delta/2}}p^\Theta(\theta) \mathrm{d}\theta}.
    \end{align*}

Taking limit $\epsilon \rightarrow 0$, as the numerator converges to $0$ by dominated convergence theorem, we have $\lim_{\epsilon \rightarrow 0} \frac{\hat{p}_\epsilon(A^c|\mathcal{D})}{\hat{p}_\epsilon(A|\mathcal{D})}=0$, which implies that $\lim_{\epsilon \rightarrow 0} \hat{p}_\epsilon(A^c|\mathcal{D}) = 0$ and $\lim_{\epsilon \rightarrow 0} \hat{p}_\epsilon(A|\mathcal{D}) = 1$.
    
\end{proof}

\subsection{Gradient of \texorpdfstring{$\theta \mapsto g_{s,a}(\theta)$}{g} for tabular \texorpdfstring{$Q_\theta$}{Qθ}}
\label{apd:gradq}
Recall that $g_{s,a}$ has the form:
     $$g_{s,a}(\theta) = Q_\theta(s,a) - \mathbb{E}[\max_{a^\prime \in \mathcal{A}_{S_{1}}}Q_\theta(S_{1}, a^\prime)|S_0=s, A_0=a].$$
 As $\mathcal{S}$ is assumed to be a finite set,
 $$\mathbb{E}[\max_{a^\prime \in \mathcal{A}_{S_{1}}}Q_\theta(S_{1}, a^\prime)|S_0=s, A_0=a] = \sum_{s^\prime \in \mathcal{S}} \max_{a^\prime \in \mathcal{A}_{s^\prime}}Q_\theta(s^\prime, a^\prime) p^S(s^\prime|s,a).$$

 Then, for any $\theta \in \{\theta \in \Theta| \theta_{\nu(s,a)} \neq \theta_{\nu(s,a^\prime)} \,\, \forall s \in \mathcal{S} \setminus \mathcal{S}^g, a,a^\prime \in \mathcal{A}_s, \text{ where } a \neq a^\prime\}$, the set of differentiable $\theta \in \Theta$, 
 $$
 \nabla_\theta g_{s,a}(\theta) =  \nabla_\theta Q_\theta(s,a) - \sum_{s^\prime \in \mathcal{S}} \sum_{a^\prime \in \mathcal{A}_{s^\prime}}  \nabla_\theta Q_\theta(s^\prime, a^\prime) p^S(s^\prime|s,a) \mathbbm{1}\Big(a^\prime \in \argmax\limits_{a^{\prime\prime} \in \mathcal{A}_{s^\prime}}Q_\theta(s^\prime,a^{\prime\prime})\Big).
 $$

 Note that each of the $\argmax$ only contains one element because of the set of differentiable $\theta$ and the fact that for any $s^g \in \mathcal{S}^g$, $\mathcal{A}_{s^g}=\{a^g\}$.\newline

As $Q_\theta(s,a) = \theta_{\nu(s,a)} = \sum_{j=1}^{d_{\Theta}} \theta_{j}\mathbbm{1}(j=\nu(s,a))$ for $s\in \mathcal{S}, a \in \mathcal{A}_s$,
 $$
 \frac{\partial g_{s,a}(\theta)}{\partial \theta_k} = \mathbbm{1}(k=\nu(s,a)) - p^S(s^k|s,a) \mathbbm{1}\Big(a^k \in \argmax\limits_{a^{\prime\prime} \in \mathcal{A}_{s^k}}\theta_{\nu(s^k,a^{\prime\prime})}\Big)
 $$
 for $k \in \{1,\dots,d_{\Theta}\}$, where $(s^k,a^k):= \nu^{-1}(k)$.\newline

Also, for deterministic transition, i.e. for any $s \in \mathcal{S}$, $a \in \mathcal{A}_s$, there exists $s^\prime \in \mathcal{S}$ such that $p^S(\bigcdot|s,a) = \delta_{s^\prime}(\bigcdot)$, the gradient becomes:
 $$
 \frac{\partial g_{s,a}(\theta)}{\partial \theta_k} = \mathbbm{1}(k=\nu(s,a)) - \mathbbm{1}(s^k=s^\prime) \mathbbm{1}\Big(a_k \in \argmax\limits_{a^{\prime\prime} \in \mathcal{A}_{s^k}}\theta_{\nu(s^k,a^{\prime\prime})}\Big).
 $$

Finally, as a simple check for the special case where $s^g \in \mathcal{S}^g$, as $\nu(s^g,a^g) > d_{\Theta}$ and $s^k \notin \mathcal{S}^g$ for any $k \in \{1,\dots,d_\Theta\}$, $\mathbbm{1}(k=\nu(s^g,a^g))=0$ and $p^S(s^k|s^g,a^g)=0$, and hence, $\frac{\partial g_{s^g,a^g}(\theta)}{\partial \theta_k}=0$.\newline